\documentclass[twoside,11pt]{article}

\usepackage{array,booktabs}
\newcolumntype{M}[1]{>{\centering\arraybackslash}m{#1}}
\usepackage[table]{xcolor}
\usepackage{tabu}
\usepackage{automl2019}
\usepackage{subcaption}
\usepackage{float}
\usepackage{amsmath}
\usepackage{amssymb}
\usepackage{cleveref}
\usepackage{subcaption}
\usepackage{units}
\usepackage{wrapfig, blindtext}
\usepackage{hyperref}

\usepackage{tikz}
\usetikzlibrary{arrows,shapes,snakes,automata,backgrounds,petri}

\usetikzlibrary{decorations.shapes}

\usetikzlibrary{calc}

\usetikzlibrary{arrows}
\usepackage{tikz}
\usetikzlibrary{
  arrows.meta, % for Straight Barb arrow tip
  fit, % to fit the group box around the central neurons
  positioning, % for relative positioning of the neurons
}

\tikzset{
  neuron/.style={ % style for each neuron
    circle,draw,thick, % drawn as a thick circle
    inner sep=0pt, % no built-in padding between the text and the circle shape
    minimum size=2.0em, % make each neuron the same size regardless of the text inside
    node distance=2ex and 2em, % spacing between neurons (y and x)
  },
  group/.style={ % style for the groups of neurons
   rectangle,draw,thick, % drawn as a thick rectangle
    inner sep=0pt, % no padding between the node contents and the rectangle shape
  },
  mlp_theta/.style={ % style for the groups of neurons
    rounded rectangle,draw,thick, % drawn as a thick rectangle
    inner sep=0pt, % no padding between the node contents and the rectangle shape
    minimum width=1.5cm,
    minimum height=1cm,
    fill=black!8
  },
  mlp/.style={ % style for the groups of neurons
    rounded rectangle,draw,thick, % drawn as a thick rectangle
    inner sep=0pt, % no padding between the node contents and the rectangle shape
    minimum width=1.5cm,
    minimum height=1cm,
    fill=black!8
  },
  joint/.style={ % style for the groups of neurons
    rounded rectangle,draw,thick, % drawn as a thick rectangle
    inner sep=0pt, % no padding between the node contents and the rectangle shape
    minimum width=1.3cm,
    minimum height=1cm
  },
  lstm/.style={ % style for the groups of neurons
    rounded rectangle,draw,thick, % drawn as a thick rectangle
    inner sep=0pt, % no padding between the node contents and the rectangle shape
    minimum width=3.5cm,
    minimum height=1cm,
    fill=black!8
  },
  io/.style={ % style for the inputs/outputs
    neuron, % inherit the neuron style
    fill=black!15, % add a fill color
  },
  conn/.style={ % style for the connections
    -{Straight Barb[angle=40:2pt 3]}, % simple barbed arrow tip
    thick, % draw in a thick weight to match other drawing elements
  },
  rec/.style={ % style for the connections
    -{Straight Barb[angle=40:2pt 3]}, % simple barbed arrow tip
    very thick, % draw in a thick weight to match other drawing elements
  },
}

\jmlrheading{M. Gargiani and A. Klein and S. Falkner and F. Hutter}

\ShortHeadings{Probabilistic Learning Curve Prediction}{ Gargiani and Klein and Falkner and Hutter}
\firstpageno{1}

\begin{document}

\title{Probabilistic Rollouts for Learning Curve Extrapolation Across Hyperparameter Settings}

\author{\name M. Gargiani \email gargiani@informatik.uni-freiburg.de \\
       \addr University of Freiburg, Germany
       \AND
       \name A. Klein \email kleinaa@informatik.uni-freiburg.de \\
       \addr University of Freiburg, Germany
       \AND
       \name S. Falkner \email stefan.falkner@de.bosch.com \\
       \addr Bosch Center for Artificial Intelligence, Germany
       \AND
       \name F. Hutter \email fh@cs.uni-freiburg.de \\
       \addr University of Freiburg, Germany \newline Bosch Center for Artificial Intelligence, Germany}

\maketitle

\begin{abstract}%   <- trailing '%' for backward compatibility of .sty file
We propose probabilistic models that can extrapolate learning curves of iterative machine learning algorithms, such as stochastic gradient descent for training deep networks, based on training data with variable-length learning curves.
We study instantiations of this framework based on random forests and Bayesian recurrent neural networks. Our experiments show that these models yield better predictions than state-of-the-art models from the hyperparameter optimization literature when extrapolating the performance of neural networks trained with different hyperparameter settings.
%FH: rewrote the abstract.
%Hyperparameter optimization deals with the problem of finding the optimal hyperparameter configuration for a given task. 
%In deep learning scenarios, where the problems are really expensive, it is fundamental to automatize the hyperparameter tuning and avoid the tedious and expensive process of human inspection of learning curves originated by different configuration. 
%In order to make that possible and fasten the optimization procedure, it is important to have a model that is able not only to predict the learning curves originated by new configurations, but also to update its predictions on the fly when new data points become available. 
%In this paper, we inspect the use of roll-out models for this purpose, and we propose a new model based on long short-term memory Bayesian networks for this purpose.
\end{abstract}

\section{Introduction}

The efficient optimization of machine learning hyperparameters is one of the most basic yet most important tasks in automated machine learning (AutoML, \citet{automl-book}). E.g., hyperparameter optimization has already achieved remarkable improvements of the state-of-the-art in different applications, such as natural language processing~\citep{melis-iclr18} or AlphaGO~\citep{chen-arxiv18a}.
A wide range of hyperparameter optimization methods exists (see, e.g., \citet{feurer-automlbook18a} for an overview), and since the objective function of interest (e.g., cross-validation error) is typically expensive, the most efficient methods tend to leverage cheap-to-evaluate proxies (so-called fidelities)~\citep{swersky-arxiv14, domhan-ijcai15, baker-arxiv17, kandasamy-icml17, klein-ejs17,klein-iclr17, li-iclr17,falkner-icml18}.

A frequently used fidelity for iterative machine learning algorithms is the performance over time or iterations, the so-called \emph{learning curve}: the early performance of a network architecture or hyperparameter configuration is typically quite indicative of its final performance when trained to convergence. Some approaches model these learning curves to decide whether to stop or continue the evaluation of a hyperparameter configuration~\citep{swersky-arxiv14, domhan-ijcai15, baker-arxiv17, klein-iclr17, li-iclr17, falkner-icml18}, while others actively choose a budget before evaluating in order to maximize the information gained per time spent~\citep{klein-ejs17, kandasamy-icml17}. 
%Here, the global model combines the information from different configurations evaluated on different fidelities.

Another key difference between previous methods lies in what the model predicts based on what information.
Several approaches~\citep{swersky-arxiv14, kandasamy-icml17, klein-iclr17, klein-ejs17} build a global model capable of predicting the performance at any fidelity based on the hyperparameter configuration alone.
Others~\citep{baker-arxiv17, falkner-icml18} only train models that predict the learning curve for a fixed set of fidelities, and a third group~\citep{li-iclr17, domhan-ijcai15} only operates on single learning curves and extrapolates them without taking the hyperparameter configuration into account.

A final notable distinction are the assumptions going into the model.
Many existing methods use hand-designed basis functions describing common characteristics of learning curves~\citep{domhan-ijcai15, klein-ejs17, klein-iclr17, swersky-arxiv14}, while others~\citep{baker-arxiv17, kandasamy-icml17, li-iclr17, falkner-icml18} make no or very weak assumptions about the shape of the learning curves, but rely more heavily on observed training data.

Surprisingly, none of the existing methods truly takes into account the sequential nature of learning curves by using a sequence model that can be rolled out for an arbitrary number of time steps. In this paper, we fill this gap; our contributions are as follows:
\vspace{-2mm}
\begin{itemize}
\addtolength\itemsep{-3mm}
    \item {We introduce the first sequence models for learning curve prediction. We provide instantiations based on random forests and Bayesian neural networks that also take hyperparameter configurations 
    %(and in principle also meta-features) 
    into account.} 
    \item {These sequence models are the first that can cheaply generate extrapolations of partially observed learning curves with similar characteristics to those in the training data.}
    \item {In preliminary experiments, we show that these models are not only more flexible and accurate than previous learning curve models, but also allow to efficiently transfer knowledge to new tasks with the same input domain.}
\end{itemize}

\section{Probabilistic Prediction of Learning Curves}

Previous work~\citep{swersky-arxiv14, klein-iclr17} casts the prediction $\tilde{y}_t \in \mathbb{R}$ of the performance $y_t \in \mathbb{R}$ of a hyperparameter configuration $\boldsymbol{\theta} \in \Theta$ at a time step $t \in \mathbb{R}$ as a mapping $\tilde{y}_t = g(\mathbf{x}_t; \boldsymbol{\omega})$ 
with $\mathbf{x}_t=[\boldsymbol{\theta}^{\top}, t]^{\top}$ and $ \boldsymbol{\omega}$ being the collection of the model parameters.

%The models for learning curves prediction can be divided into two main categories: non roll-out models and roll-out models. Given a learning curve $\mathbf{y}=\left[y_1, \dots, y_T\right]^{\top}$ and a configuration vector $\boldsymbol{\theta}\in \mathbb{R}^M$, the non roll-out models learn the mapping $y_t = f(\mathbf{x}_t; \boldsymbol{\omega})$, with $\mathbf{x}=[\boldsymbol{\theta}^{\top}, t]^{\top}$ and $ \boldsymbol{\omega}$ being the collection of the model's parameters. 
Instead, we treat learning curves as sequential time series and predict the value at the current time step based on the values observed at previous time steps.
More formally, we keep the same mapping $\tilde{y}_{t}=g(\mathbf{x}_t; \boldsymbol{\omega})$ but augment the input $\mathbf{x}_t = \left[\boldsymbol{\theta}^{\top}, y_{t-K-1}, \dots, y_{t-1}\right]^\top$ by the past $K$ observed points $y_{t-K-1}, \dots, y_{t-1}$.
We assume that the unknown true objective function $f(\boldsymbol{\theta}, t)$ is only observable with noise $y_t = f(\boldsymbol{\theta}, t) + \epsilon$, with $\epsilon \sim \mathcal{N}(0, \sigma^2).$  
%In the Bayesian probabilistic scenario, $y_t$ is considered to be a noisy observation of the true unknown value and $\boldsymbol{\omega}$ a collection of random variables. The scope of the models is then to determine
To predict for an unseen data point $\mathbf{x}_t^{\star}$ during inference time, we approximate the predictive distribution by a Gaussian:  
\begin{equation}\label{eq:model}
p(y_t^{\star} \mid \mathbf{x}_t^{\star}, \mathcal{D}) \approx \mathcal{N}\big(\mu(y^{\star}_t\mid \mathbf{x}_t^{\star}, \mathcal{D}),  \sigma^2(y^{\star}_t\mid \mathbf{x}_t^{\star}, \mathcal{D})\big)
%p(y_t^{\star} \mid \mathbf{x}_t^{\star}, \mathcal{D}) = \int p(y_t^{\star}|\mathbf{x}_t^{\star}, %\boldsymbol{\omega})\, p(\boldsymbol{\omega}|\mathcal{D})d\boldsymbol{\omega} \,\,,   
\end{equation}
where $\mathcal{D} = \{\mathbf{x}^{(0)}, \mathbf{y}^{(0)}, \dots \mathbf{x}^{(N-1)}, \mathbf{y}^{(N-1)}\}$ is the training dataset that consists of $N$ learning curves with potentially varying lengths, together with their corresponding hyperparameter configuration vectors.
We now describe how to predict $\mu(y^{\star}_t\mid \mathbf{x}_t^{\star}, \mathcal{D})$ and $\sigma^2(y^{\star}_t\mid \mathbf{x}_t^{\star}, \mathcal{D})$ in Equation~\ref{eq:model} using two different probabilistic regression models: random forests (RFs) and variational recurrent neural networks (VRNNs).
%\frank{Can you please replace the abbreviation RFR with the standard one RF everywhere (including in figures)? (RFR is only the name of Stefan's RF package.)}

\subsection{Random Forests}

%A really intuitive roll-out probabilistic model for learning curve prediction can be obtained by using random forests~\citep{breimann-mlj01a}.
First, we consider random forests~\citep{breimann-mlj01a} because of their conceptual simplicity and practical robustness against their own hyperparameters.
%Given a learning curve $\mathbf{y}=  \left[y_1, \dots, y_T\right]^{\top}$ and the corresponding hyperparameter configuration $\boldsymbol{\theta}\in \Theta$, we use a sliding window approach with size $K$ and train the random forest on a collection of joint inputs $\mathbf{x}_t = \left[y_{t-K-1},\dots,y_{t-1}, \boldsymbol{\theta}^{\top}\right]^{\top}$ to predict for the next point in the learning curve $\hat{y}_{t}$.
%\newpage
Following~\citet{hutter-aij14a}, given a forest with $B$ trees, each tree $i$ stores the empirical mean $\tilde{\mu}_i$ and variance $\tilde{\sigma}^2_i$ and, for a test point $\mathbf{x}_t^{\star}$, the forest returns a Gaussian predictive distribution
$\mathcal{N}(\tilde{\mu}(y_t^{\star} \mid \mathbf{x}_t^{\star}, \mathcal{D}),\tilde{\sigma}^2(y_t^{\star} \mid \mathbf{x}_t^{\star}, \mathcal{D}))$
where
$\tilde{\mu}(y_t^{\star} \mid \mathbf{x}_t^{\star}, \mathcal{D}) =\nicefrac1B \sum_i \tilde{\mu}_{i}$
is the mean of the individual tree predictions and 
$\tilde{\sigma}^2(y_t^{\star} \mid \mathbf{x}_t^*, \mathcal{D}) = \nicefrac1B \cdot \sum_i \tilde{\sigma}^2_i +  \nicefrac1B \cdot \sum_i [\tilde{\mu}_i - \tilde{\mu}(y_t^{\star} \mid \mathbf{x}_t^*, \mathcal{D})]^2$
is computed based on the law of total variance.
At inference time, this model requires access to the first $K$ points of an unseen learning curve, but can then extend these to arbitrary length, which we call a \textit{roll out}.
For a single roll out, we sample $y^*_{K+1}$ from the predictive distribution defined above. This process can then be consecutively applied until a whole sequence $[\tilde{y}^r_{K+1}, \ldots ,\tilde{y}^r_{T}]$ is generated up to some time step $T$.
By averaging over $R$ independent roll outs, we approximate Equation~\ref{eq:model} by a Gaussian with mean $\mu(y^{\star}_t\mid \mathbf{x}_t^{\star}, \mathcal{D}) = \frac{1}{R}\sum_{r=1}^R \tilde{y}^{r}_{t}$ and variance $\sigma^2(y^{\star}_t\mid \mathbf{x}_t^{\star}, \mathcal{D}) = \frac{1}{R}\sum_{r=1}^R (\tilde{y}^{r}_{t} - \mu(y^{\star}_t\mid \mathbf{x}_t^{\star}, \mathcal{D}))^2$.

%Now to predict $y^{\star}_t \in \mathbf{y}^{\star} = (y^{\star}_0, \ldots, y^{\star}_T)$ at any time step $t$ we first roll out our model $R$ times by feeding back the predictions produced by the model at earlier time steps by either using the mean $\tilde{\mu}$ or a sample $\tilde{y_t} \sim \mathcal{N}(\tilde{\mu}, \tilde{\sigma}^2)$ from the predictive distribution of the random forest.
%This results in a set of $R$ roll outs $\{\hat{y}^{\star}_{t0}, \ldots, \hat{y}^{\star}_{tr}\}$ for $y^{\star}_t$  and

%Each realization $\tilde{\mathbf{y}}_{t}$ is obtained by sampling from a Gaussian distribution $\mathcal{N}\left(m_{t}, \sqrt{v}_{t}\right)$, where $m_{t}$ and $ {v}_{t}$ are respectively the mean and variance of the model's prediction for ${\mathbf{y}}_{t}$, generated by its inherent random nature.

\subsection{Variational Recurrent Neural Networks (VRNNs)}\label{vrnn_model}

%\frank{I don't see the reason for the notation $\mathbf{h}^{1}_{t-1}= h_1(\boldsymbol{\theta})$; why not just $\mathbf{h}_1= h_1(\boldsymbol{\theta})$? In contrast to y, this actually does not depend on the epoch, so there shouldn't be a $t$ or $t-1$ subscript.}

%Learning curves are intrinsically time series data. Therefore, the learning curve prediction task is at best addressed by models such as recurrent neural networks~\citep{minka-uai01}, and their long short-term memory (LSTM) version~\citep{hochreiter-neural97}, which, with their structure, are able to account for the ineherent sequenciality of these data. 
\usetikzlibrary{shapes.misc, positioning}
\tikzset{decorate sep/.style 2 args=
{decorate,decoration={shape backgrounds,shape=circle,shape size=#1,shape sep=#2}}}
\begin{wrapfigure}{R}[10pt]{4.5cm}\centering
\vspace{-10pt}
\resizebox{3cm}{7cm}{
\begin{tikzpicture}
      \node[neuron] (vt) {$\mathbf{y}$};
      \node[neuron,left=5em of vt] (kt) {$\boldsymbol{\theta}$};  
     \node[mlp_theta,above=1em of kt] (mlp_config) {$h_1$};%{$\textbf{MLP}_{\theta}$};
      \node[joint, above=1.4cm of vt] (joint_input) {$[\mathbf{h}_1\odot \mathbf{z}_1 , \mathbf{y}]$};
      \node[lstm,above =3.5cm of vt] (lstm_1) {$r_1$};%{\textbf{LSTM 1}};
     %\node[mlp,above=1.5em of lstm_1] (mlp_1) {\textbf{MLP 1}};
      \node[neuron,above=10em of mlp_config] (theta_2) {$\boldsymbol{\theta}$}; 
     \node[mlp_theta,above=1em of theta_2] (mlp_config2) {$h_2$};%{$\textbf{MLP}_{\theta}$};
     \node[joint, above=2.6cm of lstm_1] (joint_input2) {$[\mathbf{h}_2\odot \mathbf{z}_2 , \mathbf{\tilde{h}}_1]$};
      
     \node[lstm,above=5.5cm of lstm_1] (lstm_2) {$r_2$};%{\textbf{LSTM 2}};
     \node[mlp,above=1.5em of lstm_2]   (mlp_2) {$h_3$};%{\textbf{MLP}};
      \node[neuron,above=1.4em of mlp_2] (output) {$\tilde{\mathbf{y}}$};  
      \draw[conn] (vt) -- (joint_input);
      \draw[conn] (joint_input) -- (lstm_1);  
      \draw[conn] (kt) -- (mlp_config);
      \draw[conn] (mlp_config) |- (joint_input);
      \draw[conn] (lstm_1) -- (joint_input2);
      %\draw[decorate sep={2mm}{4mm},fill] (lstm_1) -- (mlp_1);
      \draw[conn] (lstm_2) -- (mlp_2);
      \draw[conn] (mlp_2) -- (output);
      \draw[conn] (theta_2) -- (mlp_config2);
      \draw[conn] (mlp_config2) |- (joint_input2);
      \draw[conn] (joint_input2) -- (lstm_2);
      %\draw[thick,->] (lstm_1.160) arc (1:264:5mm);
      \draw[rec] (lstm_1) to [looseness=5, out= 345, in=15]  (lstm_1);
      \draw[rec] (lstm_2) to [looseness=5, out= 345, in=15]  (lstm_2);
      %\draw[conn] (mlp_1) -- (joint_input2);
    \end{tikzpicture}}
    \caption{VRNN model.}
    \label{fig:vrnn_model_graph}
%    \vspace{-10pt}
\end{wrapfigure}
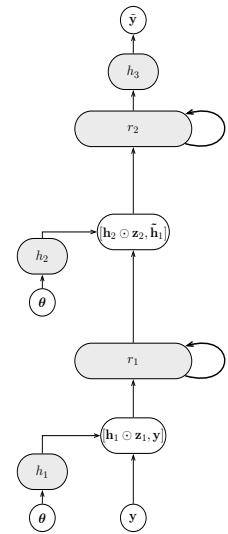
Due to their success for time series prediction (e.g. \citet{rangapuram-nips18}), we also consider recurrent neural networks in form of long short-term memory (LSTM) cells~\citep{hochreiter-neural97}. 
To obtain uncertainty estimates, we use variational dropout~\citep{gal-icml16,gal-nips16} to allow for a Bayesian treatment of the weights.
Given a  hyperparameter configuration $\boldsymbol{\theta}$, a dropout rate $d \in (0,1)$, and the previous observed point in the learning curve $y_{t-1}$, we predict the next step by:
\begin{equation*}
    \tilde{y}_t = h_3\left( r_2\left(  \mathbf{h}_2\odot\mathbf{z}_2 , r_1\left( \mathbf{h}_1\odot\mathbf{z}_1, y_{t-1} \right) \right)  \right),
\end{equation*}
where
\begin{align*}
\mathbf{h}_{1}&= h_1(\boldsymbol{\theta}) &  \mathbf{h}_{2} &= h_2(\boldsymbol{\theta})\\
\mathbf{z}_1 &\sim Bernoulli(1-d)          &\mathbf{z}_2&\sim Bernoulli(1-d)
\end{align*}
and $\odot$ denotes the element-wise product, $r_i(\cdot)$ are LSTM blocks and $h_i(\cdot)$ feedforward neural networks.
See Figure~\ref{fig:vrnn_model_graph} and Section~\ref{appendix_model_graph} in the Appendix for a graphical representation of our model, where we used $\mathbf{\tilde{h}}_1$ to indicate the output of $r_1$.
As described above for the random forest, to predict for an unobserved point in a learning curve $y^{\star}_t$ at any time step $t$, we first perform $R$ rollouts by keeping dropout active and feed the prediction of our model back to itself.
%\aaron{If I am not mistaken, the VDRNN actually doesn't have a predictive distribution and just predicts a single value which is returned to model.}
%\maty{I think it has. See paper 'Dropout as a Bayesian Approximation' eq.(5)}
The final prediction is then the mean and variance of the rollouts (\textit{MC dropout}).
Note that, compared to the random forest, we set $K=1$ and implicitly accumulate memory of the previous observed points in the hidden state of the LSTMs. 
By introducing a dummy value $y_0 = 0$, we do not even require to observe any points of the learning curve at test time.

\section{Experiments}

In this section, we first empirically evaluate the performances of our probabilistic models, dubbed VRNN and RF, and compare them against other non roll-out state-of-the-art probabilistic regression models for learning curve prediction.
Afterwards, we present some preliminary results that show the potential of our model to predict learning curves from unseen datasets, and that hint towards possible meta-learning and transfer-learning extensions of this work.
%Afterwards, we present preliminary experiments showing the potentials of our model for multi-task hyperparameter optimization. 

%\subsection{Settings}
\subsection{Learning Curve Prediction}\label{setting_subsection}

\begin{figure}[t]
%\vspace*{-1cm}
    \begin{center}
        \includegraphics[width=.24\textwidth]{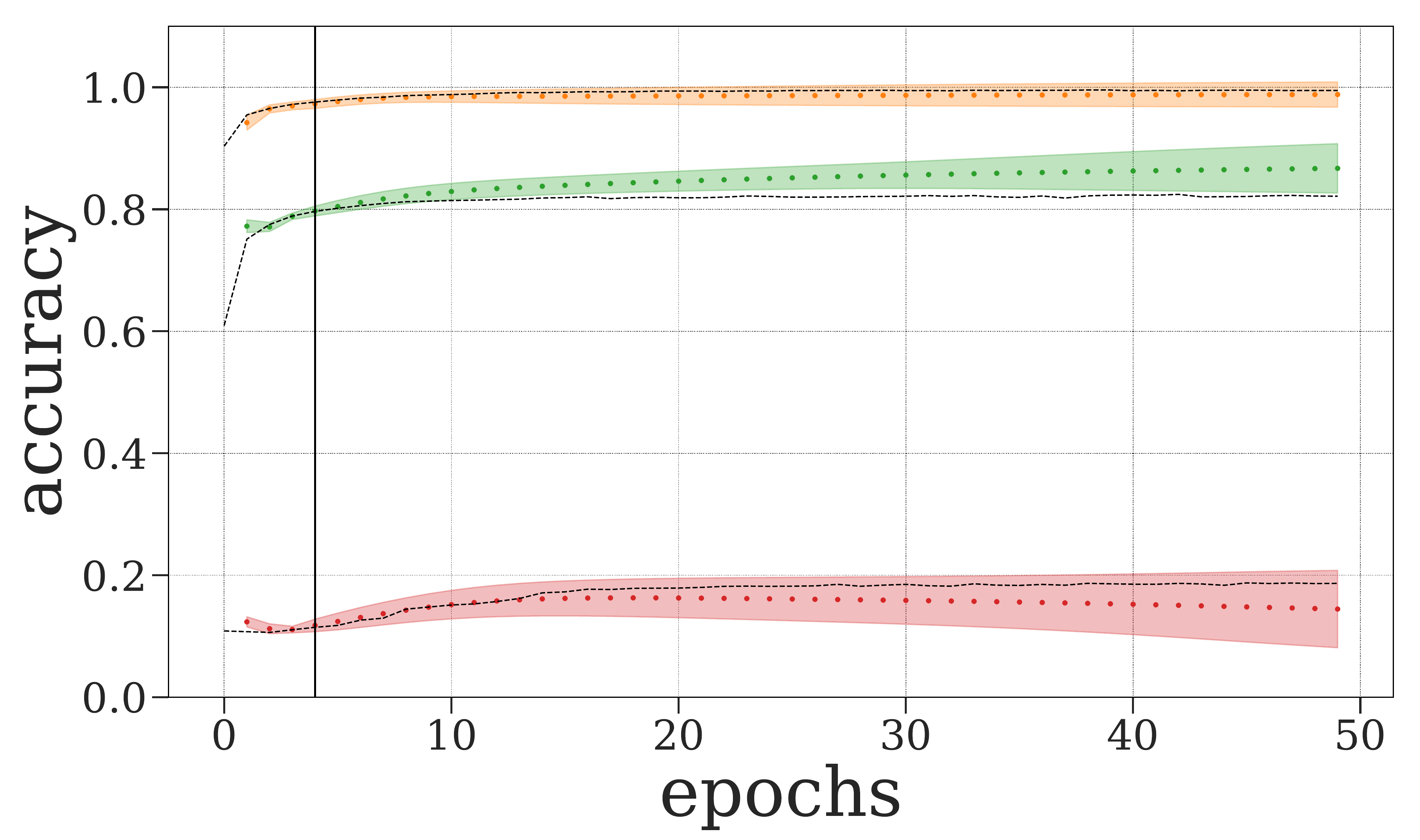}
        \includegraphics[width=.24\textwidth]{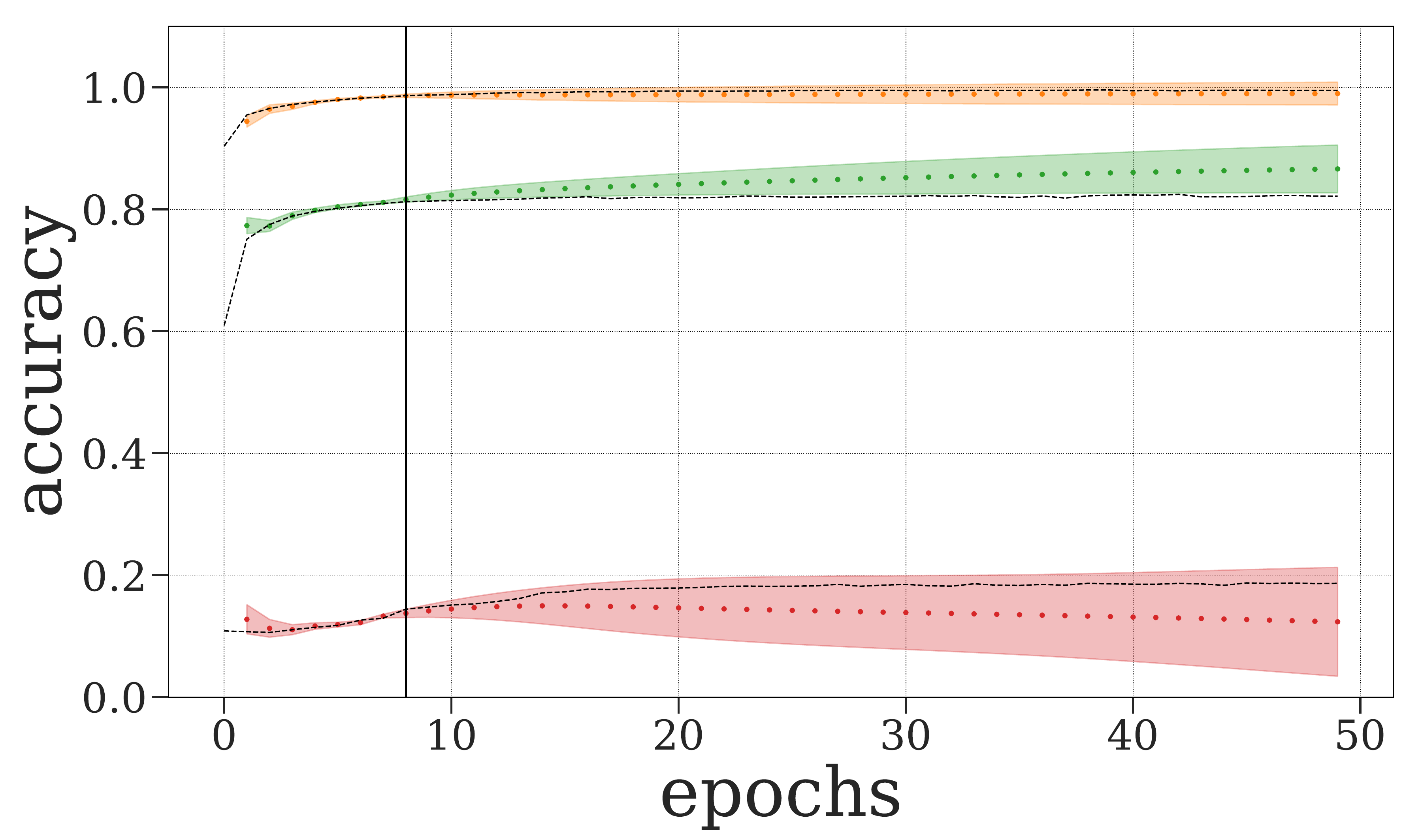}
        \includegraphics[width=.24\textwidth]{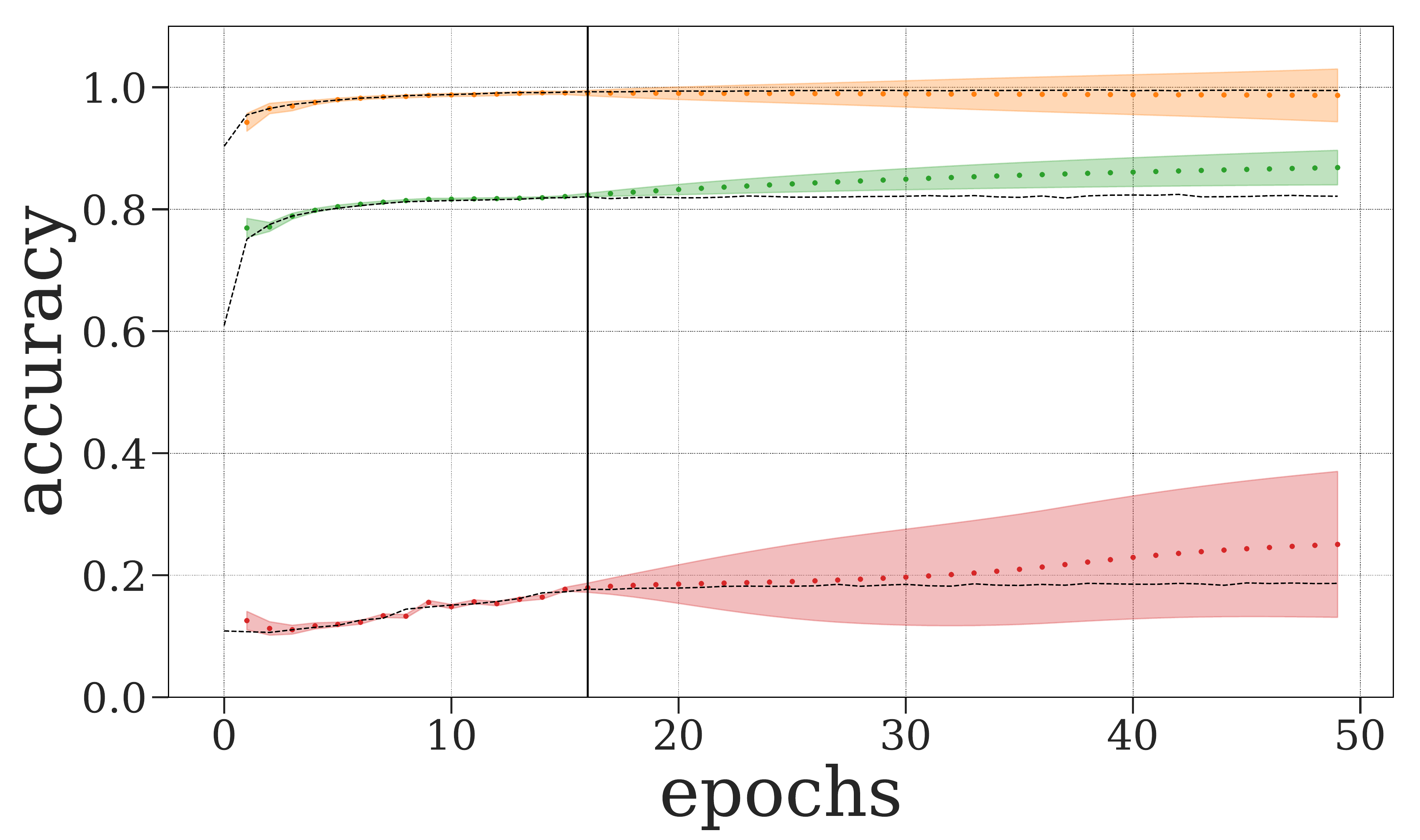}
        \includegraphics[width=.24\textwidth]{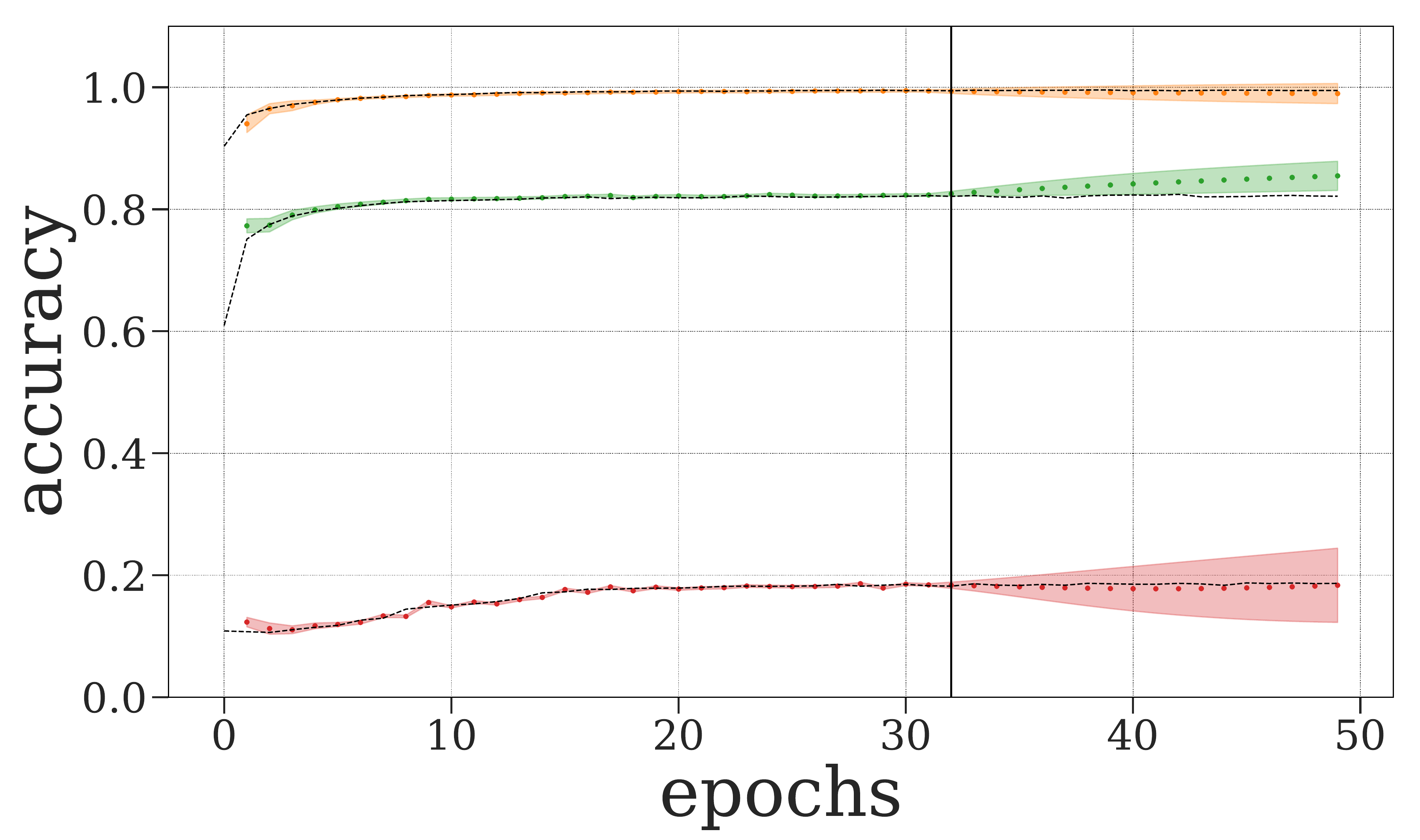}
    \end{center}
    \vspace*{-0.6cm}
    \caption{Qualitative assessment of the test roll-out performances of VRNN for different numbers of observed epochs (black vertical line) on the MNIST benchmark. Different colors stand for different configurations. The shaded area corresponds to $1\sigma$.}
    \label{fig:mnist_unrolling}
\end{figure}
%Different colors stands for different configurations, the dotted line is the predicted mean and the shaded area corresponds to $1\sigma$, the black dashed line is the true learning curve, and the black vertical line indicates the number of observed epochs.
\begin{figure}[t]
\begin{subfigure}{.24\textwidth}
  \centering
  \includegraphics[width=\linewidth]{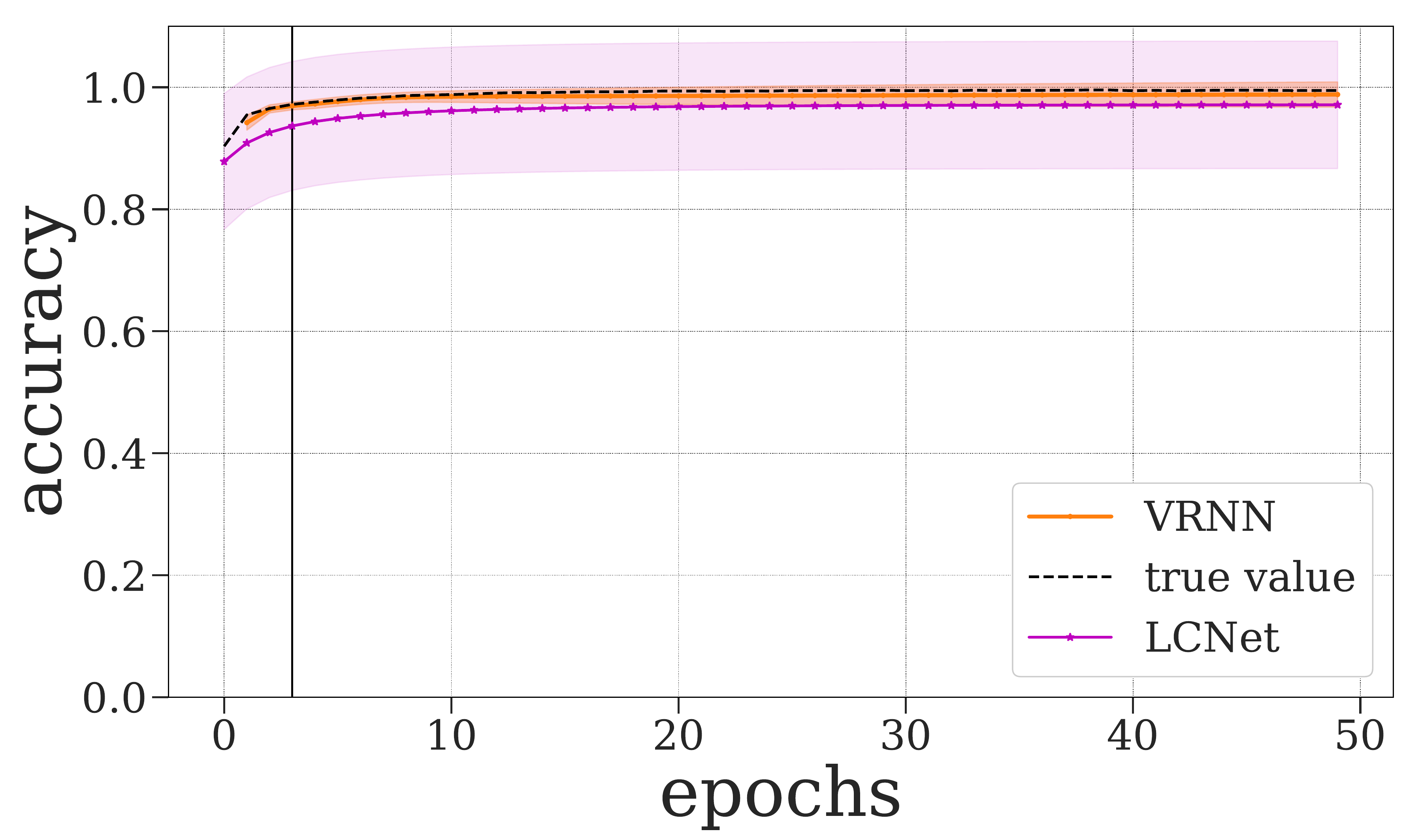}
  \caption{\small{4 observed epochs}}
  \label{fig:e}
\end{subfigure}%
\begin{subfigure}{.24\textwidth}
  \centering
  \includegraphics[width=\linewidth]{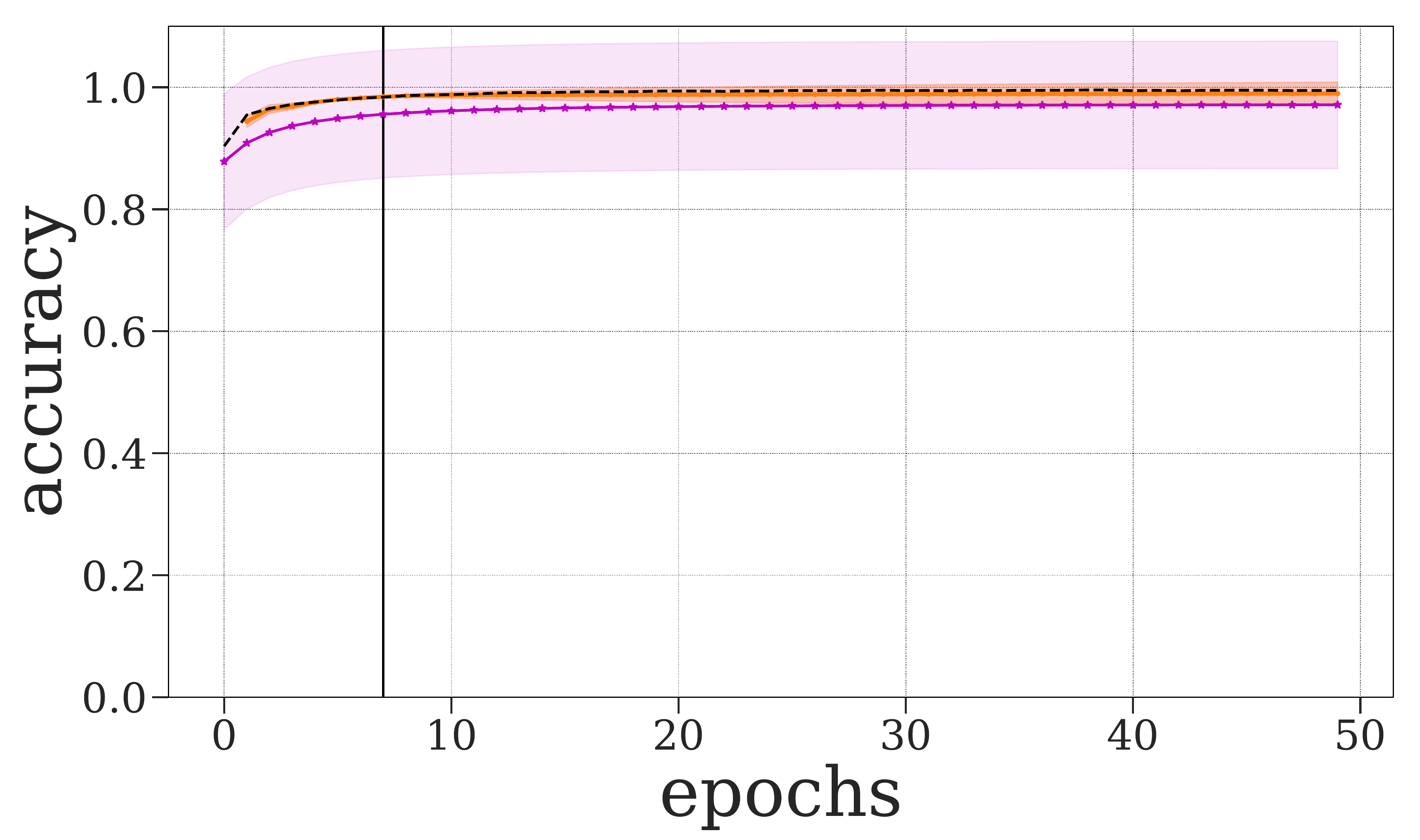}
  \caption{8 observed epochs}
  \label{fig:f}
\end{subfigure}
\begin{subfigure}{.24\textwidth}
  \centering
  \includegraphics[width=\linewidth]{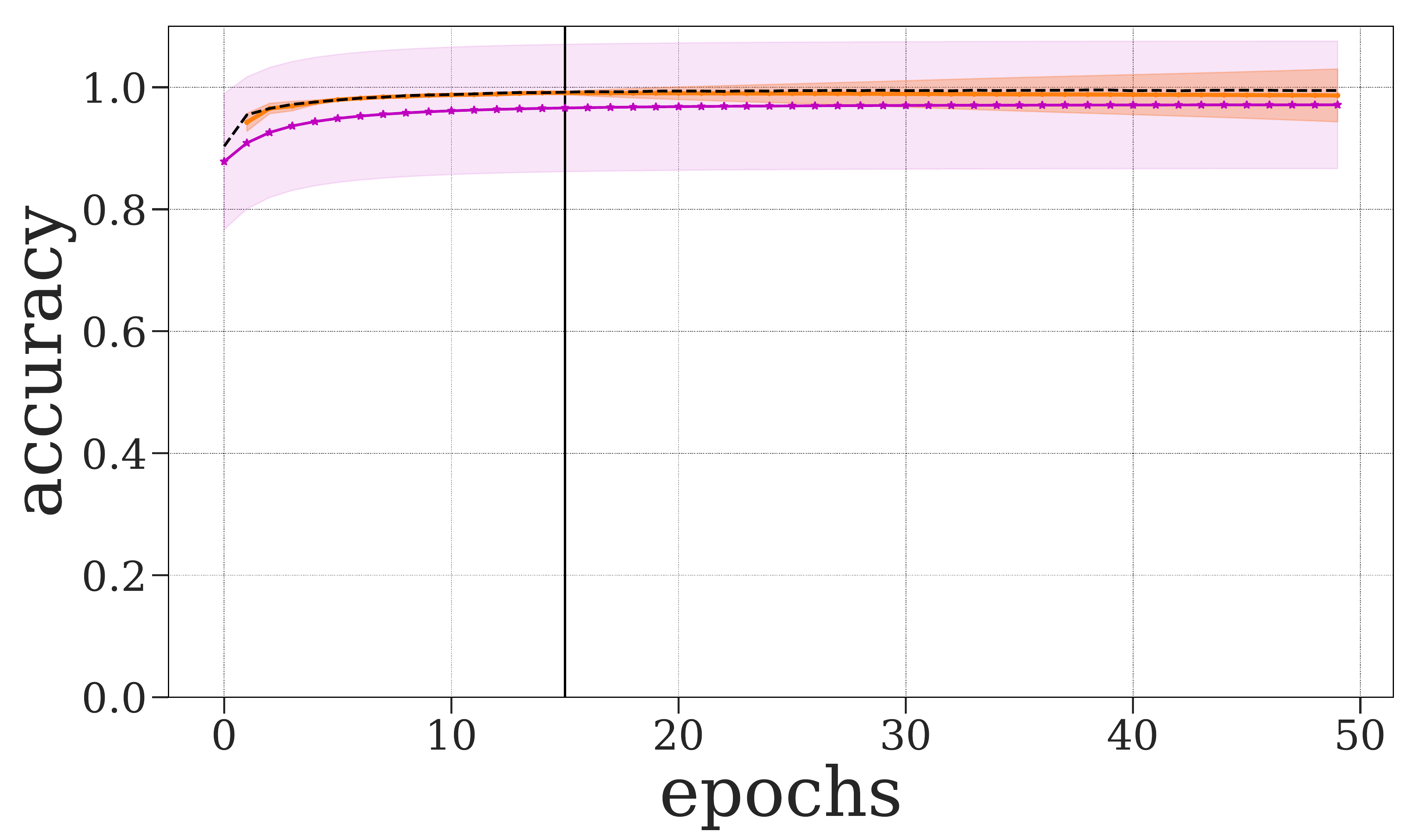}
  \caption{16 observed epochs}
  \label{fig:g}
\end{subfigure}
\begin{subfigure}{.24\textwidth}
  \centering
  \includegraphics[width=\linewidth]{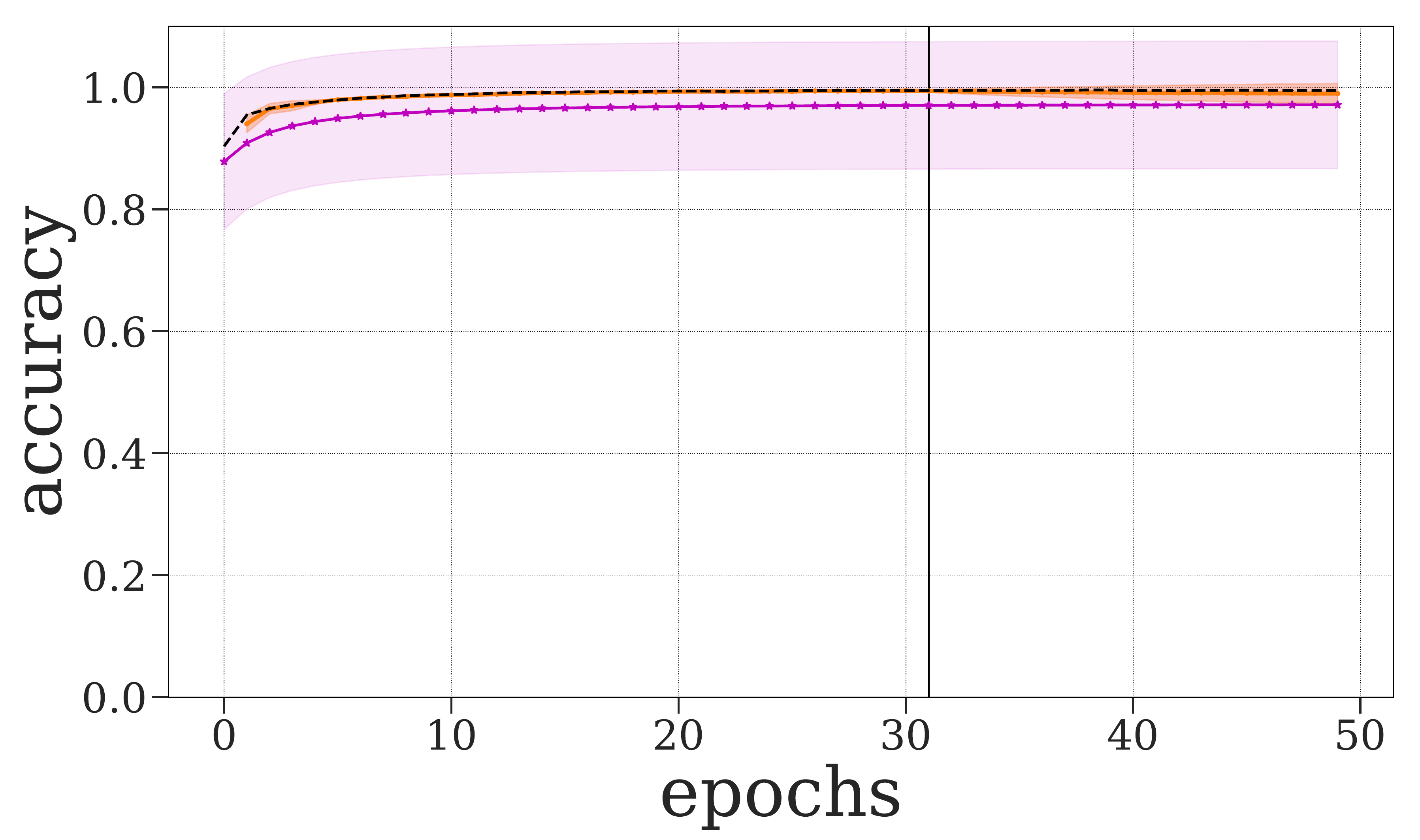}
  \caption{32 observed epochs}
  \label{fig:h}
\end{subfigure}
\caption{Qualitative comparison of the test predictions of VRNN and LCNet models for different numbers of observed epochs (the black vertical line) for one learning curve randomly sampled from the MNIST dataset. The shaded area corresponds to $1\sigma$.}
\vspace*{-0.5cm}
\label{fig:comparison_fig}
\end{figure}

To test the predictive strength of our learning curve model, we generated four different sets of learning curves of a feed forward neural network as training data.
For each dataset, we sampled 5000 hyperparameter configurations randomly from the configuration space described in Table~\ref{tab:table_hyper_mlps} in Section~\ref{datasets_details} in the Appendix and trained each configuration for 50 epochs with Adam~\citep{kingma-iclr14} on the datasets MNIST~\citep{lecun-isp01a}, Adult~\citep{kohavi1996scaling}, Higgs~\citep{baldi2014searching} and Vehicle~\citep{siebert-turing87} collected from OpenML~\citep{vanschoren-sigkdd13a}. The same learning curve datasets were also used in~\citep{falkner-icml18}. For our experiments we used $25 \%$ of each dataset as test data, and trained each method on the remaining part with full-length learning curves. Due to space constraints, we only show experiments on the MNIST benchmark (see the Appendix for the experiments with the other datasets).
%\frank{As I had asked during my last pass: is this not exactly the data from the LCNet paper? Saying so would make the argument much stronger; otherwise sceptical readers might think we generated the data in a way that makes our model look good.}

As baselines, we consider LCNet~\citep{klein-iclr17}, a random forest baseline (RF-B) as described by~\citet{klein-iclr17} and the last seen value (LSV) heuristic, that, despite its simplicity, is successfully used in Hyperband~\citep{li-iclr17}.
Note that, LSV does not provide uncertainty estimates, therefore we used it only to compare against mean predictions. 
While we found the random forest based methods to be robust against their own hyperparameters, we used BOHB~\citep{falkner-icml18} to optimize the hyperparameters of our model and LCNet on the MNIST dataset and then used the best found configuration for all experiments (see Section~\ref{sec:hyper_opt_supp} in the Appendix for more details).
%The considered non roll-out and roll-out probabilistic regression models enable for the prediction of both partially and/or completely unobserved learning curves of new configurations (this is achieved in the roll-out scenario by augmenting the dimensionality of the input sequence and setting the added elements of each input sequence to dummy values), since, in contrast to previously proposed approaches, i.e.~\citep{domhan-ijcai15}, they model performance across different hyperparameter settings. 

Even though LCNet and RF-B allow to predict for completely unobserved curves, only our models are able to correct their test predictions on the fly without the need for retraining after observing initial points from the true learning curve.
This property, illustrated in Figures~\ref{fig:mnist_unrolling},~\ref{fig:comparison_fig} and~\ref{fig:mnist_metrics_epoch_40}, (see Section~\ref{figures} in the Appendix for additional Figures), is fundamental for multi-fidelity hyperparameter optimization methods, such as Hyperband~\citep{li-iclr17} or BOHB~\citep{falkner-icml18}, where learning curves of different configurations are extended to different budgets.
%TODO : add details about figures
In particular, in Figures~\ref{fig:comparison_fig} and~\ref{fig:mnist_metrics_epoch_40}, the test performances of different methods for different numbers of observed points at test time are shown. All the models were trained on the MNIST benchmark for full-length learning curves. The roll-out methods take as input also the extra information provided at test time from the partially observed learning curves. Therefore their predictions do not remain unchanged, as those produced by LCNet and RF-B models, but improve with increasing number of observed epochs.
%\aaron{The experimental setup for Figure 4 and 5 is not explained: what is the test data? Explain why LCNET and RF-B are not decreasing with more observed data. What's the difference between 4 and 5.}
%\frank{This comment was not dealt with yet.}

This flexibility comes together with a higher quality of predictions, as shown in Figures~\ref{fig:comparison_fig} and~\ref{fig:lcnet_mnist_predicted_true} (see also the Tables in Section~\ref{tables} of the Appendix).
In addition, another benefit compared to the LCNet model is that our model does not rely on prior user knowledge of the learning curves shape through the use of the parametric basis functions. %\vspace*{-2cm}
%plot with mse and median LL on mnist for all different methods
\begin{figure}[t]
\setlength\belowcaptionskip{-0.8cm}
%\vspace*{-1cm}
    \centering
    \includegraphics[width=0.8\linewidth]{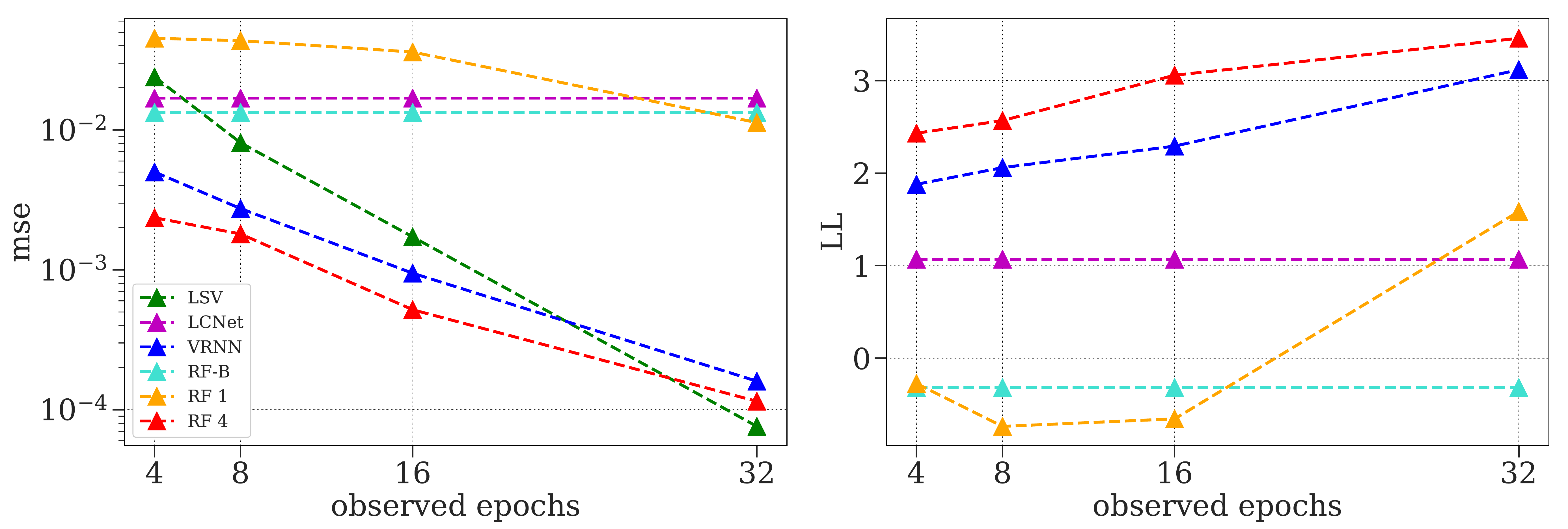}
    %\vspace*{-0.3cm}
    \caption{Assessment of the test predictive quality of the different models at target epoch 40 on the MNIST benchmark for different numbers of observed points from the learning curves. Note that RF-B and LCNet are not capable of adapting their predictions online without retraining, hence their constant error across epochs. Note that LSV does not provide a predictive variance.}
    \label{fig:mnist_metrics_epoch_40}
\vspace{0.9mm}
%\vspace{-0.cm}
\end{figure}
%comparing with LCNet
%TODO add also predicted true for LCNet
%In Figure~\ref{fig:comparison_fig}, we compare the predictions of VRNN and LCNet models. From this visual analysis it also emerges that our VRNN model yields to more flexible and accurate predictions (see also the Tables in Section~\ref{tables} in the appendix). 

As shown in Figure~\ref{fig:mnist_metrics_epoch_40}, the performance of roll-out models based on random forests degrades significantly with the reduction of the input size (see also Figures~\ref{fig:rfr_s4_rollout_4}--\ref{fig:rfr_s1_rollout_32},~\ref{fig:rfr_s2_predicted_true_ll} and~\ref{fig:rfr_s1_predicted_true_ll} in Appendix~\ref{figures}). 
This is due to their intrinsic inability of modeling sequence-type data, such as learning curves. 
Therefore, this class of methods might require an input size whose cost could realistically be non-negligible for scenarios such as learning curves generated by the training of state-of-the-art deep neural networks. 
%\begin{figure}[ht]
%  \begin{subfigure}[b]{0.5\textwidth}
%    \includegraphics[width=60mm,scale=0.6]{plot/vrnn/mnist_predicted_true.pdf}
%    \caption{VRNN, 4 observed epochs.}
%    \label{fig:mnist_predicted_true}
%  \end{subfigure}
  %
%  \begin{subfigure}[b]{0.5\textwidth}
%    \includegraphics[width=60mm,scale=0.6]{plot/vrnn/lcnet_predicted_true.pdf}
%    \caption{LCNet.}
%    \label{fig:lcnet_predicted_true}
%  \end{subfigure}
%  \caption{Qualitative assessment of the roll-out performances of VRNN and LCNet for different epochs on the MNIST benchmark. Each plot shows on the horizontal axis the true value and on the vertical axis the predicted values. Each point is colored based on its log-likelihood value. \textcolor{red}{ SF : Could the colormap be the same for both plots? That would make it more obvious that the VRRNs actually have equal log likelihoods! Plus the plots need to be bigger somehow...}}
%  \label{fig:lcnet_mnist_predicted_true}
%\end{figure}
\begin{figure}[ht]
\vspace*{-0.26cm}           
    \centering
    \includegraphics[height=2.6cm, width=15.3cm]{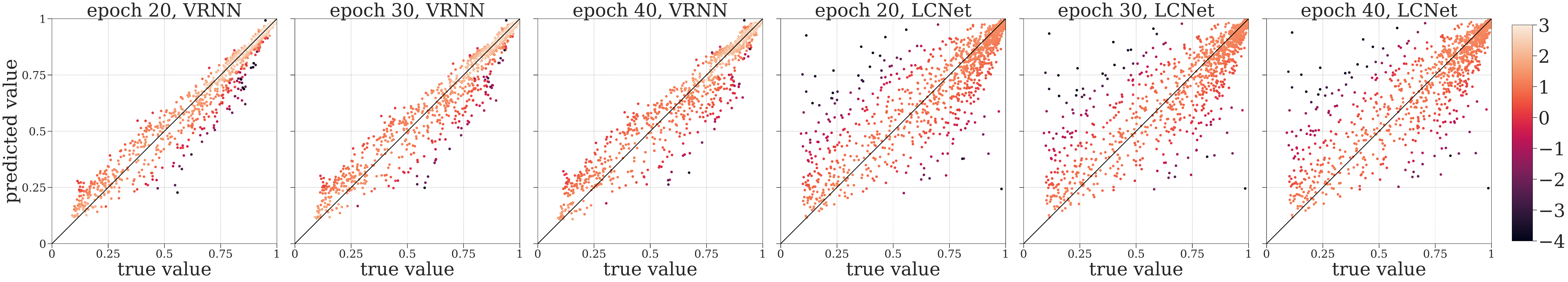}
  \caption{Qualitative assessment at different target epochs on the MNIST benchmark of the test roll-out performances of VRNN with 4 observed epochs and LCNet. Each plot shows on the horizontal axis the true values and on the vertical axis the predicted values. Each point is colored based on its log-likelihood value.}
  \label{fig:lcnet_mnist_predicted_true}
\end{figure} 
\subsection{Predictions for Unseen Datasets}\label{future_work}
\begin{figure}[t]
%\vspace*{-1cm}
    \begin{center}
        \includegraphics[width=.24\textwidth]{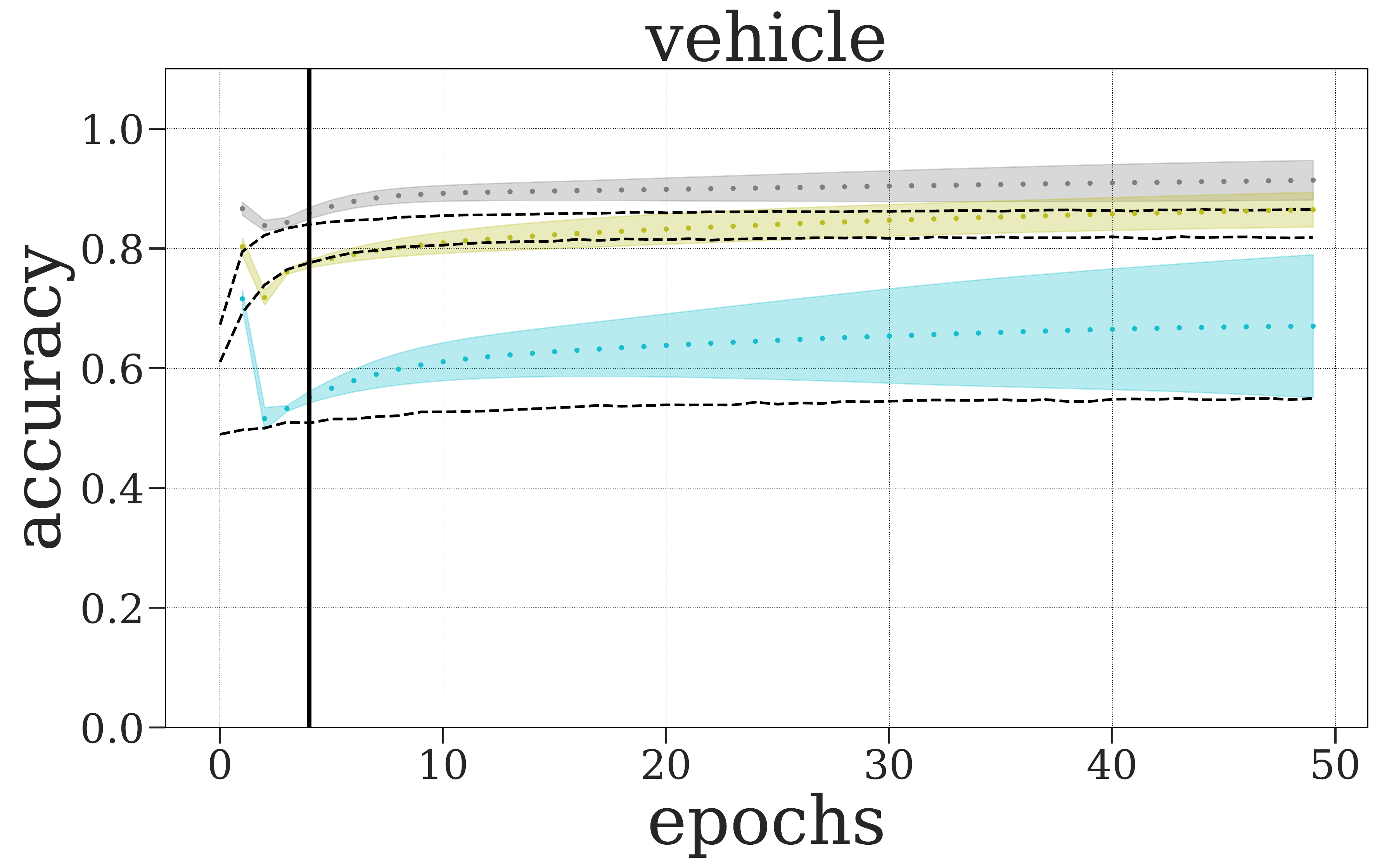}
        \includegraphics[width=.24\textwidth]{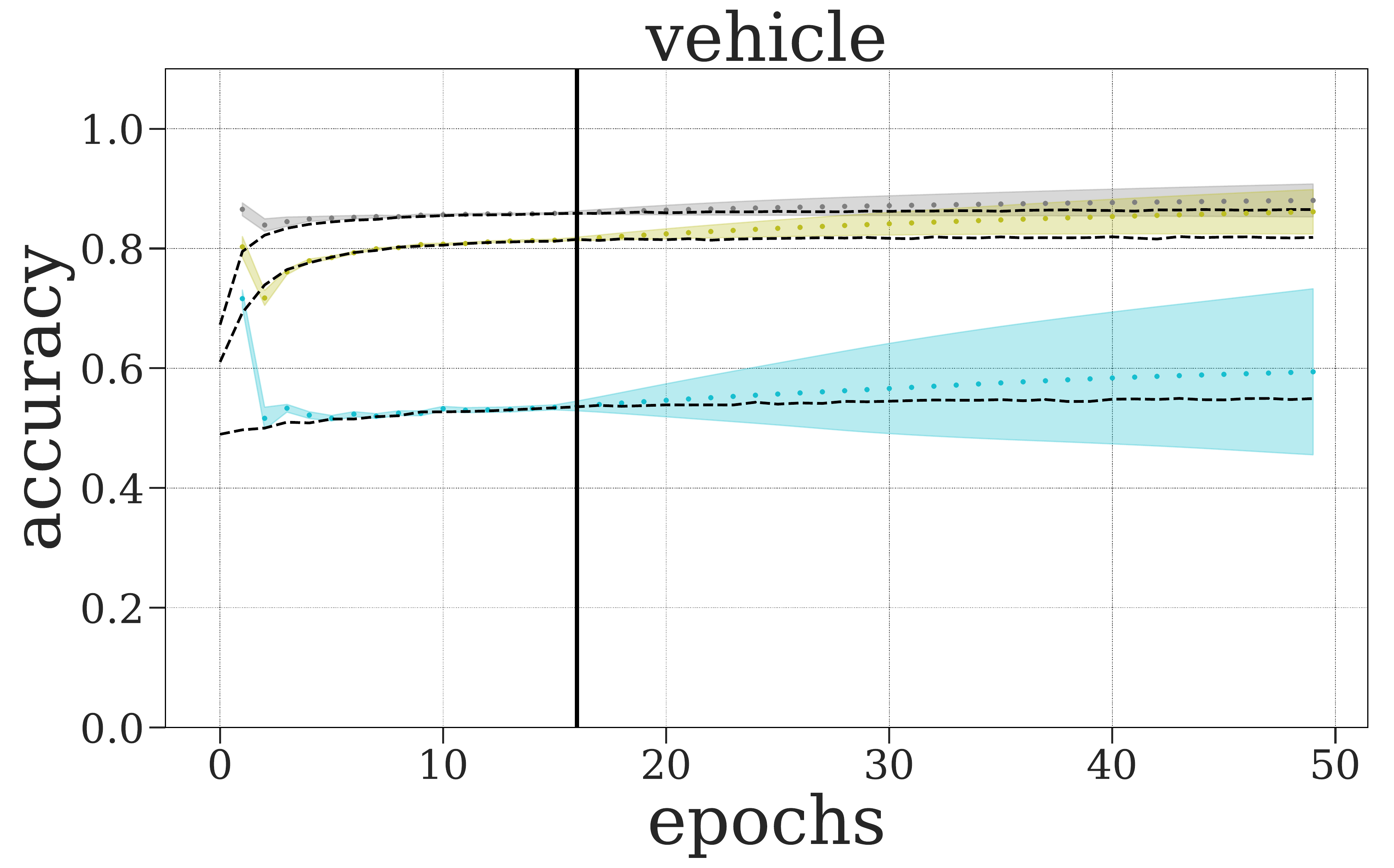}
        \includegraphics[width=.24\textwidth]{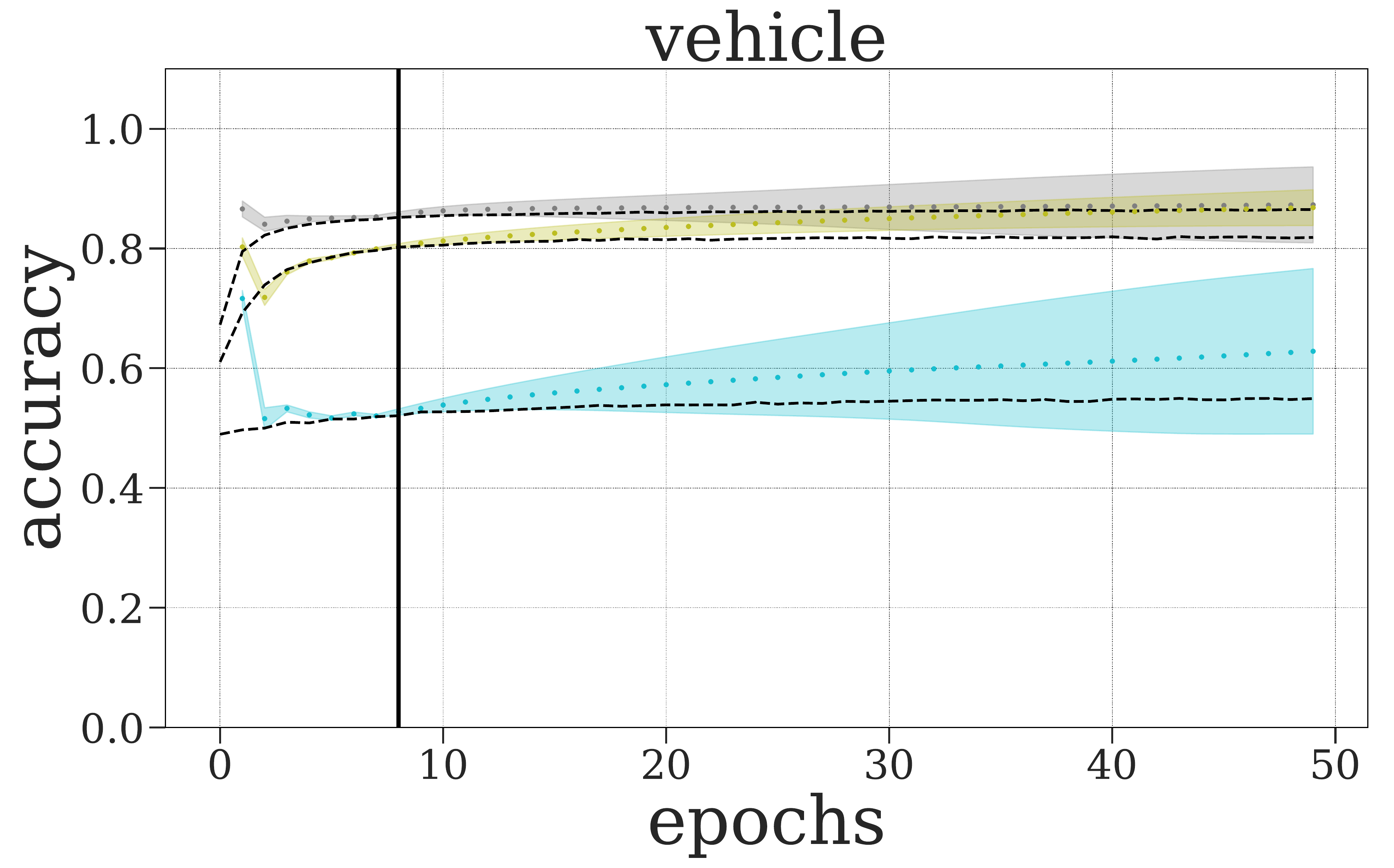}
        \includegraphics[width=.24\textwidth]{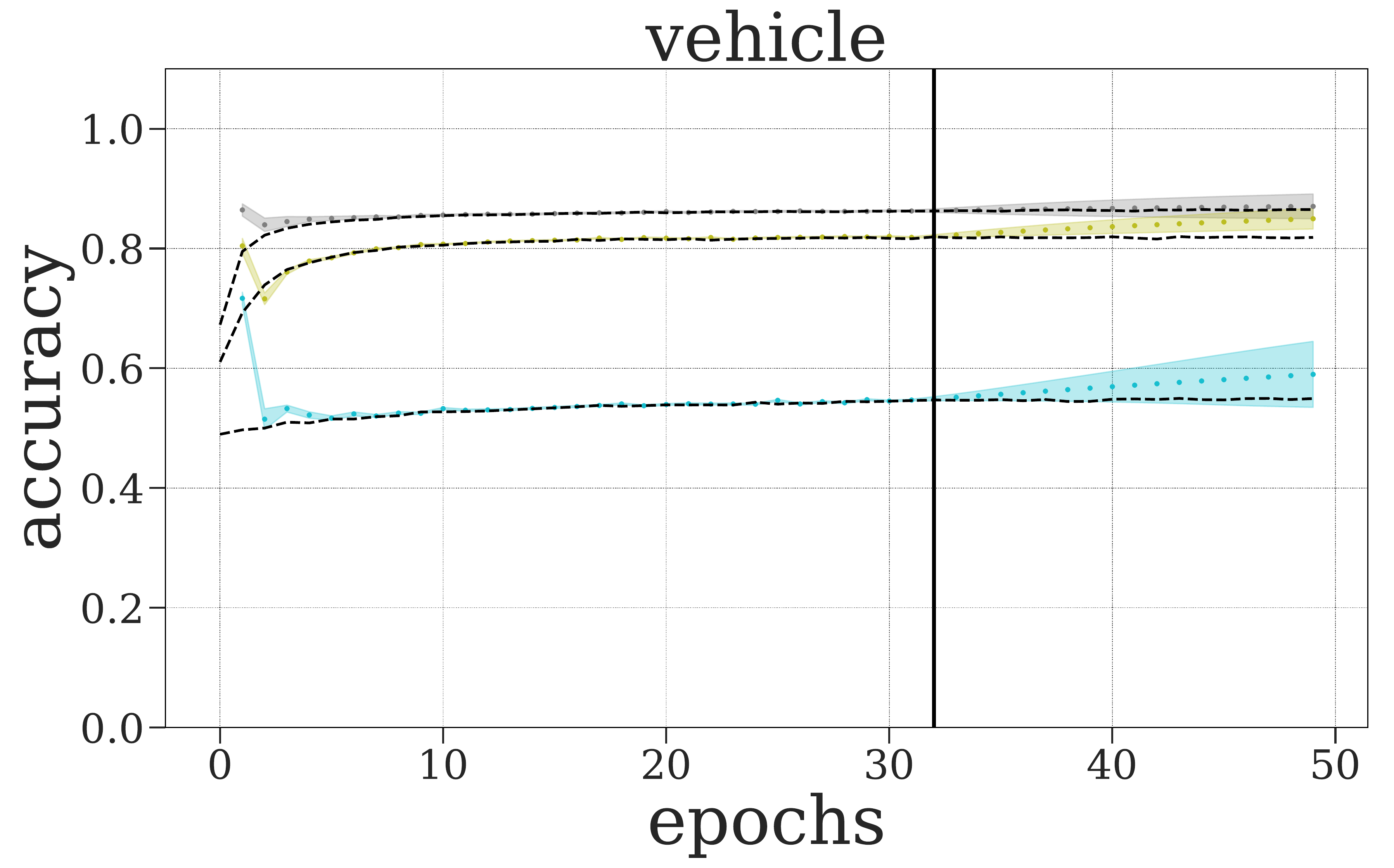}
    \end{center}
    \vspace*{-0.3cm}
    \caption{Qualitative assessment of VRNN predictions on the Vehicle benchmark when trained on MNIST for different numbers of observed epochs (black vertical line).}
    \label{fig:unseen_vehicle}
    \vspace*{-0.5cm}
\end{figure}

We now conduct preliminary experiments to study the performances of our roll-out models on unseen datasets, by training them on the MNIST benchmark and using these trained models to extrapolate partial learning curves on other datasets without retraining.
Even though the same configuration potentially leads to vastly different performances across different datasets (as also observed in our benchmarks), Figure~\ref{fig:unseen_vehicle} suggests that the VRNN model can adjust its predictions on the fly by starting its roll-outs with the initial learning curves observed on the new dataset. 
Based on these results (see also Figures 
\ref{fig:rfr_4_vehicle_unseen_dataset}--\ref{fig:unseen_higgs_predicted_true}
%Figures~\ref{fig:vrnn_higgs_unseen_dataset},~\ref{fig:vrnn_adult_unseen_dataset} and Figures~\ref{fig:rfr_4_vehicle_unseen_dataset},~\ref{fig:rfr_4_higgs_unseen_dataset},~\ref{fig:rfr_4_adult_unseen_dataset} for RFR 4 
in Appendix~\ref{figures}), we believe that our model is very promising for
a variety of meta-learning and transfer-learning extensions.

%In the wake of what has been explored in Section~\ref{hyperband_vrnn}, we also plan to investigate the transfer-learning direction. Namely, instead of keeping the pre-trained VRNN model fixed, the model's parameters are updated when new data points from selected configurations become available.
%Finally, we are also interested in using our model, with the meta-learning and/or transfer-learning extensions, as model inside hyperparameter optimization methods, such as BOHB.
%%We indeed expect that BOHB greatly benefits from the exploitation of the prediction power of the VRNN.
%\textcolor{red}{SF: Because this work is actually part of the Bosch workpackages, we should have filed an Invention Report for any patentable idea. I talked to Sina before Easter, and could convince here that if we simply apply known methods to known benchmarks that we don't need to go through the whole IP process. That being said, I think we should be cautious about details in the conclusion like "learn the meta-features that characterize different datasets". Maybe we could be a bit more vague. Unfortunately, Sina is on vacation right now, so I can't ask her.}
%In this section, we show preliminary experiments to hint at the potential strengths of our model in the meta-learning scenario. 

\section{Conclusion and Possible Use Cases of our Roll-Out Models}
We proposed new roll-out models for the learning curve prediction task, based on random forests and variational recurrent neural networks. 
%We then studied the prediction quality of our models, and compared them against non roll-out techniques from the literature on different datasets. 
These models offer more flexibility and better performances than previous state-of-the-art learning curve prediction methods from the literature. In addition, they show to be capable of adapting their predictions to unseen datasets. We now list some of the possible future extensions of this work in AutoML tasks:%In addition, we showed that the prediction quality of variational recurrent neural networks is more robust with respect to the input size than that of the roll-out random forests. 
%%In contrast to previous models, our roll-out models can also efficiently extrapolate learning curves on new datasets, without retraining.
%Finally, we conducted some preliminary experiments on unseen datasets, that revealed the great potential of roll-out models to adapt also to new datasets and therefore hint towards their use for hyperparameter optimization with meta-learning and transfer-learning extensions.
%\frank{I moved this part here after all, just not calling it Future Work.}
%%We believe the strength and flexibility of our model will be useful for many different AutoML tasks:
%We now list some of the possible future extensions of this work in AutoML tasks:
\vspace{-1.3mm}%{-1.9mm}
%we would like to study in future work:
\begin{itemize}
\addtolength\itemsep{-3.28mm}%{-3.38mm}
    \item Explicit dataset meta-features can be integrated and/or a latent task embedding learnt in order to  enable direct learning across datasets. %In absence of good meta-features, a latent task embedding vector for the training tasks can be learned. %for a new dataset, due to the fast rollouts our model offers, these can then be sampled using MCMC based on very few observed data points.
    \item Our model could be used to warmstart bandit-based hyperparameter optimizers, such as Hyperband and BOHB, and replace the individual models learnt by BOHB for each fidelity.
    \item Our model could also be used to directly make decisions about which learning curves to extend, akin to Freeze-Thaw Bayesian optimization~\citep{swersky-arxiv14}.%, but more efficiently and more versatile than their approximate Gaussian process model.
    \item High-quality and flexible uncertainty estimates of time-series predictions are important in scenarios based on the exploitation-exploration paradigm. This makes our model attractive also for reinforcement-learning applications.%The possibility of using high-quality and flexible uncertainty estimates of time-series predictions is very important in scenarios based on the exploitation-exploration paradigm, which also makes our model very attractive for reinforcement-learning applications.  
\end{itemize}
Due to this breadth of possible use cases, we expect our model to be useful in developing different types of new AutoML systems. To facilitate this, we make open-source code available for our model and experiments at \url{https://github.com/gmatilde/vdrnn}.

%as future work, we plan to investigate more explicit meta-learning and transfer-learning directions for roll-out models in order to strengthen their across datasets prediction quality. 
%Regarding the meta-learning scenario, we intend to extend our model in order to make it able to also learn meta-features that characterize different datasets. This can also be combined with transfer-learning techniques. Namely, instead of keeping the pre-trained roll-out model fixed, its parameters are updated when new data points become available, potentially enlarging the model's knowledge.

%With these extensions, roll-out models can potentially be used for automatic hyperparmater optimization, i.e. in combination with bandit-based solvers such as Hyperband and BOHB.

\newpage
\acks{This work has partly been supported by Robert Bosch GmbH, the state of Baden-W\"{u}rttemberg through bwHPC and the German Research Foundation (DFG) through grant no INST 39/963-1 FUGG,
and the European Research Council (ERC) under the European Union’s Horizon 2020 research and innovation programme under grant no.\ 716721.
}
\bibliography{vrnn}

% Appendix goes to a new page

\newpage

\appendix

\section{Details about the models}\label{appendix_model_graph}

\usetikzlibrary{shapes.misc, positioning}
\tikzset{decorate sep/.style 2 args=
{decorate,decoration={shape backgrounds,shape=circle,shape size=#1,shape sep=#2}}}

\begin{figure}[H]
    \centering
    \resizebox{.25\textwidth}{!}{
    \begin{tikzpicture}
      \node[neuron] (vt) {$\mathbf{y}$};
      \node[neuron,left=5em of vt] (kt) {$\boldsymbol{\theta}$};  
     \node[mlp_theta,above=1em of kt] (mlp_config) {$h_1$};%{$\textbf{MLP}_{\theta}$};
      \node[joint, above=1.4cm of vt] (joint_input) {$[\mathbf{h}_1\odot \mathbf{z}_1 , \mathbf{y}]$};
      \node[lstm,above =3.5cm of vt] (lstm_1) {$r_1$};%{\textbf{LSTM 1}};
     %\node[mlp,above=1.5em of lstm_1] (mlp_1) {\textbf{MLP 1}};
      \node[neuron,above=10em of mlp_config] (theta_2) {$\boldsymbol{\theta}$}; 
     \node[mlp_theta,above=1em of theta_2] (mlp_config2) {$h_2$};%{$\textbf{MLP}_{\theta}$};
     \node[joint, above=2.6cm of lstm_1] (joint_input2) {$[\mathbf{h}_2\odot \mathbf{z}_2 , \mathbf{\tilde{h}}_1]$};
      
     \node[lstm,above=5.5cm of lstm_1] (lstm_2) {$r_2$};%{\textbf{LSTM 2}};
     \node[mlp,above=1.5em of lstm_2]   (mlp_2) {$h_3$};%{\textbf{MLP}};
      \node[neuron,above=1.4em of mlp_2] (output) {$\tilde{\mathbf{y}}$};  
      \draw[conn] (vt) -- (joint_input);
      \draw[conn] (joint_input) -- (lstm_1);  
      \draw[conn] (kt) -- (mlp_config);
      \draw[conn] (mlp_config) |- (joint_input);
      \draw[conn] (lstm_1) -- (joint_input2);
      %\draw[decorate sep={2mm}{4mm},fill] (lstm_1) -- (mlp_1);
      \draw[conn] (lstm_2) -- (mlp_2);
      \draw[conn] (mlp_2) -- (output);
      \draw[conn] (theta_2) -- (mlp_config2);
      \draw[conn] (mlp_config2) |- (joint_input2);
      \draw[conn] (joint_input2) -- (lstm_2);
      %\draw[thick,->] (lstm_1.160) arc (1:264:5mm);
      \draw[rec] (lstm_1) to [looseness=5, out= 345, in=15]  (lstm_1);
      \draw[rec] (lstm_2) to [looseness=5, out= 345, in=15]  (lstm_2);
      %\draw[conn] (mlp_1) -- (joint_input2);
    \end{tikzpicture}
    }
    \caption{Folded schematic of the VRNN model for learning curve prediction. $h_1$, $h_2$ and $h_3$ are feedforward neural networks, $r_1$ and $r_2$ are LSTM blocks and $\mathbf{\tilde{h}}_1$ is the output of $r_1$. Given the learning curve $\mathbf{\tilde{y}^{(i)}}=\left(y_0, \dots, y_T  \right)$, in the graph $\mathbf{\tilde{y}}=\left(\tilde{y}_1, \dots, \tilde{y}_T  \right)$ is the vector of the predicted values, $\mathbf{y}=\left( y_0, y_1 \dots, y_{T-1} \right)$ is the input at training time and $\mathbf{y}=\left( y_0,\dots \tilde{y}_{M-1},\tilde{y}_{M}, \dots, \tilde{y}_{T-1} \right)$ is the input at evaluation time and where $M$ is the number of observed points, $\mathbf{z_1}$ and $\mathbf{z_2}$ are dropout masks sampled from a Bernoulli distribution and are kept fixed across time steps, and $\boldsymbol{\theta}$ is the configuration vector, which is fed into a feedforward neural network and then used to initialize the hidden states of the LSTM blocks. The bold arrows indicate the recurrence.}
    %\label{fig:vrnn_model}
\end{figure}
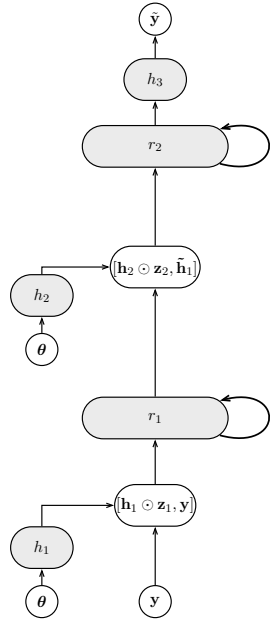

\newpage
\section{Datasets}\label{datasets_details}

Table~\ref{tab:table_hyper_mlps} reports the hyperparameters of the benchmarks from which the learning curves datasets described in Section~\ref{setting_subsection} are generated.

\begin{table}[h]
\vspace*{2cm}
\centering
\begin{tabular}{|l|c|c|}
\hline
Hyperparameter Name & Range & Log-Scale \\
\hline
initial learning rate & $[10^{-6}, 10^{-2}]$ & \checkmark
 \\
batch size & [16, 256] & \checkmark \\
average units per layer &$[2^4, 2^8]$ & \checkmark \\
final learning rate fraction &$[10^{-4}, 10^0]$  & \checkmark \\
shape parameter 1 & $[0,1]$ & \checkmark \\
dropout 0 & $[0.0,0.5]$ & $-$ \\
dropout 1 & $[0.0,0.5]$ & $-$ \\
number of layers & $[1,5]$ & $-$ \\
\hline
\end{tabular}
\vspace*{-0.08cm}
\caption{}
\label{tab:table_hyper_mlps}
\end{table}

\newpage

\section{Hyperparameter optimization}\label{sec:hyper_opt_supp}
 
In order to select the architecture for our model and the hyperparameters that control the training procedure (see Table~\ref{tab:table_hyper_vrnn} for a list of the hyperparameters), we used BOHB~\citep{falkner-icml18} as hyperparameter optimizer on the MNIST learning curves benchmark and the set-up described in Table~\ref{tab:bohb_setting}. % 10 minutes as maximum time budget. 
The configuration returned as incumbent was then used in all the VRNN experiments and showed good performances across all the considered datasets. The selected configuration is reported in Table~\ref{tab:config_vrnn}. 

In order to optimize the LCNet's hyperparameters, we also run BOHB with the same set-up (see Table~\ref{tab:bohb_setting}) on MNIST learning curves benchmark. Since numerical instability problems were occurring during the training procedure when the incumbent configuration returned by BOHB was applied, for the experiments with this model we then used the default configuration, which appeared to be more robust (Table~\ref{tab:table_hyper_LCNet} reports a list of the hyperparameters of LCNet together with their default values). 

As optimizers we used SGD with momentum and adaptive SGHMC for the VRNN and LCNet experiments, respectively. Regarding the experiments with the VRNN model, in order to speed up the training procedure, we also used a curriculum learning based technique~\citep{curriculum_learning_09} and linearly increased the length of the input sequence during training, starting from a selected number of initial observed epochs (this hyperparameter, dubbed  ``initial observed epochs'', was also optimized with BOHB and the selected value is reported in Table~\ref{tab:config_vrnn}).   

\begin{table}[H]
\centering
\begin{tabular}{|l|c|c|c|}
\hline
Hyperparameter Name & Value\\
\hline
$\eta$ & 2\\
number of iterations & 1000 \\
min time budget (min) & 2\\
max time budget (min) & 10 \\
\hline
\end{tabular}
\caption{Set-up of BOHB optimizer used to optimize VRNN and LCNet's hyperparameters.}
\label{tab:bohb_setting}
\end{table}

\begin{table}[H]
\centering
\begin{tabular}{|l|c|c|c|}
\hline
Hyperparameter Name & Range & Log-Scale & Type\\
\hline
initial learning rate & $[10^{-5}, 10^{-1}]$ & \checkmark & FLOAT
 \\
momentum & $[0, 0.99]$ & $-$ & FLOAT
 \\
final learning fraction & $[10^{-4}, 10^{0}]$ & \checkmark & FLOAT
 \\
batch size & [4, 128] & \checkmark & INTEGER \\
initial observed epochs &$[5, 50]$ & \checkmark & INTEGER \\
number of stacked LSTMs &$[1, 2]$  & $-$ & INTEGER \\
number of layers for final MLP & $[1,2]$ & $-$ & INTEGER \\
number of layers for config. MLP & $[1,2]$ & $-$  & INTEGER\\
number of units for LSTM & $[2^2, 2^7]$ & \checkmark & INTEGER \\
number of units for final MLP & $[2^2, 2^7]$ & \checkmark & INTEGER \\
number of units for config. MLP & $[2^2, 2^7]$ & \checkmark & INTEGER \\
learning rate scheduler & $[\text{cos, exp, const}]$ & $-$ & CATEGORICAL \\
\hline
\end{tabular}
\caption{Hyperparameter configuration space of the VRNN model described in Section~\ref{vrnn_model}.}
\label{tab:table_hyper_vrnn}
\end{table}

\begin{table}[H]
\centering
\begin{tabular}{|l|c|}
\hline
Hyperparameter Name & Selected Value\\
\hline
initial learning rate & $0.027$ \\
final learning fraction & $0.0008$
 \\
batch size & $22$ \\
initial observed epochs & $5$ \\
number of stacked LSTMs &$2$ \\
number of layers for final MLP & $1$ \\
number of layers for config. MLP & $1$ \\
number of units for LSTM & $6$ \\
number of units for final MLP & $103$ \\
number of units for config. MLP & $115$ \\
learning rate scheduler & $\text{cos}$ \\
\hline
\end{tabular}
\caption{Incumbent hyperparameter configuration selected by BOHB optimizer.}
\label{tab:config_vrnn}
\end{table}

%table for LCNet hyperparameters
\begin{table}[h]
\centering
\vspace{-0.5cm}
\begin{tabular}{|l|c|c|c|c|}
\hline
Hyperparameter Name & Range & Log-Scale & Type & Default\\
\hline
learning rate & $[10^{-5}, 10^{-1}]$ & \checkmark & FLOAT
 & 0.001\\
momentum & $[0, 0.99]$ & $-$ & FLOAT
 & 0.05 \\
batch size & $[4, 128]$ & \checkmark & INTEGER & 40 \\
\hline
\end{tabular}
\caption{Hyperparameter configuration space of LCNet.}
\label{tab:table_hyper_LCNet}
\end{table}

%section with some details about the optimization procedure for VRNN experiments (curriculum learning)   

\section{Tables with Mean Squared Error and Median Log-Likelihood}\label{tables}
\begin{figure}[H]
    \begin{center}
        \includegraphics[width=.45\textwidth]{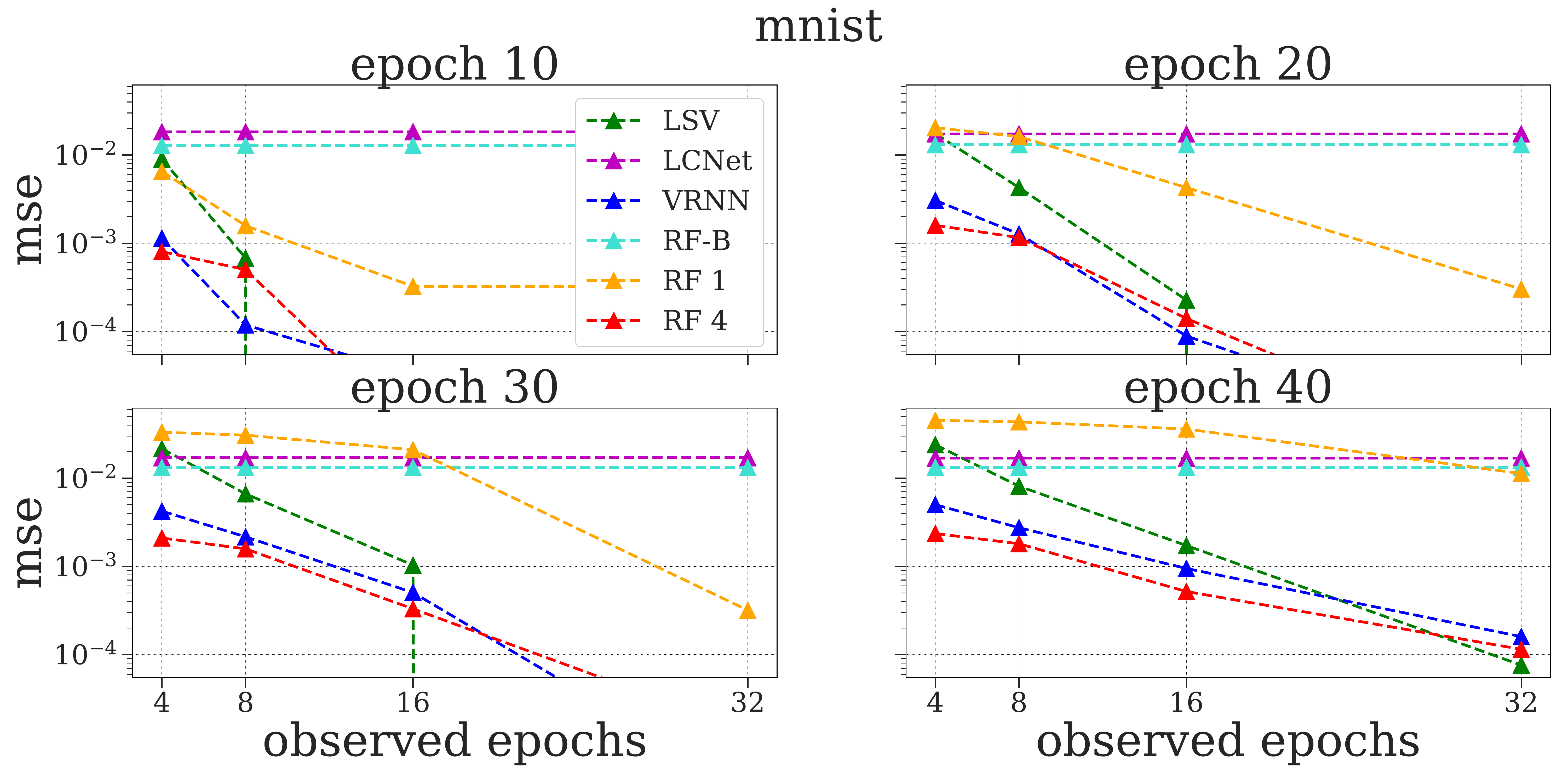}
        \includegraphics[width=.45\textwidth]{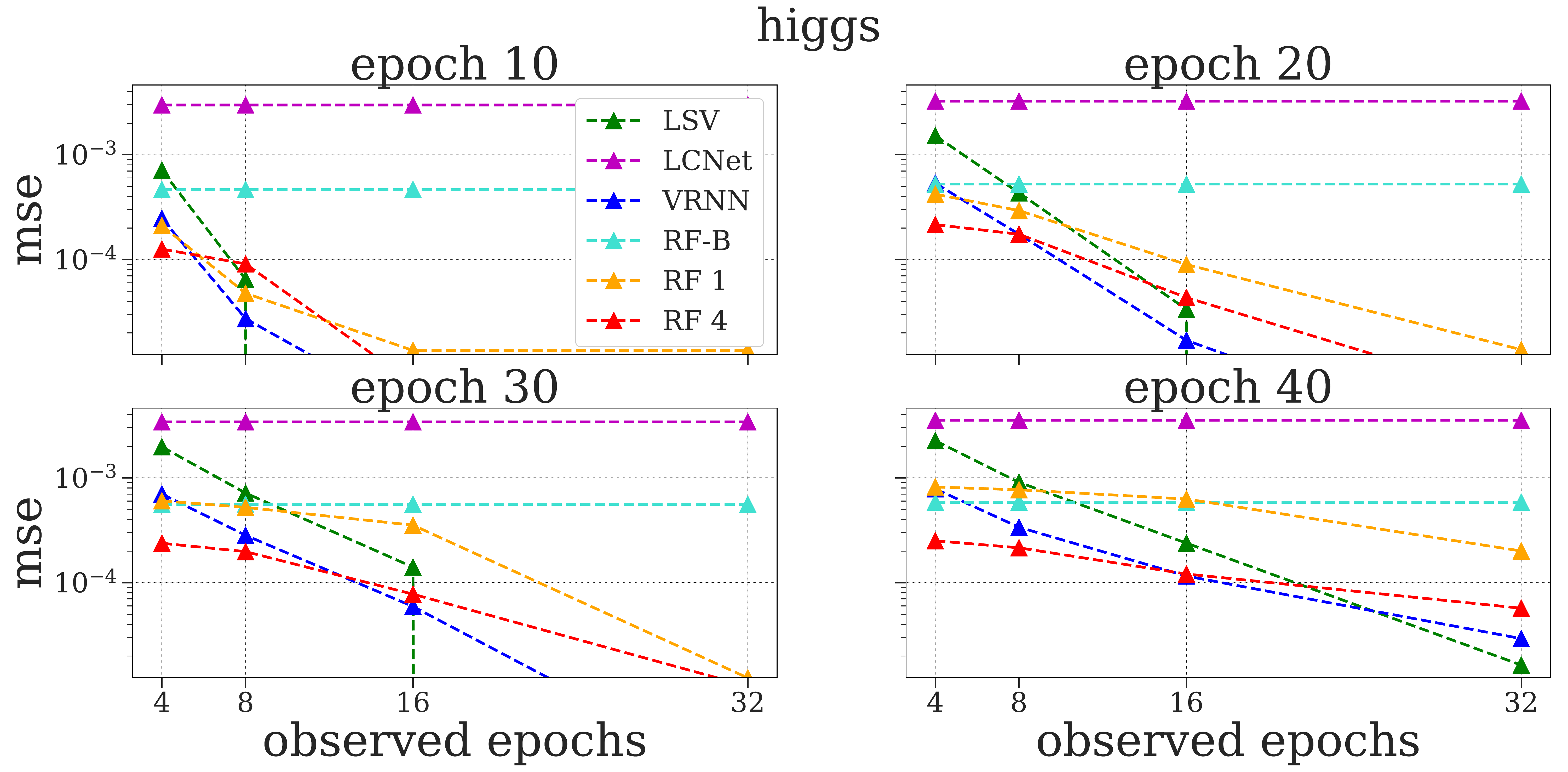}
        \includegraphics[width=.45\textwidth]{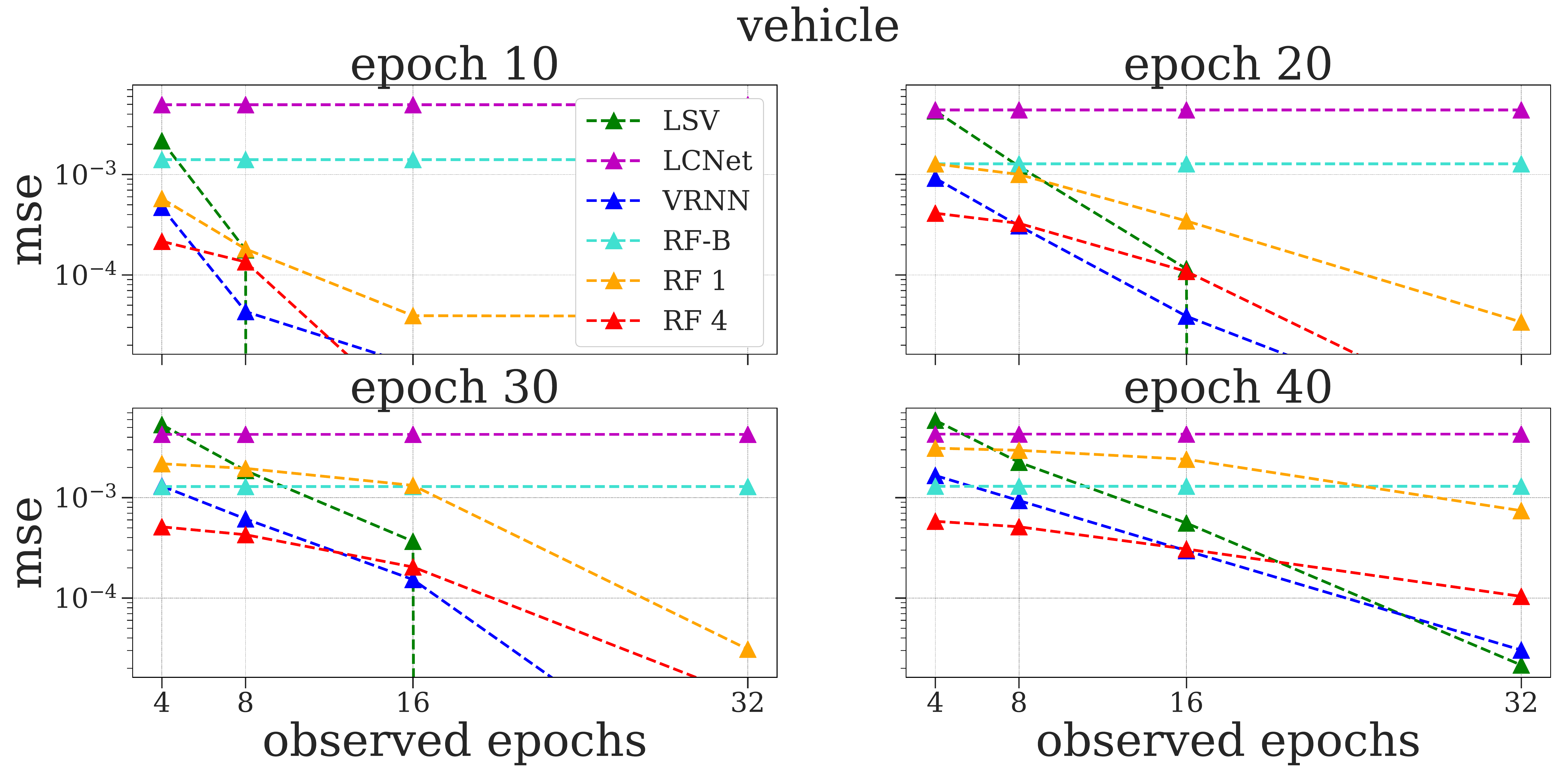}
        \includegraphics[width=.45\textwidth]{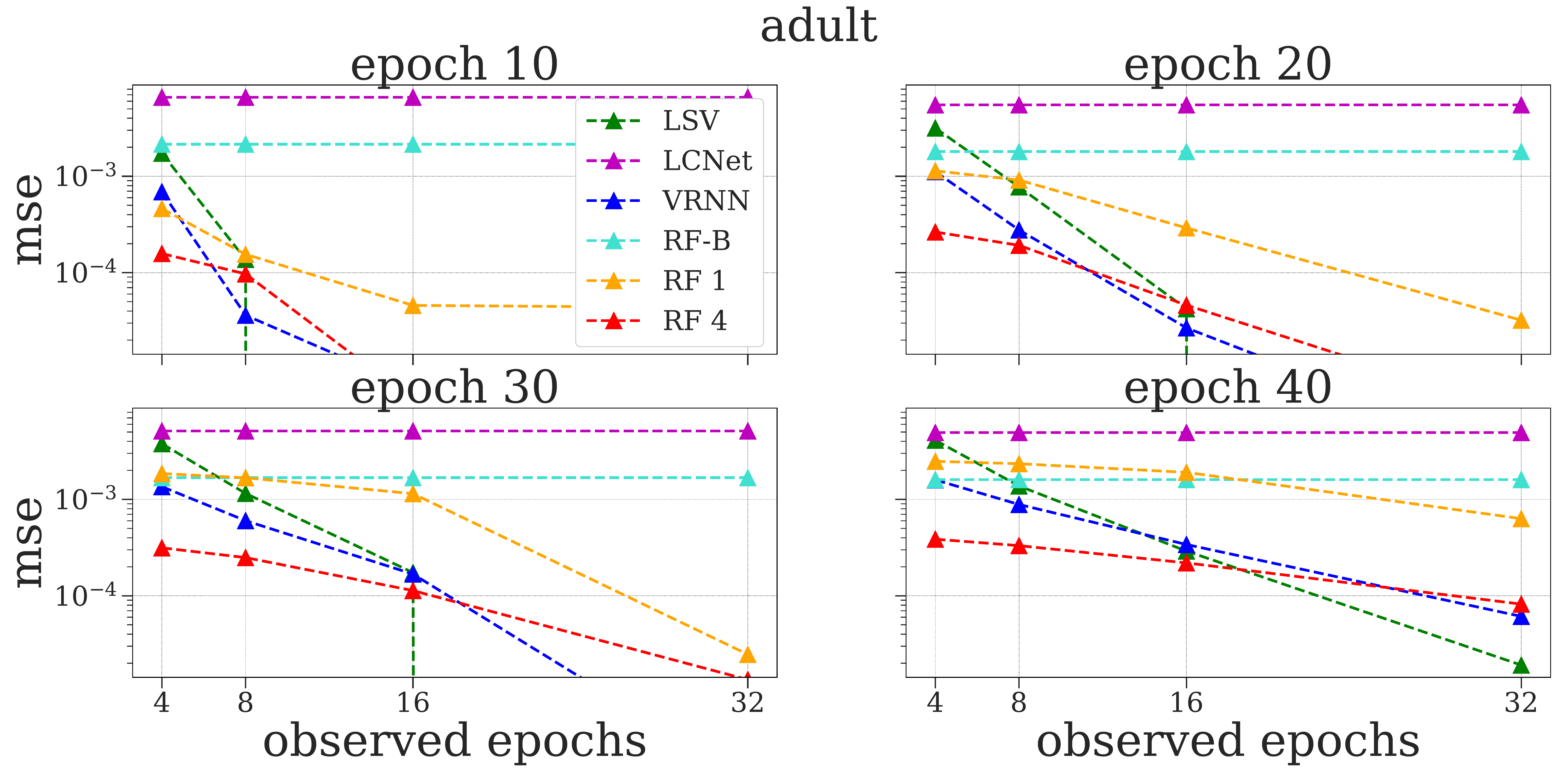}
    \end{center}
    \caption{The panels show how the mean squared error at different target epochs (y-axis) varies with the number of observed points from the true learning curve at evaluation time (x-axis) for different methods on the four considered benchmarks. }
    \label{fig:mse_observed_epochs}
    \end{figure}

\begin{figure}[H]
    \begin{center}
        \includegraphics[width=.45\textwidth]{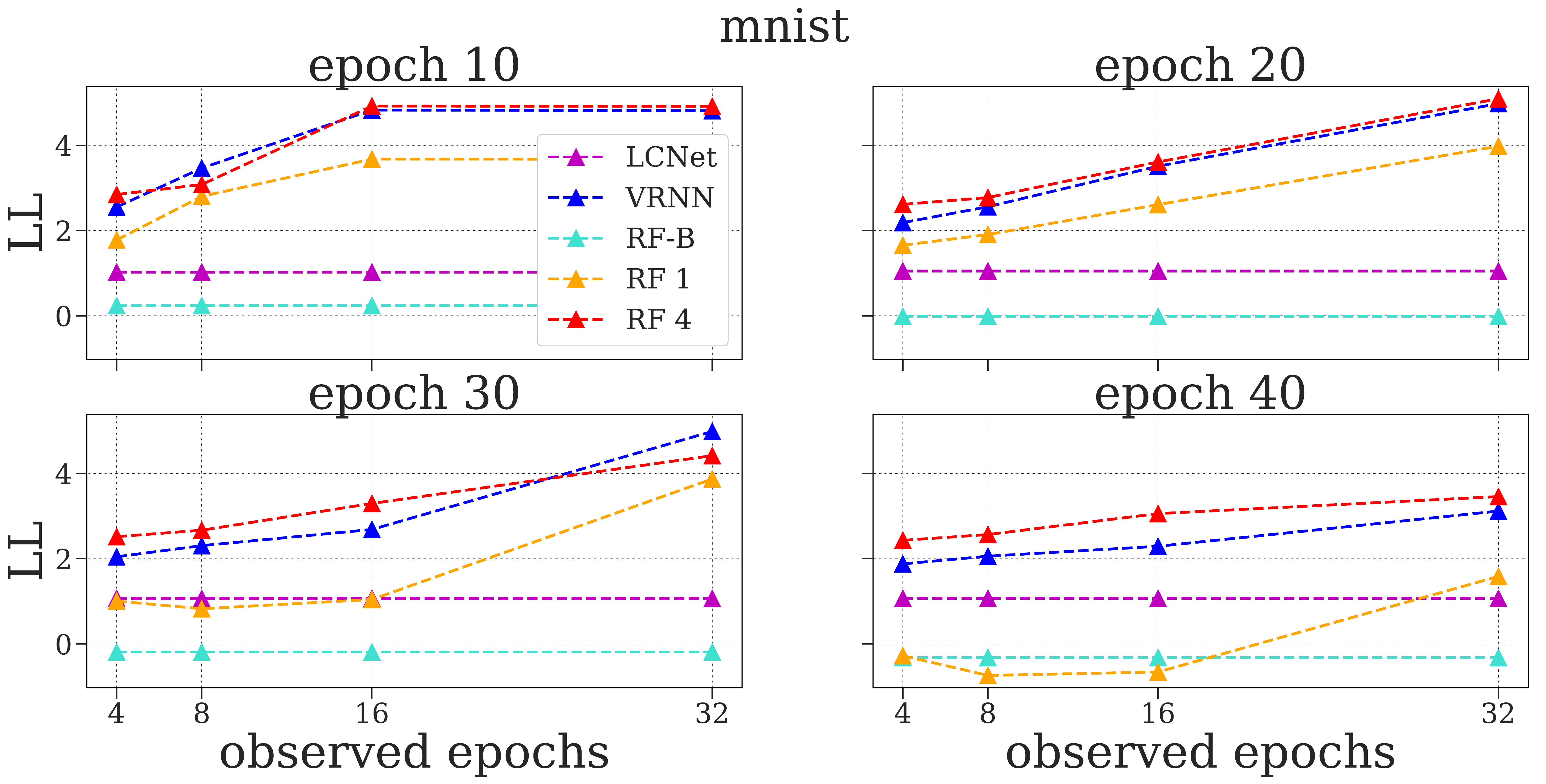}
        \includegraphics[width=.45\textwidth]{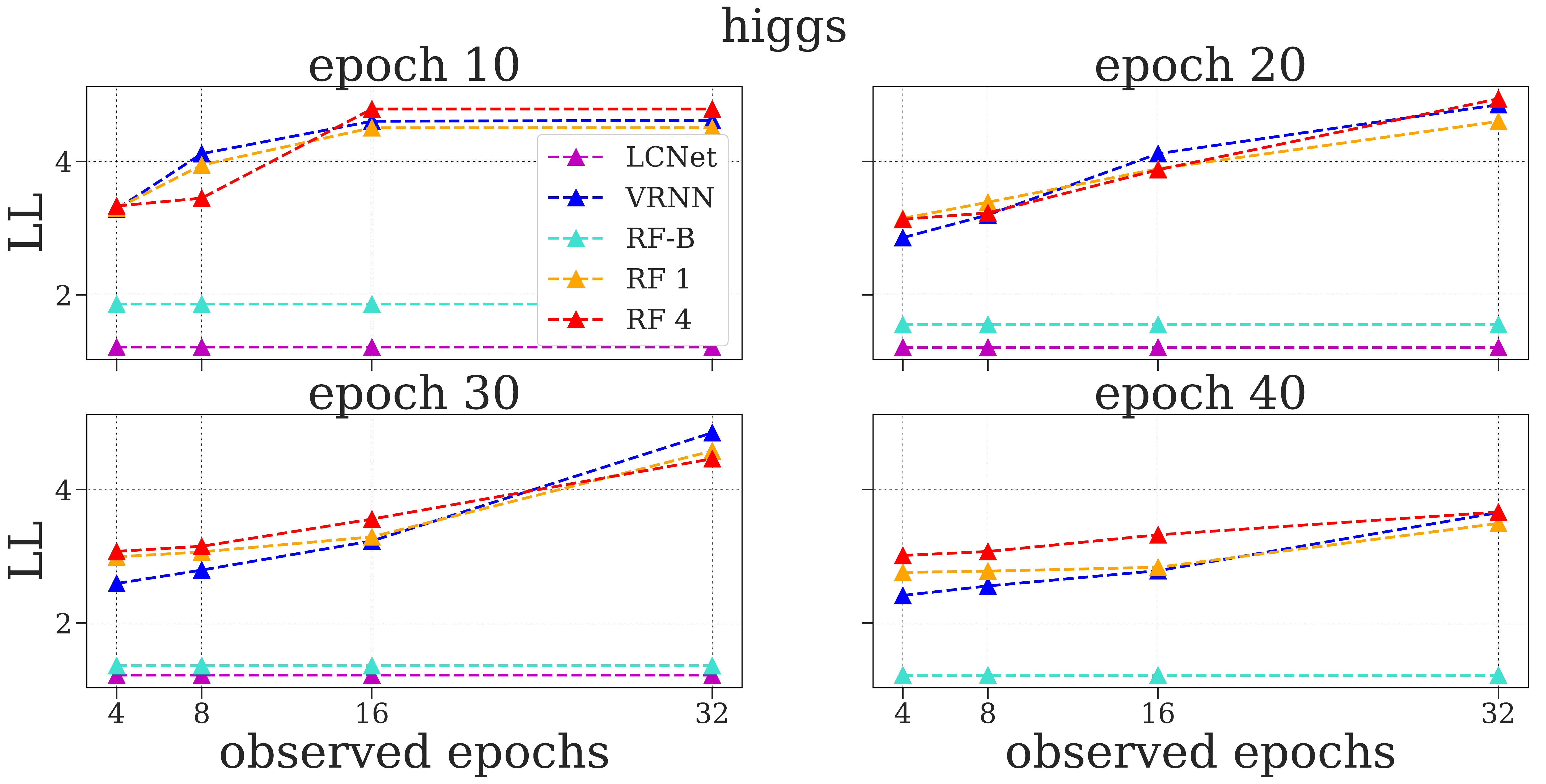}
        \includegraphics[width=.45\textwidth]{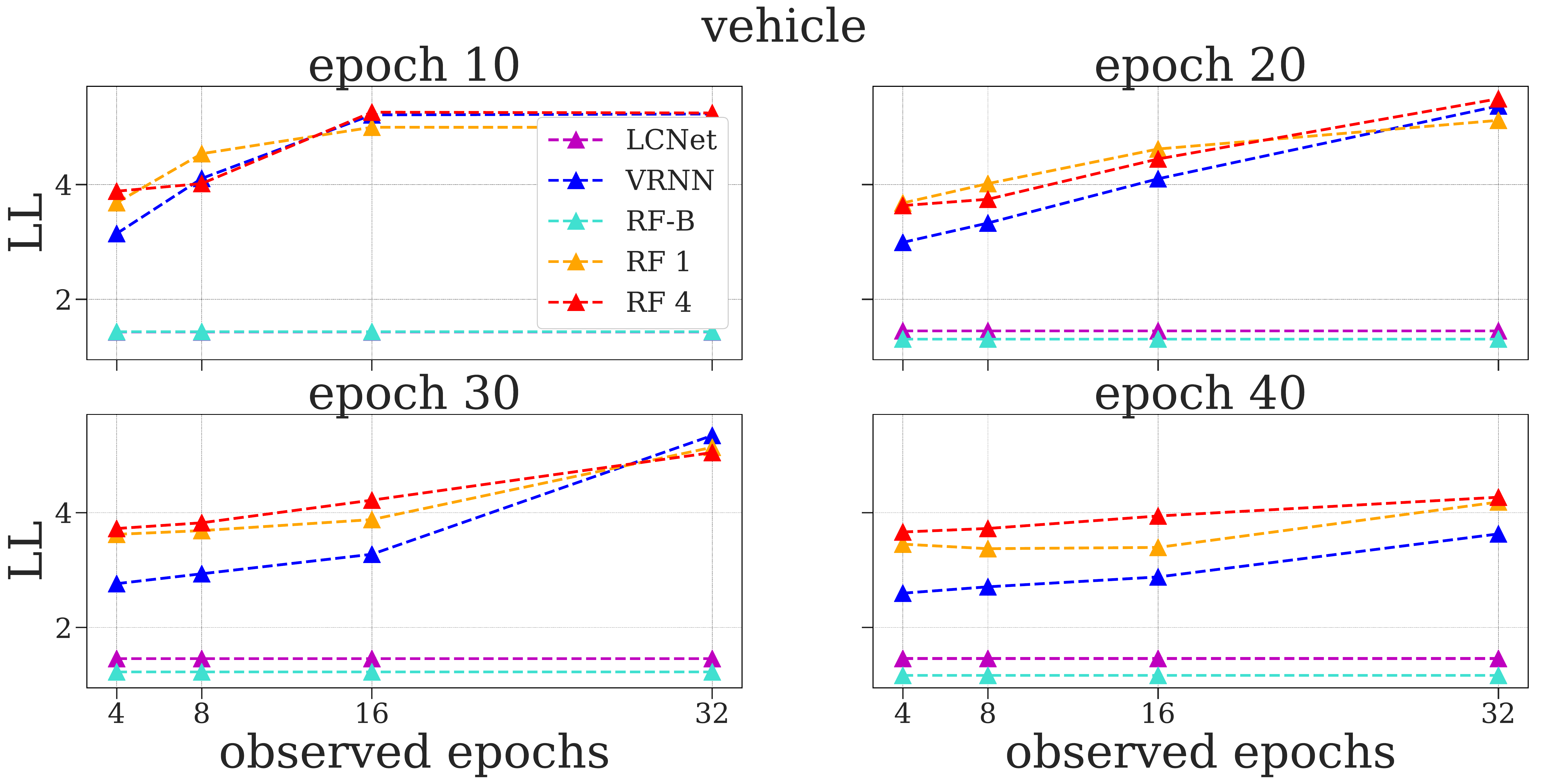}
        \includegraphics[width=.45\textwidth]{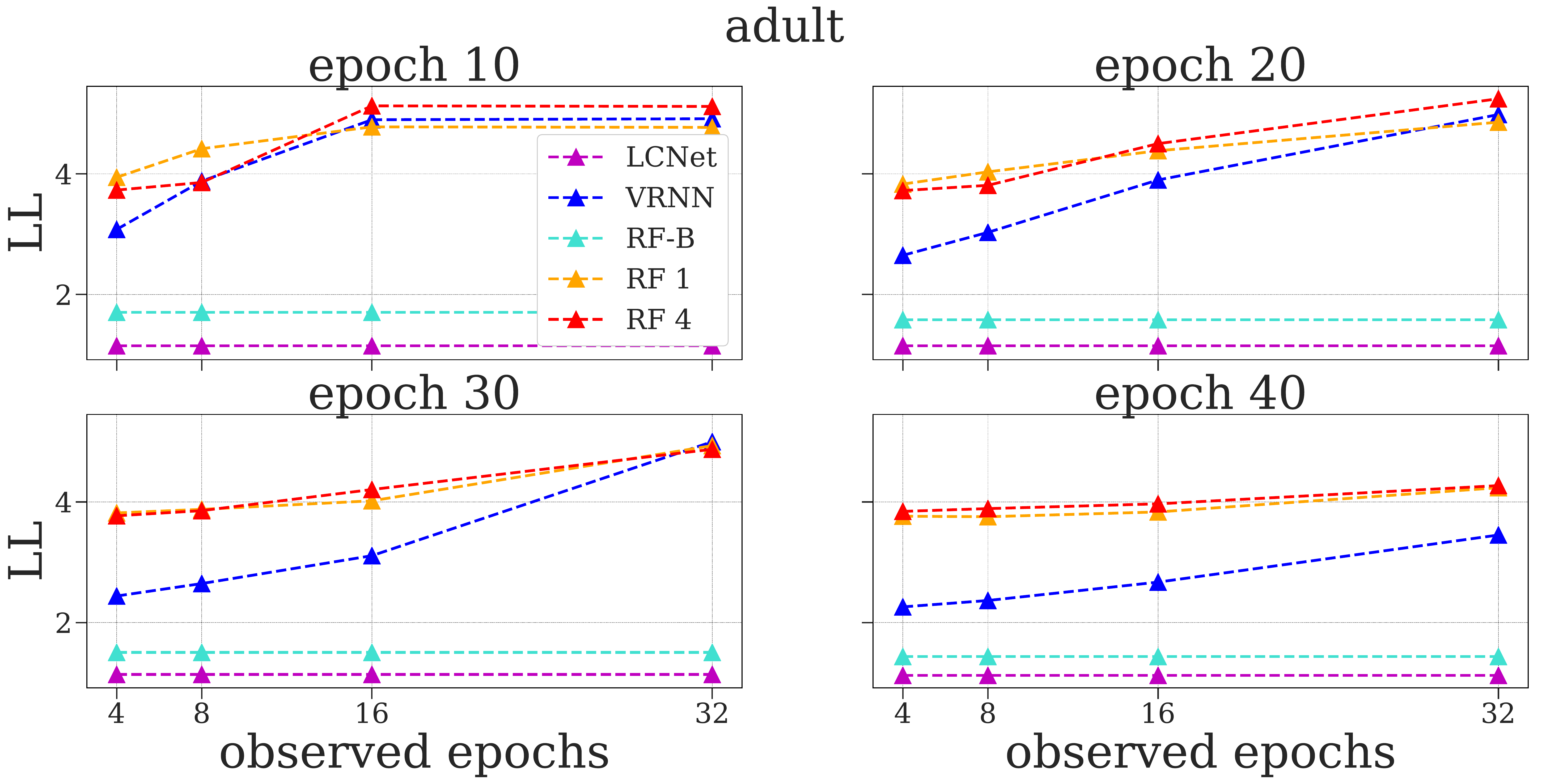}
    \end{center}
    \caption{The panels show how the median log-likelihood at different target epochs (y-axis) varies with the number of observed points from the true learning curve at evaluation time (x-axis) for different methods on the four considered benchmarks.}
    \label{fig:ll_observed_epochs}
    \end{figure}

Tables~\ref{tab:table_observed_4}--\ref{tab:table_observed_32} report for each method the achieved mean squared error and median log-likelihood averaged over all the epochs on the four considered datasets for different numbers of observed epochs at test time, respectively. The same metrics but across different epochs are also plotted in Figures~\ref{fig:mse_observed_epochs} and~\ref{fig:ll_observed_epochs}. We observe that our VRNN model and RF 4 consistently yield to better predictions. VRNN* and RF* 4 are used in the Tables to denote respectively VRNN and RF 4 models trained on the MNIST benchmark and then used to evaluate on the other unseen datasets. 
%table for 4 observed epochs

\begin{table}[H]
%\vspace{-0.2cm}
\scriptsize
\centering
\resizebox{\columnwidth}{!}{%
\begin{tabular}{|p{1.3cm}||@{\rule[-0.4cm]{0pt}{1cm}}*{8}{M{1.5cm} |}}% p{1.6cm}|p{0.82cm}|p{0.7cm}|p{0.82cm}|p{0.7cm}|p{0.82cm}|p{0.7cm}|p{0.82cm}|p{0.7cm}|   }
 \hline
 \multicolumn{1}{|c}{\textbf{epochs 4}} 
 &
 \multicolumn{2}{|c}{\textbf{mnist} }
 &
 \multicolumn{2}{|c}{\textbf{higgs} }
 &
 \multicolumn{2}{|c}{\textbf{adult} }
 &
 \multicolumn{2}{|c|}{\textbf{vehicle} }\\
 \hline
 \textbf{Methods}  &  \textbf{mse}   & \textbf{ll} & \textbf{mse}  & \textbf{ll}    & \textbf{mse} & \textbf{ll} & \textbf{mse} & \textbf{ll} \\
 \hline
 \textbf{VRNN}  & $3\mathrm{e}{-3}\pm 0.01 $ & $2.2\pm 1.1$ &  $6\mathrm{e}{-4} \pm 0.0$ & $2.75\pm 1.8$    & $1\mathrm{e}{-3} \pm 0.01$ &  $2.57\pm 0.9$ & $1\mathrm{e}{-3} \pm 0.0$& $2.9\pm 1.3$\\
 \hline
 \textbf{RF 1} & $0.03  \pm  0.07$    & $0.89 \pm 927$ &   $5\mathrm{e}{-4}  \pm  0.0$ & $2.9\pm 5.5$    &$1\mathrm{e}{-3} \pm  0.01$&   $3.7\pm 8.6$ & $2\mathrm{e}{-3} \pm  0.01$& $3.4\pm 16$\\
 \hline
 \textbf{RF 4} & $0.02  \pm  0.01$    & $2.7 \pm 4.5$ &   $2\mathrm{e}{-4} \pm  0.0$ & $3.2\pm 0.9$    &$3\mathrm{e}{-4} \pm  0.00$&   $3.8\pm 1.12$ & $4\mathrm{e}{-4} \pm  0.0$& $3.9\pm 1.12$\\
 \hline
 \textbf{LSV} & $0.02 \pm 0.03$    & $--$ &    $1\mathrm{e}{-3} \pm 0.0$ &  $--$   & $3\mathrm{e}{-3} \pm 0.01$ &    $--$& $4\mathrm{e}{-3} \pm 0.01 $ &  $--$\\
 \hline 
 \textbf{RF-B }   & $0.01 \pm 0.05$   &$0.01\pm 1.0$&   $5\mathrm{e}{-4} \pm 0.0$ & $1.6\pm 0.4$    &$2\mathrm{e}{-3} \pm 0.01$&   $1.6\pm 1.1$& $1\mathrm{e}{-3} \pm 0.0$& $ 1.3\pm 0.9$\\
 \hline 
 \textbf{LCNet} & $0.02\pm 0.04$   &$1.03 \pm 1.0$&   $3\mathrm{e}{-3} \pm 0.0$ & $1.21 \pm 0.16$    &$0.01 \pm 0.02$&   $1.12 \pm 0.74$& $5\mathrm{e}{-3} \pm 0.01$& $ 1.41 \pm 0.6$\\
 \hline 
\textbf{VRNN*}  & $3\mathrm{e}{-3}\pm 0.01 $ & $2.2\pm 1.1$ &  $0.01 \pm 0.01$ &  $-2.7\pm 3.4$ &$3\mathrm{e}{-3} \pm 0.0$ & $0.86\pm 1.94$& $3\mathrm{e}{-3} \pm 0.0$& $1.16\pm 2.3$\\
 \hline
 \textbf{RF* 4} & $0.02  \pm  0.01$    & $2.7 \pm 4.5$ & $1\mathrm{e}{-3}\pm 0.0 $   &  $2.16\pm 1.37$  &$1\mathrm{e}{-3}\pm 0.0 $  &$2.15\pm 1.3$   &$1\mathrm{e}{-3}\pm 0.0 $   & $2.37\pm1.15$ \\
 \hline
\end{tabular}}
\caption{Average total mean squared error and median log-likelihood achieved by the different models for 4 observed epochs at evaluation time.}
\label{tab:table_observed_4}
\end{table}
% table for 8 observed epochs
\begin{table}[H]
\small
\centering
\resizebox{\columnwidth}{!}{%
\begin{tabular}{|p{1.4cm}||@{\rule[-0.4cm]{0pt}{1cm}}*{8}{M{1.7cm} |}}% p{1.6cm}|p{0.82cm}|p{0.7cm}|p{0.82cm}|p{0.7cm}|p{0.82cm}|p{0.7cm}|p{0.82cm}|p{0.7cm}|   }
 \hline
 \multicolumn{1}{|c}{\textbf{epochs 8}} 
 &
 \multicolumn{2}{|c}{\textbf{mnist} }
 &
 \multicolumn{2}{|c}{\textbf{higgs} }
 &
 \multicolumn{2}{|c}{\textbf{adult} }
 &
 \multicolumn{2}{|c|}{\textbf{vehicle} }\\
 \hline
 \textbf{Methods}  &  \textbf{mse}   & \textbf{ll} & \textbf{mse}  & \textbf{ll}    & \textbf{mse} & \textbf{ll} & \textbf{mse} & \textbf{ll} \\
 \hline
 \textbf{VRNN}  & $2\mathrm{e}{-3} \pm 0.01 $ & $2.53\pm 2.1$ &  $2\mathrm{e}{-4} \pm 0.0$ & $3\pm1.5$    & $5\mathrm{e}{-4} \pm 0.0$ &  $2.9\pm 0.8$ & $5\mathrm{e}{-4} \pm 0.0$& $3.25 \pm 1.1$\\
 \hline
 \textbf{RF 1} & $0.02  \pm  0.06$    & $0.68\pm 243$ &   $4\mathrm{e}{-4}  \pm  0.0$ & $3.2\pm 5.5$    &$1\mathrm{e}{-3} \pm  0.01$&   $3.9 \pm 8.6$ & $1\mathrm{e}{-3} \pm  0.01$& $3.7\pm 13.5$\\
 \hline
 \textbf{RF 4} & $1\mathrm{e}{-3}  \pm  0.05$    & $2.88\pm 3.9$ &   $2\mathrm{e}{-4}  \pm  0.0$ & $3.32 \pm 0.73$    &$2\mathrm{e}{-4} \pm  0.0$&   $3.94\pm  1.1$ & $3\mathrm{e}{-4} \pm  0.1
 0$& $3.96\pm 1.1$\\
 \hline
 \textbf{LSV} & $5\mathrm{e}{-3} \pm 0.1$    & \vspace{0.05cm}$--$ &    $5\mathrm{e}{-4}\pm 0.0$ &  $--$   & $8\mathrm{e}{-4} \pm 0.01$ &    $--$& $1\mathrm{e}{-3} \pm 0.0 $ &  $--$\\
 \hline 
 \textbf{RF-B }   & $0.01 \pm 0.05$   &$0.01\pm 1.0$&   $5\mathrm{e}{-4} \pm 0.0$ & $1.6\pm 0.4$    &$2\mathrm{e}{-3} \pm 0.01$&   $1.6\pm 1.1$& $1\mathrm{e}{-3} \pm 0.0$& $ 1.3\pm 0.9$\\
 \hline 
 \textbf{LCNet} & $0.02\pm 0.04$   &$1.03 \pm 1.0$&   $3\mathrm{e}{-3} \pm 0.0$ & $1.21 \pm 0.16$    &$0.01 \pm 0.02$&   $1.12 \pm 0.74$& $5\mathrm{e}{-3} \pm 0.01$& $ 1.41 \pm 0.6$\\
 \hline 
\textbf{VRNN*}  & $2\mathrm{e}{-3} \pm 0.01 $ & $2.53\pm 2.1$ &  $3\mathrm{e}{-3} \pm 0.0$ & $-0.78\pm 2.4$    & $1\mathrm{e}{-3}\pm 0.0 $ & $1.75\pm 1.34$ & $1\mathrm{e}{-3} \pm 0.0$ &  $2.3\pm 2.1$ \\
 \hline
 \textbf{RF* 4} & $1\mathrm{e}{-3}  \pm  0.05$    & $2.88\pm 3.9$ &  $1\mathrm{e}{-3}\pm 0.0 $    &    $2.5\pm 0.6$ &  $7\mathrm{e}{-3}\pm 0.0 $ &$2.52 \pm 1.0$    & $1\mathrm{e}{-3}\pm 0.0 $  & $2.65 \pm 1.26$ \\
 \hline
\end{tabular}}
\caption{Average total mean squared error and median log-likelihood achieved by the different models for 8 observed epochs at evaluation time.}
\label{tab:table_observed_8}
\end{table}
%table for 16 observed epochs
\begin{table}[H]
\small
\centering
\resizebox{\columnwidth}{!}{%
\begin{tabular}{|p{1.4cm}||@{\rule[-0.4cm]{0pt}{1cm}}*{8}{M{1.7cm} |}}% p{1.6cm}|p{0.82cm}|p{0.7cm}|p{0.82cm}|p{0.7cm}|p{0.82cm}|p{0.7cm}|p{0.82cm}|p{0.7cm}|   }
 \hline
 \multicolumn{1}{|c}{\textbf{epochs 16}} 
 &
 \multicolumn{2}{|c}{\textbf{mnist} }
 &
 \multicolumn{2}{|c}{\textbf{higgs} }
 &
 \multicolumn{2}{|c}{\textbf{adult} }
 &
 \multicolumn{2}{|c|}{\textbf{vehicle} }\\
 \hline
 \textbf{Methods}  &  \textbf{mse}   & \textbf{ll} & \textbf{mse}  & \textbf{ll}    & \textbf{mse} & \textbf{ll} & \textbf{mse} & \textbf{ll} \\
 \hline
 \textbf{VRNN}  & $5\mathrm{e}{-4} \pm 0.0 $ & $3\pm 1.1$ &  $7\mathrm{e}{-5} \pm 0.0$ & $3.3\pm 0.9$    & $2\mathrm{e}{-4} \pm 0.0$ &  $3.38\pm 0.7$ & $2\mathrm{e}{-4} \pm 0.0$& $3.7\pm1.3$\\
 \hline
 \textbf{RF 1} & $0.02  \pm  0.04$    & $1.13\pm 89.6$ &   $3\mathrm{e}{-4}  \pm  0.0$ & $3.48\pm 2.9$    &$1\mathrm{e}{-3} \pm  0.0$&   $4.1\pm 5.6$ & $1\mathrm{e}{-3} \pm  0.0$& $3.99\pm 7.8$\\
 \hline
 \textbf{RF 4} & $2\mathrm{e}{-4}  \pm  0.0$    & $3.64\pm 1.2$ &   $6\mathrm{e}{-5}  \pm  0.0$ & $3.81\pm 0.7$    &$9\mathrm{e}{-5} \pm  0.0$&   $4.4\pm 0.8$ & $2\mathrm{e}{-4} \pm  0.0$& $4.47\pm 1.0$\\
 \hline
 \textbf{LSV} & $8\mathrm{e}{-4}\pm 0.0$    & $--$ &    $1\mathrm{e}{-4} \pm 0.0$ &  $--$   & $1\mathrm{e}{-4} \pm 0.0$ &    $--$& $3\mathrm{e}{-4} \pm 0.0 $ &  $--$\\
 \hline 
 \textbf{RF-B}   & $0.01 \pm 0.05$   &$0.01\pm 1.0$&   $5\mathrm{e}{-4} \pm 0.0$ & $1.6\pm 0.4$    &$2\mathrm{e}{-3} \pm 0.01$&   $1.6\pm 1.1$& $1\mathrm{e}{-3} \pm 0.0$& $ 1.3\pm 0.9$\\
 \hline 
 \textbf{LCNet} & $0.02\pm 0.04$   &$1.03 \pm 1.0$&   $3\mathrm{e}{-3} \pm 0.0$ & $1.21 \pm 0.16$    &$0.01 \pm 0.02$&   $1.12 \pm 0.74$& $5\mathrm{e}{-3} \pm 0.01$& $ 1.41 \pm 0.6$\\
 \hline 
\textbf{VRNN*}  & $5\mathrm{e}{-4} \pm 0.0 $ & $3\pm 1.1$ & $2\mathrm{e}{-3} \pm 0.0$ &  $1\mathrm{e}{-3}\pm 2.35$ &  $7\mathrm{e}{-4} \pm 0.0$ & $2.35\pm 1.41$   &$1\mathrm{e}{-3} \pm 0.0$& $2.89\pm 2.1$\\
 \hline
 \textbf{RF* 4} & $2\mathrm{e}{-4}  \pm  0.0$    & $3.64\pm 1.2$ &   $1\mathrm{e}{-3}\pm 0.0$  & $3.1\pm 0.6$    & $3\mathrm{e}{-4}\pm 0.0$  &$3.2\pm 0.5$    & $3\mathrm{e}{-4}\pm 0.0$  &  $3.4\pm 1.2$ \\
 \hline
\end{tabular}}
\caption{Average total mean squared error and median log-likelihood achieved by the different models for 16 observed epochs at evaluation time.}
\label{tab:table_observed_16}
\end{table}

%table for 32 observed epochs

\begin{table}[H]
\small
\centering
\resizebox{\columnwidth}{!}{%
\begin{tabular}{|p{1.4cm}||@{\rule[-0.4cm]{0pt}{1cm}}*{8}{M{1.7cm} |}}% p{1.6cm}|p{0.82cm}|p{0.7cm}|p{0.82cm}|p{0.7cm}|p{0.82cm}|p{0.7cm}|p{0.82cm}|p{0.7cm}|   }
 \hline
 \multicolumn{1}{|c}{\textbf{epochs 32}} 
 &
 \multicolumn{2}{|c}{\textbf{mnist} }
 &
 \multicolumn{2}{|c}{\textbf{higgs} }
 &
 \multicolumn{2}{|c}{\textbf{adult} }
 &
 \multicolumn{2}{|c|}{\textbf{vehicle} }\\
 \hline
 \textbf{Methods}  &  \textbf{mse}   & \textbf{ll} & \textbf{mse}  & \textbf{ll}    & \textbf{mse} & \textbf{ll} & \textbf{mse} & \textbf{ll} \\
 \hline
 \textbf{VRNN}  & $1\mathrm{e}{-4}\pm 0.0 $ & $3.7\pm 1.2$ &  $2\mathrm{e}{-5} \pm 0.0$ & $3.81\pm 1.5$    & $6\mathrm{e}{-5} \pm 0.0$ &  $4.04\pm 0.8$ & $4\mathrm{e}{-5} \pm 0.0$& $4.49\pm 1.6$\\
 \hline
 \textbf{RF 1} & $5\mathrm{e}{-3}  \pm  0.01$    & $2.62\pm 3.7$ &   $1\mathrm{e}{-4}  \pm  0.0$ & $3.97\pm 0.9$    &$3\mathrm{e}{-4} \pm  0.0$&   $4.43\pm 1.8$ & $4\mathrm{e}{-4} \pm  0.0$& $4.64\pm 2.5$\\
 \hline
 \textbf{RF 4} & $4\mathrm{e}{-5}  \pm  0.0$    & $4.31 \pm 0.8$ &   $2\mathrm{e}{-5}  \pm  0.0$ & $4.31\pm 0.3$    &$3\mathrm{e}{-5} \pm  0.0$&   $4.8\pm 0.6$ & $4\mathrm{e}{-5} \pm  0.0$& $5.0 \pm 1.3$\\
 \hline
 \textbf{LSV} & $3\mathrm{e}{-5} \pm 0.0$    & $--$ &    $7\mathrm{e}{-6} \pm 0.0$ &  $--$   & $8\mathrm{e}{-6} \pm 0.0$ &    $--$& $9\mathrm{e}{-6} \pm 0.0 $ &  $--$\\
 \hline 
 \textbf{RF-B }   & $0.01 \pm 0.05$   &$0.01\pm 1.0$&   $5\mathrm{e}{-4} \pm 0.0$ & $1.6\pm 0.4$    &$2\mathrm{e}{-3} \pm 0.01$&   $1.6\pm 1.1$& $1\mathrm{e}{-3} \pm 0.0$& $ 1.3\pm 0.9$\\
 \hline 
 \textbf{LCNet} & $0.02\pm 0.04$   &$1.03 \pm 1.0$&   $3\mathrm{e}{-3} \pm 0.0$ & $1.21 \pm 0.16$    &$0.01 \pm 0.02$&   $1.12 \pm 0.74$& $5\mathrm{e}{-3} \pm 0.01$& $ 1.41 \pm 0.6$\\
 \hline 
\textbf{VRNN*} & $1\mathrm{e}{-4}\pm 0.0 $ & $3.7\pm 1.2$ &  $1\mathrm{e}{-3} \pm 0.0$ &  $0.54\pm 2.4$ &  $4\mathrm{e}{-4} \pm 0.0$ & $3.123\pm 1.63$    & $5\mathrm{e}{-4} \pm 0.0$& $3.75\pm 2.9$\\
 \hline
 \textbf{RF* 4}& $4\mathrm{e}{-5}  \pm  0.0$    & $4.31 \pm 0.8$ &  $1\mathrm{e}{-4}\pm 0.0 $ &    $3.8\pm 0.3$& $7\mathrm{e}{-5}\pm 0.0 $ & $4.03\pm 0.36$   &  $6\mathrm{e}{-5}\pm 0.0 $ & $4.25\pm 0.6$\\
 \hline
\end{tabular}}
\caption{Average total mean squared error and median log-likelihood achieved by the different models for 32 observed epochs at evaluation time.}
\label{tab:table_observed_32}
\end{table}

\section{Additional Plots}\label{figures}

%VRNN unrolling the learning curves when different epochs are  observed
\begin{figure}[H]
    \begin{center}
        \includegraphics[width=.8\textwidth]{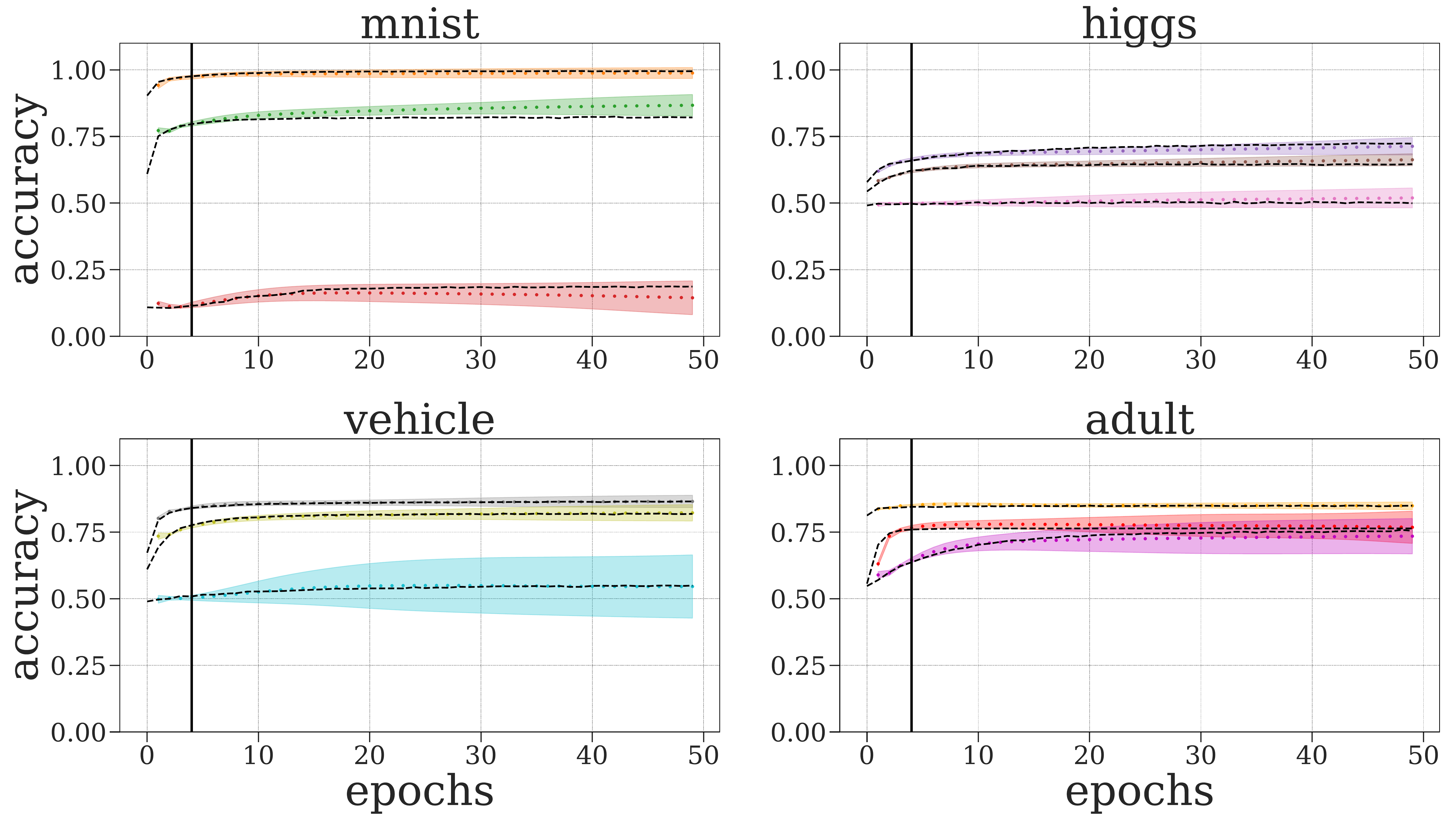}
    \end{center}
    \caption{Qualitative assessment of the test roll-out performances of VRNN for 4 observed epochs (the black vertical line). Different colors of learning curves stand for different configurations, while the black dashed lines represent the true learning curves.}
    \label{fig:vrnn_rollout_4}
\end{figure}
\begin{figure}[H]
    \begin{center}
        \includegraphics[width=.8\textwidth]{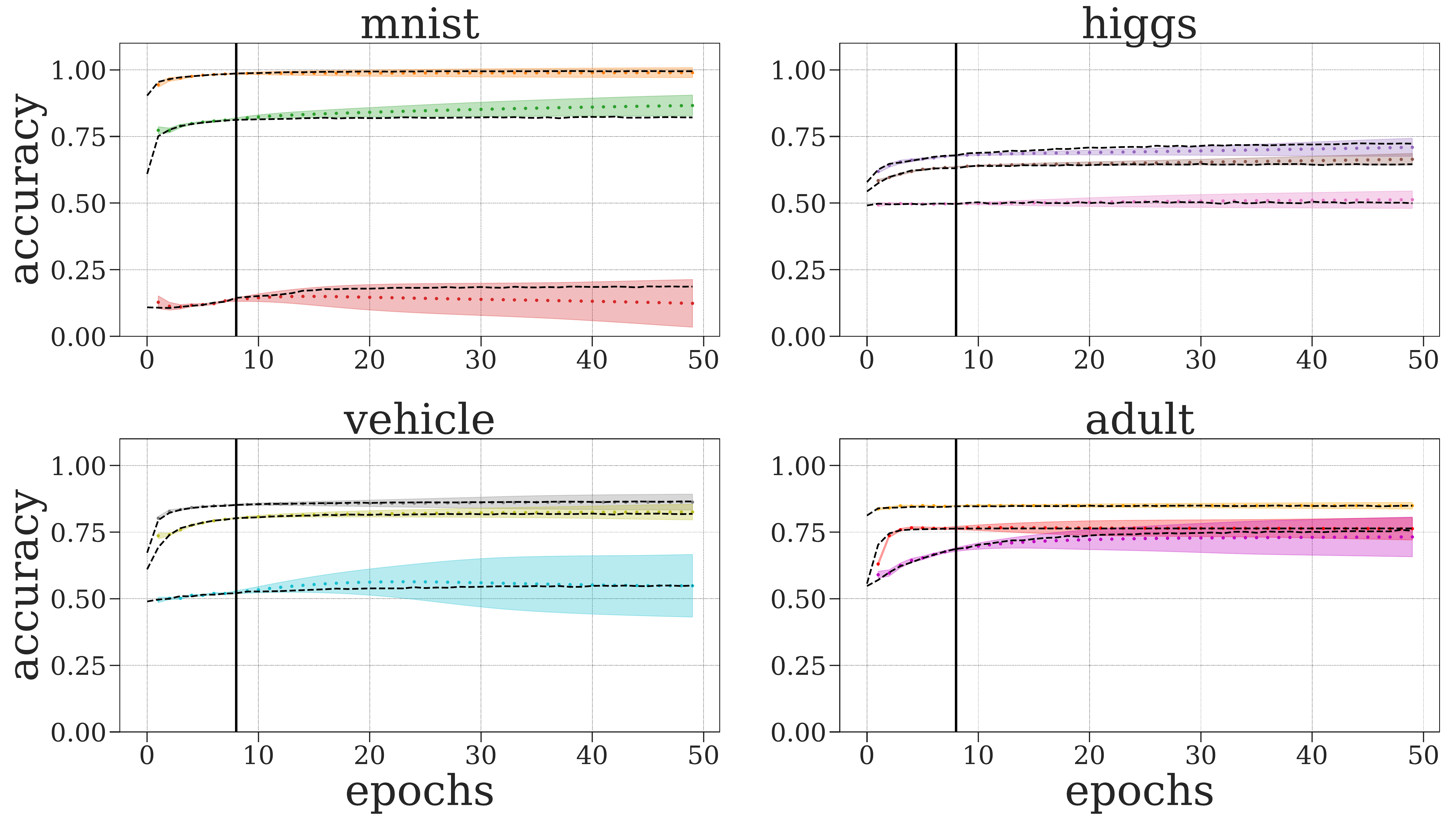}
    \end{center}
    \caption{Qualitative assessment of the test roll-out performances of VRNN for 8 observed epochs (the black vertical line). Different colors of learning curves stand for different configurations, while the black dashed lines represent the true learning curves.}
    \label{fig:vrnn_rollout_8}
\end{figure}
\begin{figure}[H]
    \begin{center}
        \includegraphics[width=.8\textwidth]{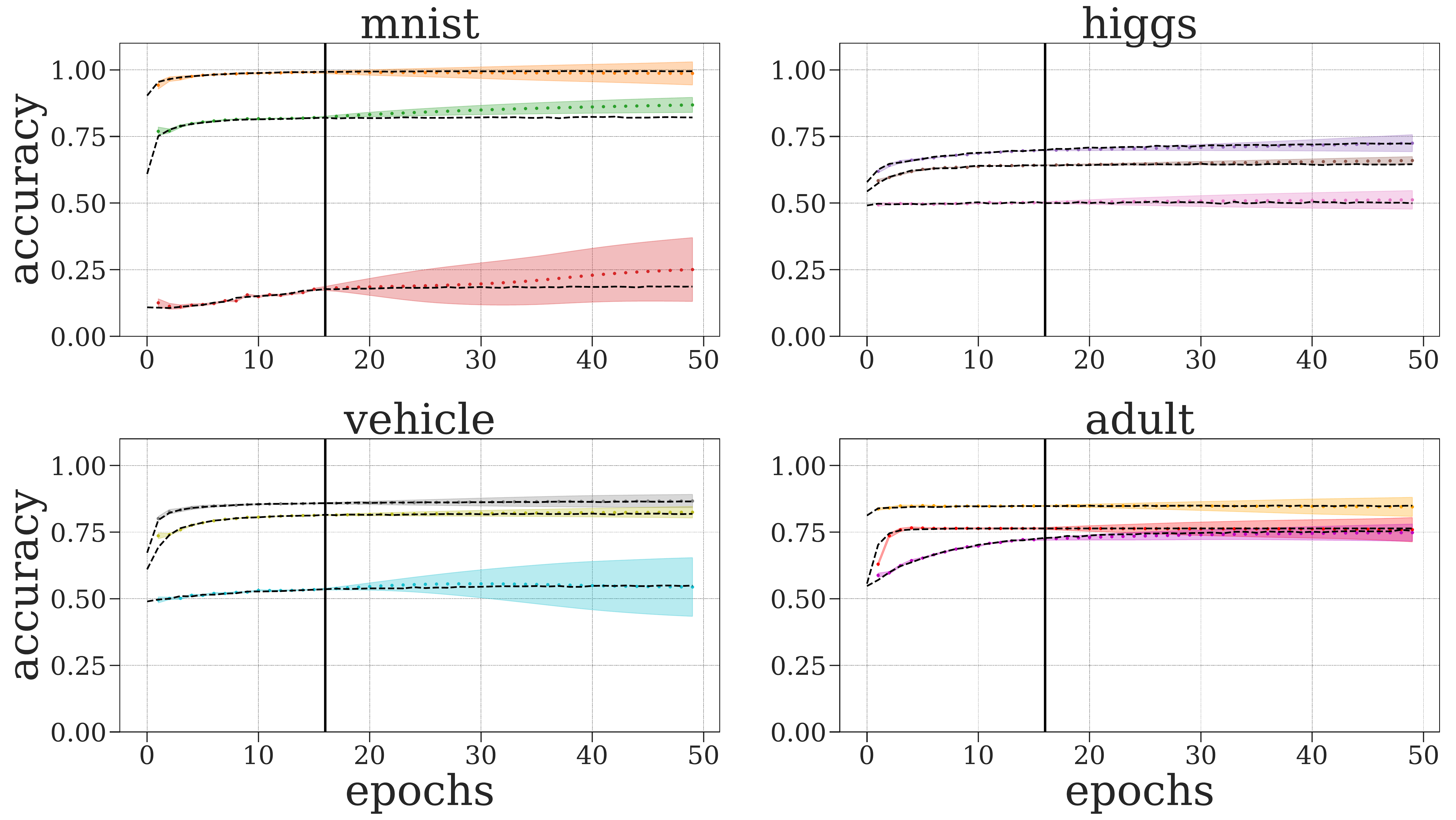}
    \end{center}
    \caption{Qualitative assessment of the test roll-out performances of VRNN for 16 observed epochs (the black vertical line). Different colors of learning curves stand for different configurations, while the black dashed lines represent the true learning curves.}
    \label{fig:vrnn_rollout_16}
\end{figure}
\begin{figure}[H]
    \begin{center}
        \includegraphics[width=.8\textwidth]{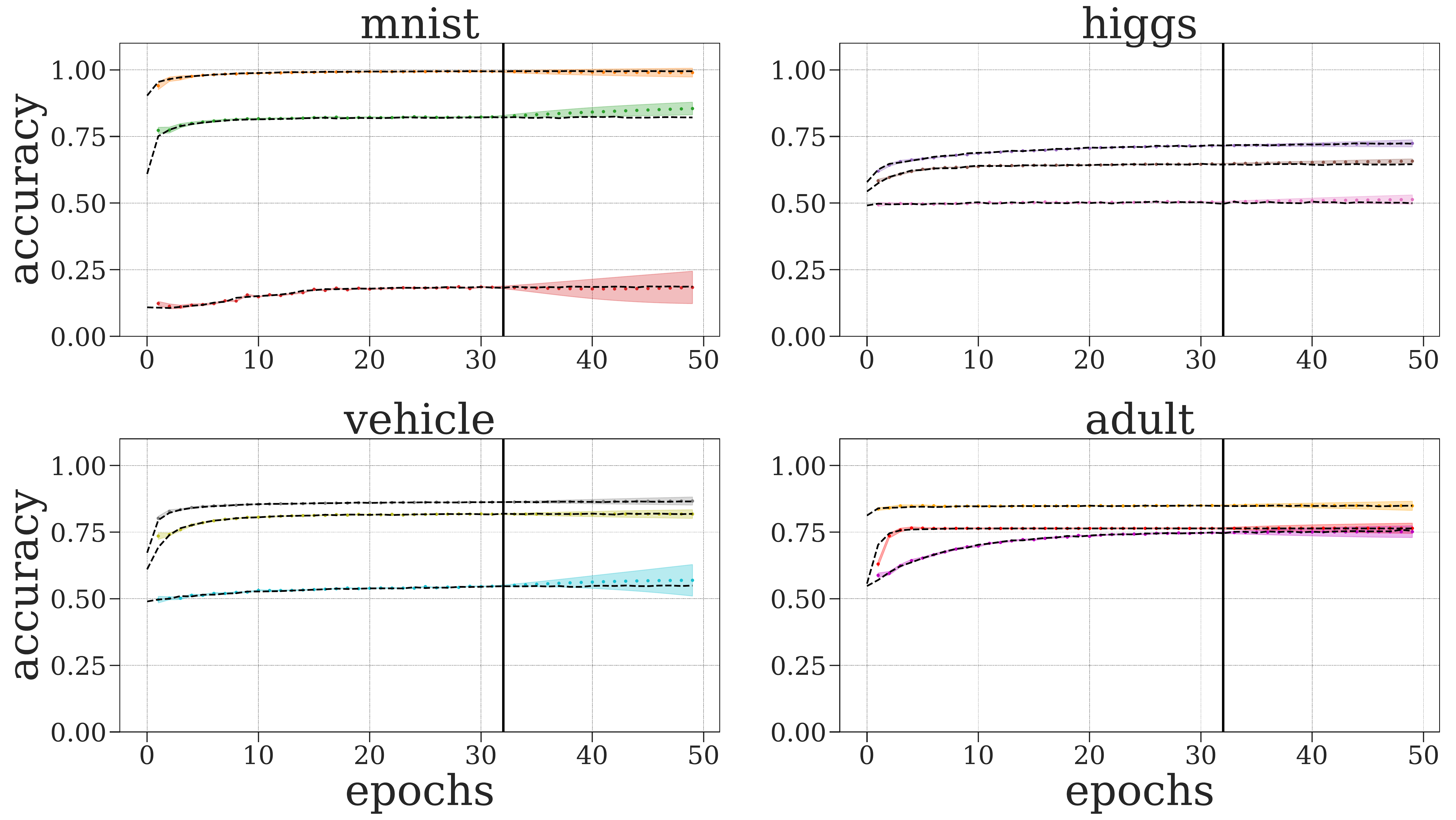}
    \end{center}
    \caption{Qualitative assessment of the test roll-out performances of VRNN for 32 observed epochs (the black vertical line). Different colors of learning curves stand for different configurations, while the black dashed lines represent the true learning curves.}
    \label{fig:vrnn_rollout_32}
\end{figure}
%RFR input size 4 unrolling the learning curves when different epochs are  observed
\begin{figure}[H]

    \begin{center}
        \includegraphics[width=.8\textwidth]{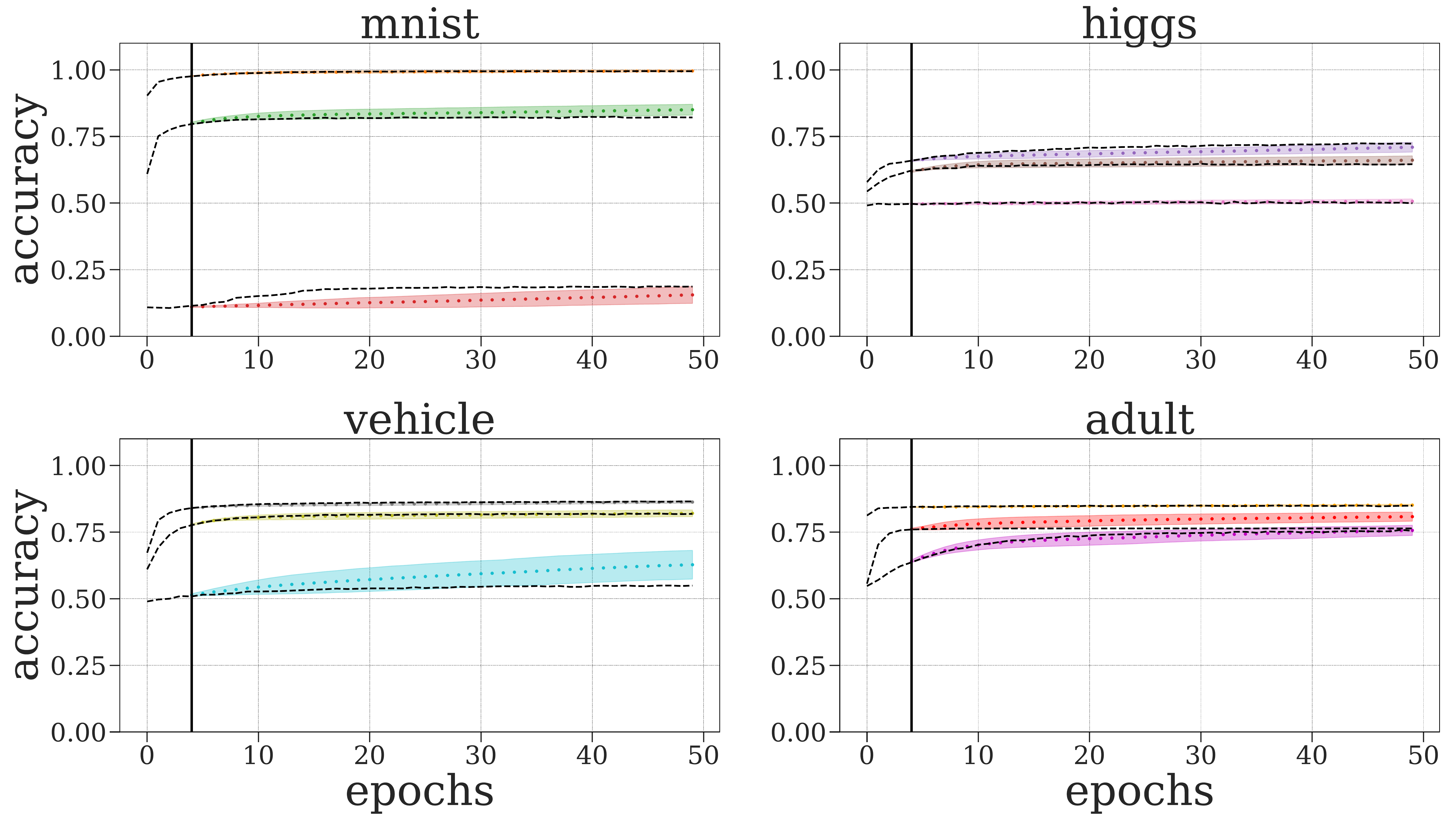}
    \end{center}
    \caption{Qualitative assessment of the test roll-out performances of RF 4 for 4 observed epochs (the black vertical line). Different colors of the learning curves stand for different configurations, while the black dashed lines represent the true learning curves.}
    \label{fig:rfr_s4_rollout_4}
\end{figure}

\begin{figure}[H]

    \begin{center}
        \includegraphics[width=.8\textwidth]{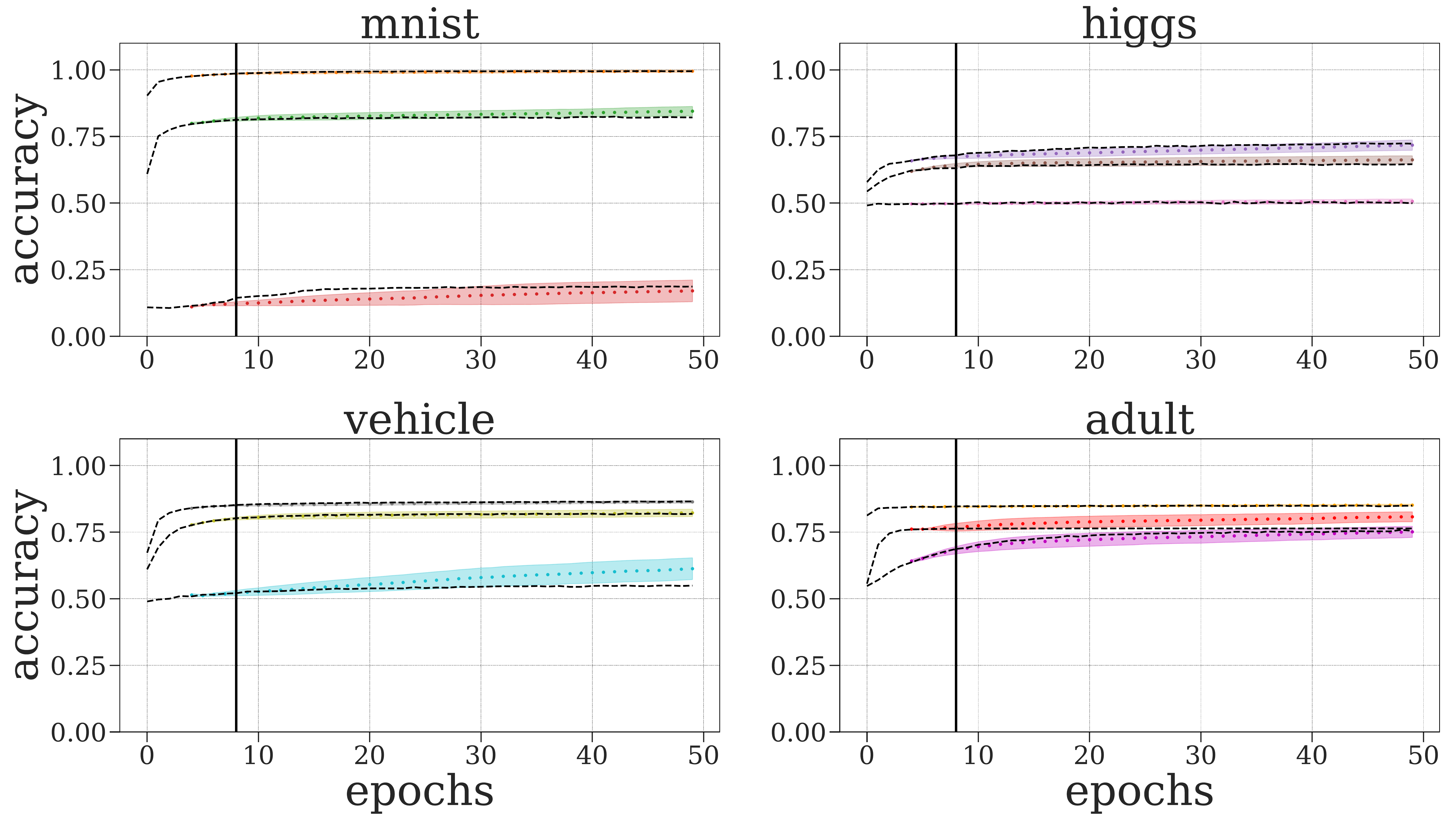}
    \end{center}
    \caption{Qualitative assessment of the test roll-out performances of RF 4 for 8 observed epochs (the black vertical line). Different colors of the learning curves stand for different configurations, while the black dashed lines represent the true learning curves.}
    \label{fig:rfr_s4_rollout_8}
\end{figure}

\begin{figure}[H]

    \begin{center}
        \includegraphics[width=.8\textwidth]{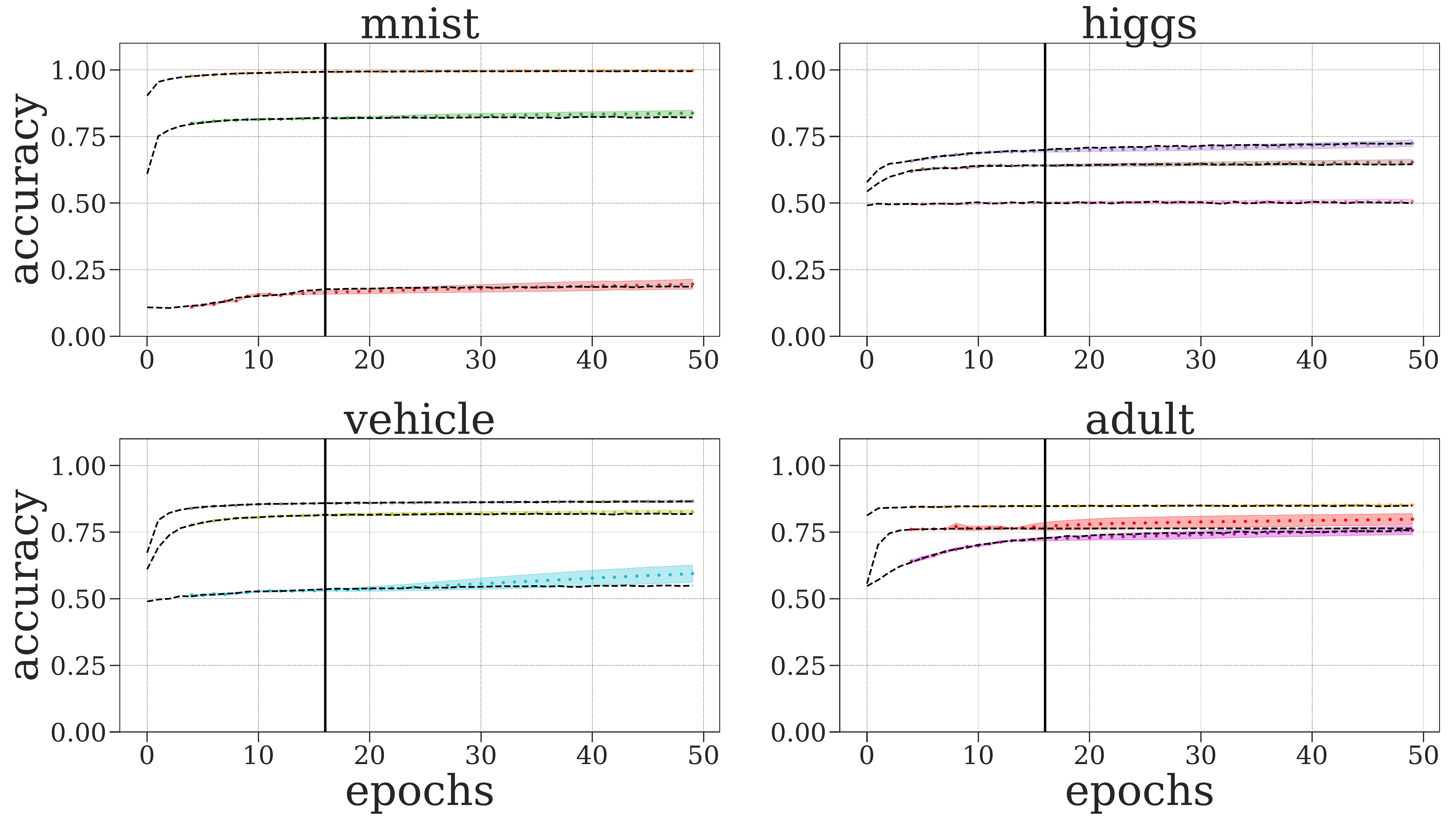}
    \end{center}
    \caption{Qualitative assessment of the test roll-out performances of RF 4 for 16 observed epochs (the black vertical line). Different colors of the learning curves stand for different configurations, while the black dashed lines represent the true learning curves.}
    \label{fig:rfr_s4_rollout_16}
\end{figure}

\begin{figure}[H]

    \begin{center}
        \includegraphics[width=.8\textwidth]{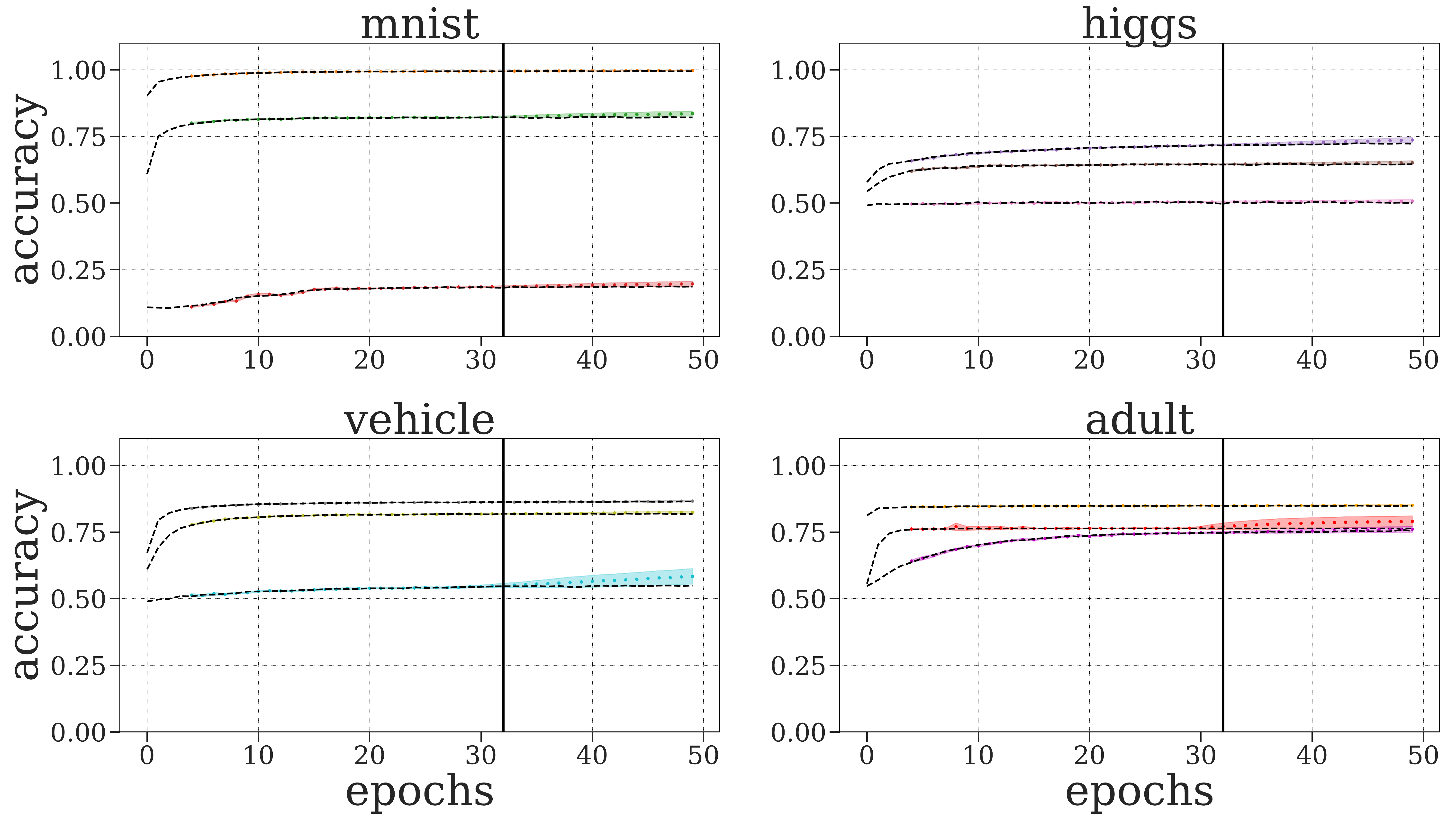}
    \end{center}
    \caption{Qualitative assessment of the test roll-out performances of RF 4 for 32 observed epochs (the black vertical line). Different colors of the learning curves stand for different configurations, while the black dashed lines represent the true learning curves.}
    \label{fig:rfr_s4_rollout_32}
\end{figure}
%RFR input size 1 unrolling the learning curves when different epochs are  observed
\begin{figure}[H]
    \begin{center}
        \includegraphics[width=.8\textwidth]{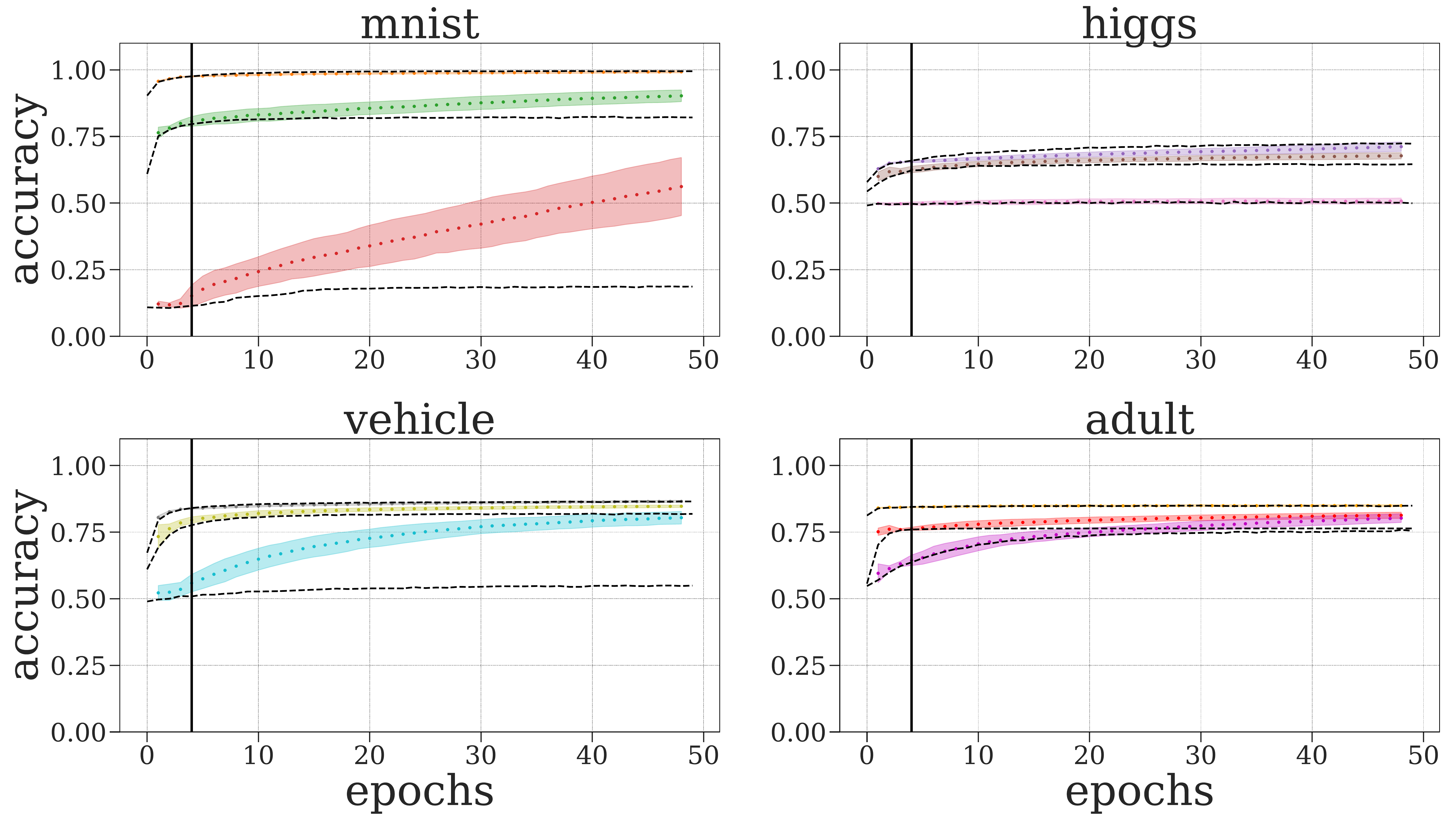}
    \end{center}
    \caption{Qualitative assessment of the test roll-out performances of RF 1 for 4 observed epochs (the black vertical line). Different colors of the learning curves stand for different configurations, while the black dashed lines represent the true learning curves.}
    \label{fig:rfr_s1_rollout_4}
\end{figure}

\begin{figure}[H]
    \begin{center}
        \includegraphics[width=.8\textwidth]{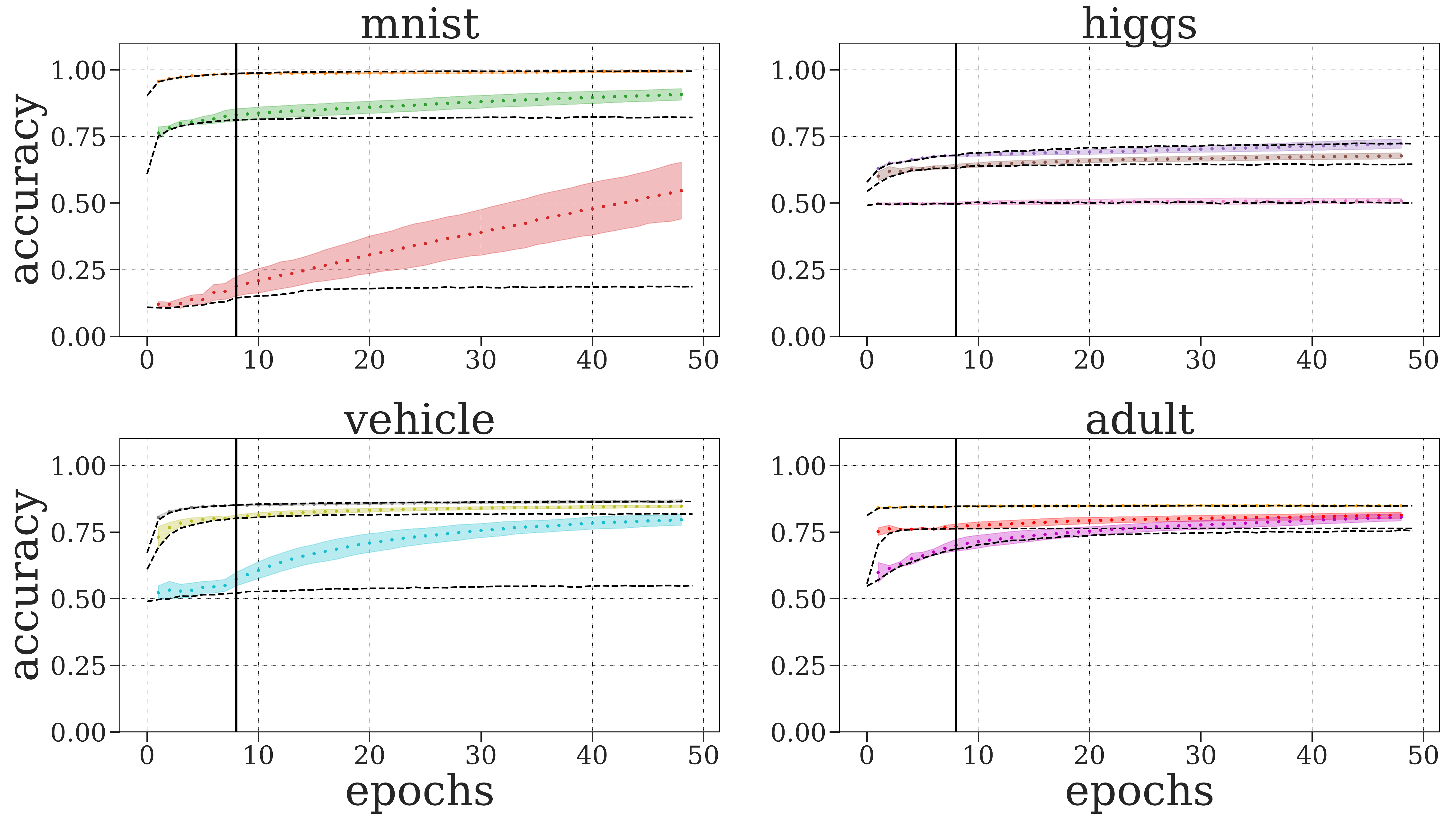}
    \end{center}
    \caption{Qualitative assessment of the test roll-out performances of RF 1 for 8 observed epochs (the black vertical line). Different colors of the learning curves stand for different configurations, while the black dashed lines represent the true learning curves.}
    \label{fig:rfr_s1_rollout_8}
\end{figure}

\begin{figure}[H]
    \begin{center}
        \includegraphics[width=.8\textwidth]{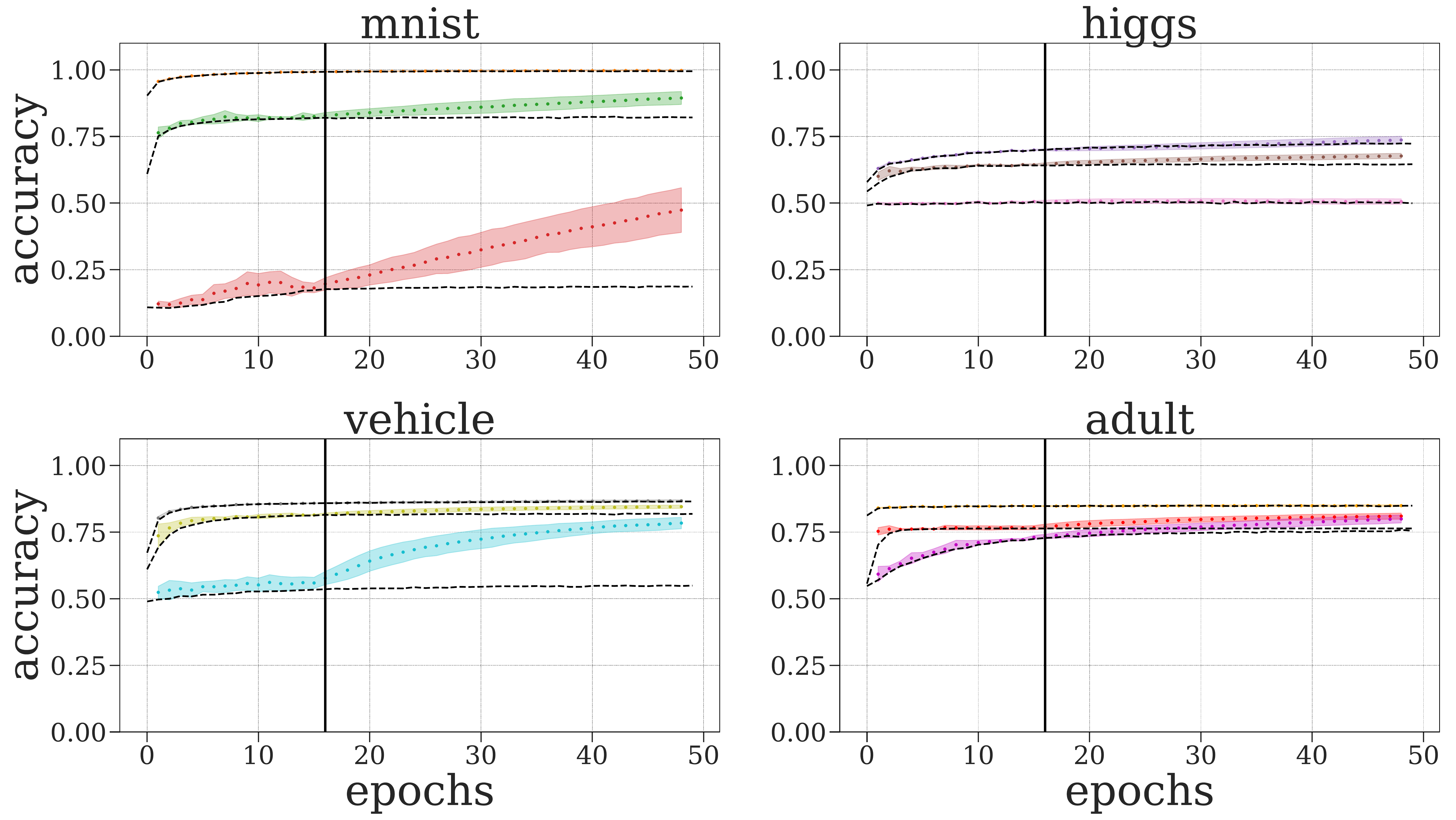}
    \end{center}
    \caption{Qualitative assessment of the test roll-out performances of RF 1 for 16 observed epochs (the black vertical line). Different colors of the learning curves stand for different configurations, while the black dashed lines represent the true learning curves.}
    \label{fig:rfr_s1_rollout_16}
\end{figure}

\begin{figure}[H]
    \begin{center}
        \includegraphics[width=.8\textwidth]{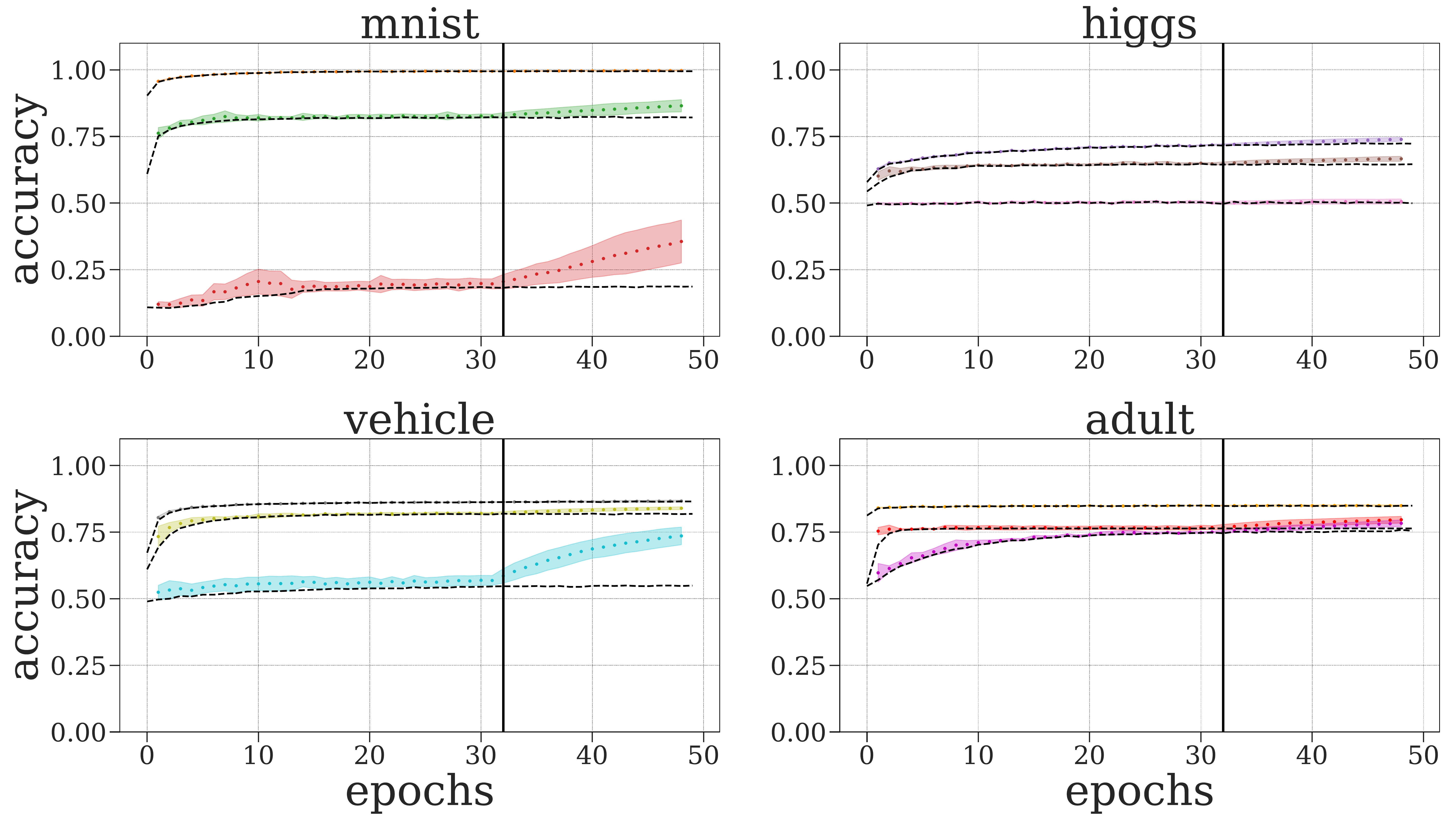}
    \end{center}
    \caption{Qualitative assessment of the test roll-out performances of RF 1 for 32 epochs (the black vertical line). Different colors of the learning curves stand for different configurations, while the black dashed lines represent the true learning curves.}
    \label{fig:rfr_s1_rollout_32}
\end{figure}
%VRNN predicted true plot for 4 observed epochs 
\begin{figure}[H]
    \begin{center}
        \includegraphics[width=.45\textwidth]{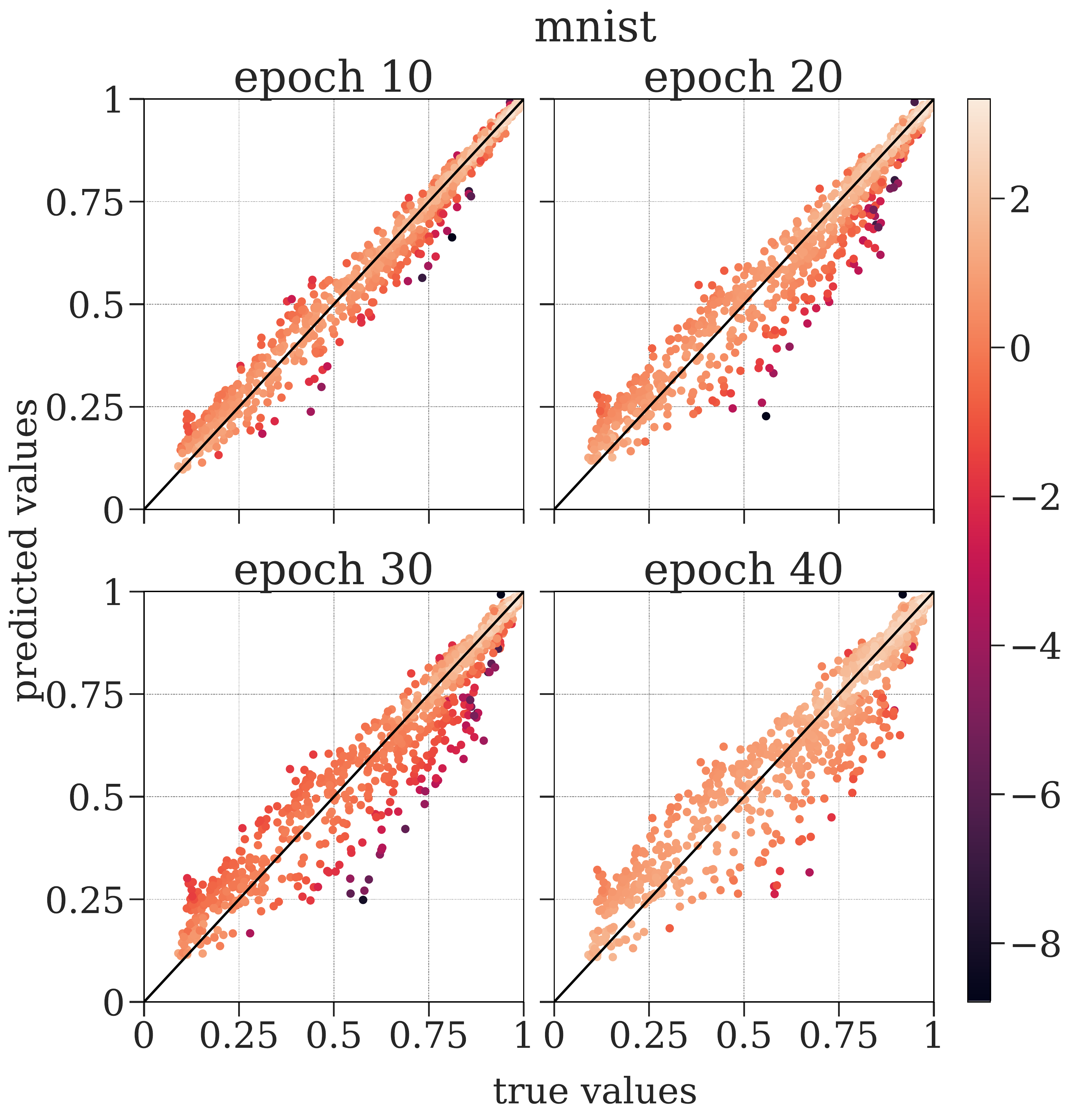}
        \includegraphics[width=.45\textwidth]{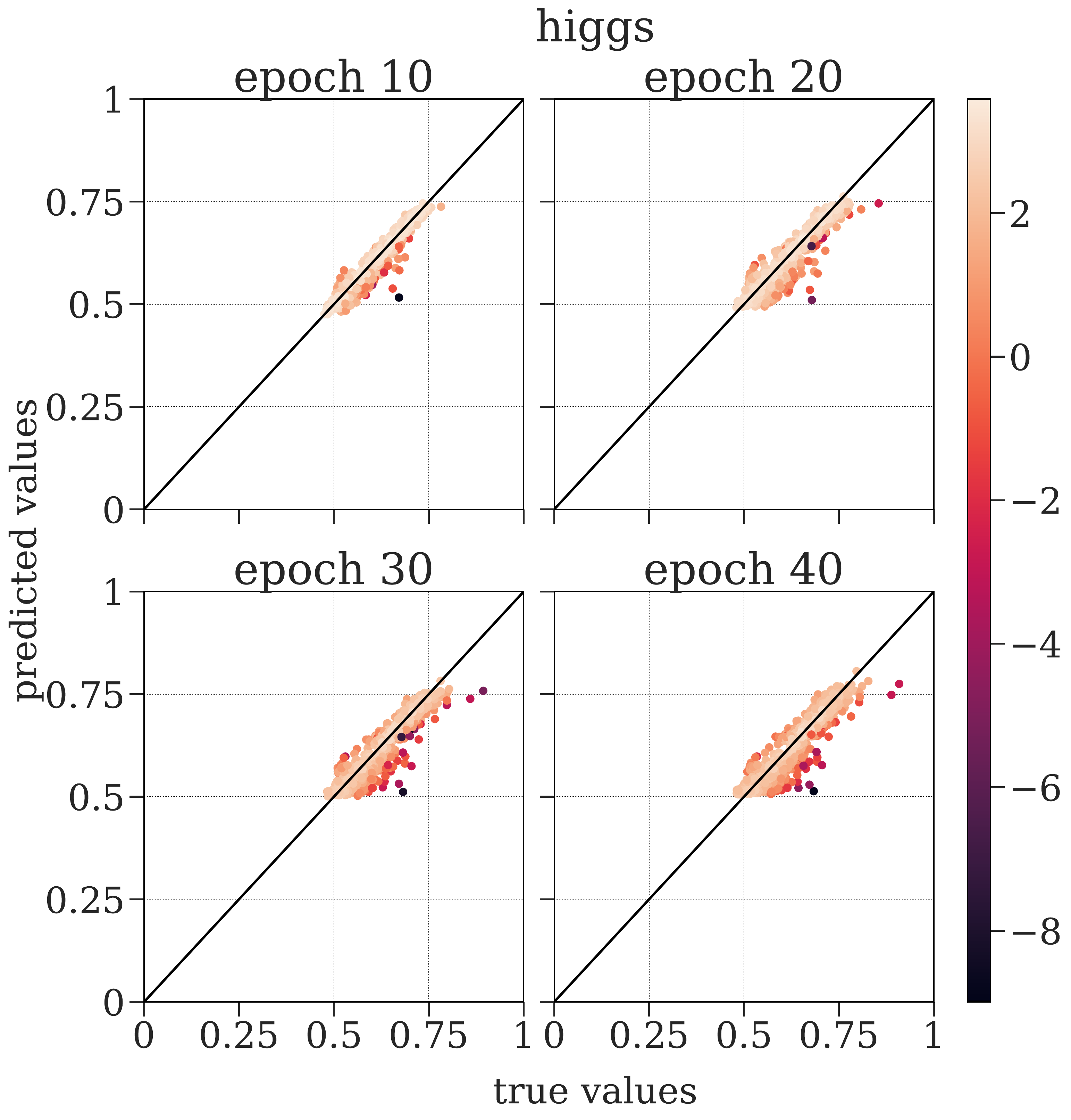}
        \includegraphics[width=.45\textwidth]{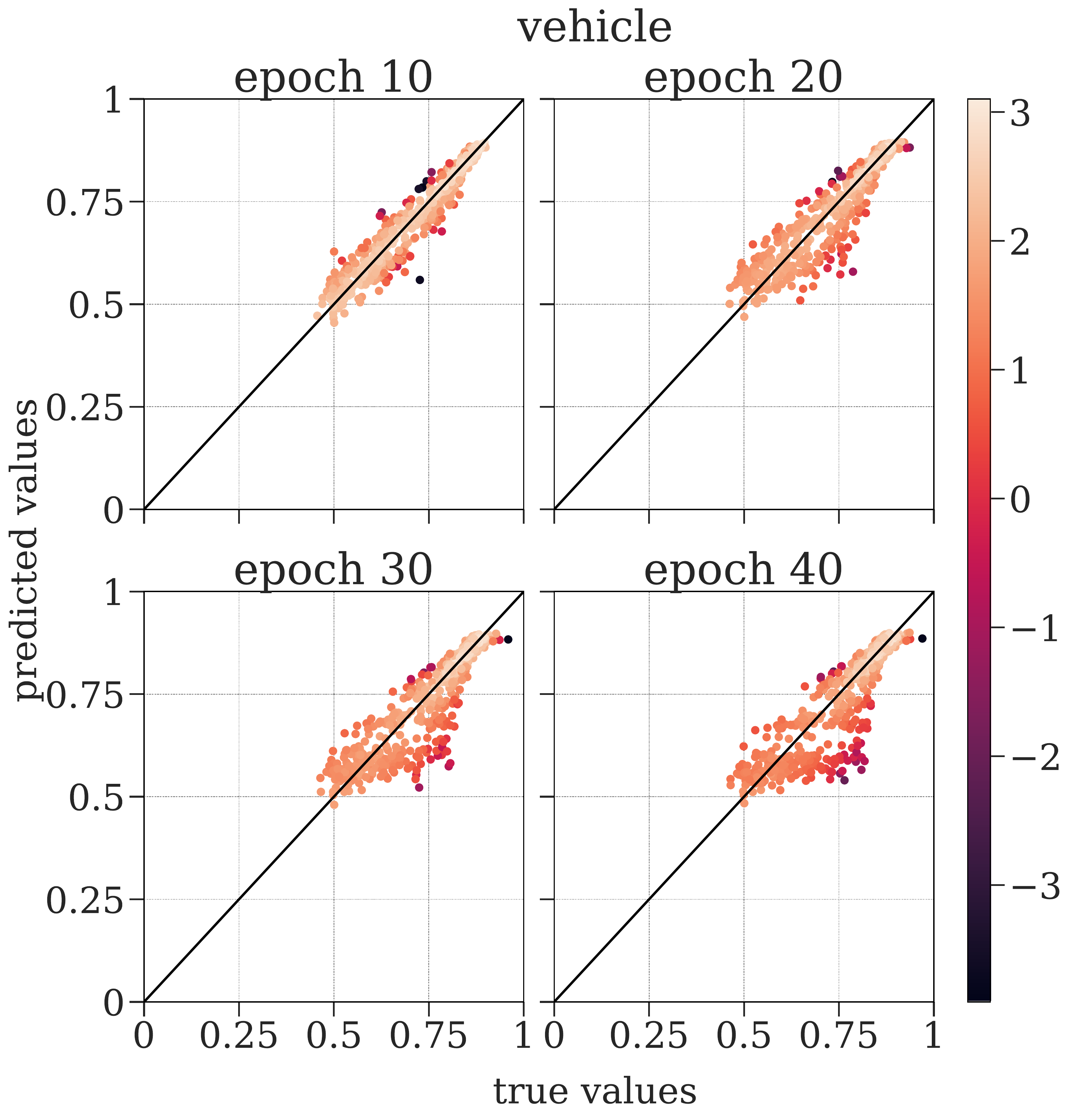}
        \includegraphics[width=.45\textwidth]{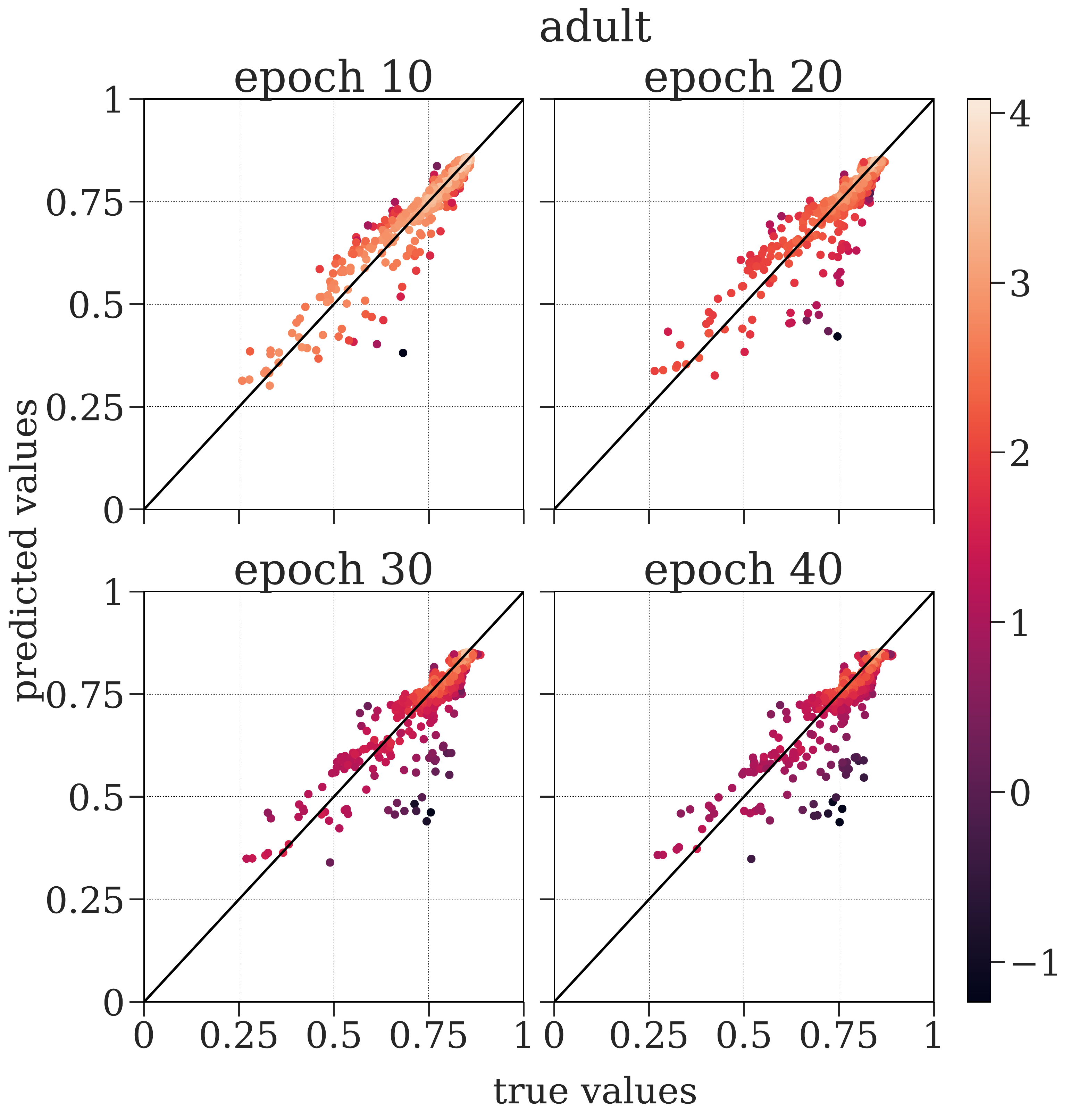}
    \end{center}
    \caption{Qualitative assessment at different target epochs of the test roll-out performances of VRNN with 4 observed epochs on the four different datasets. Each plot shows on the horizontal axis the true values and on the vertical axis the predicted values. Each point is colored based on its log-likelihood value. }
    \label{fig:vrnn_predicted_true_ll}
\end{figure}

%RFR 4 predicted true plot for 4 observed epochs 
\begin{figure}[H]
    \begin{center}
        \includegraphics[width=.45\textwidth]{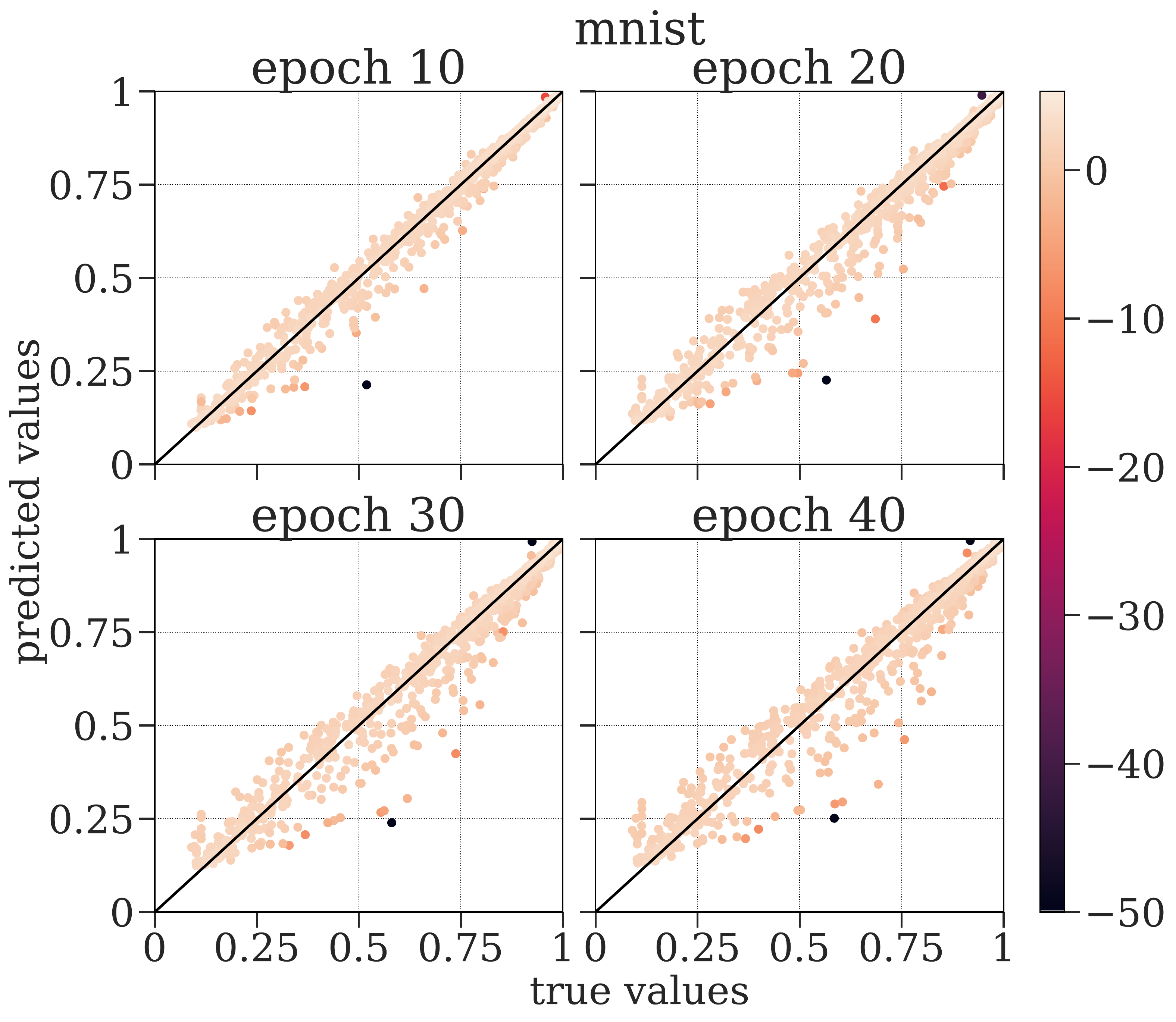}
        \includegraphics[width=.45\textwidth]{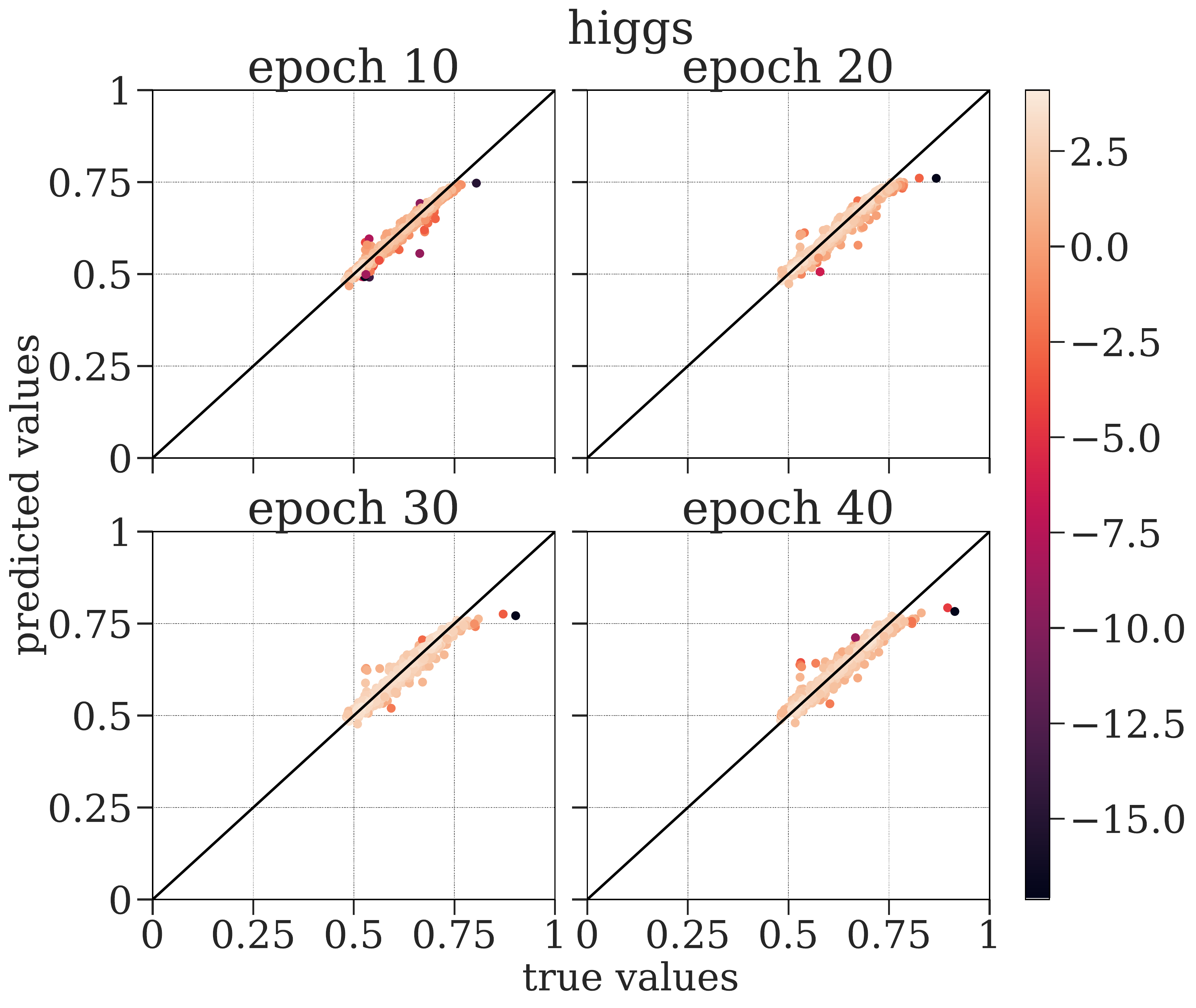}
        \includegraphics[width=.45\textwidth]{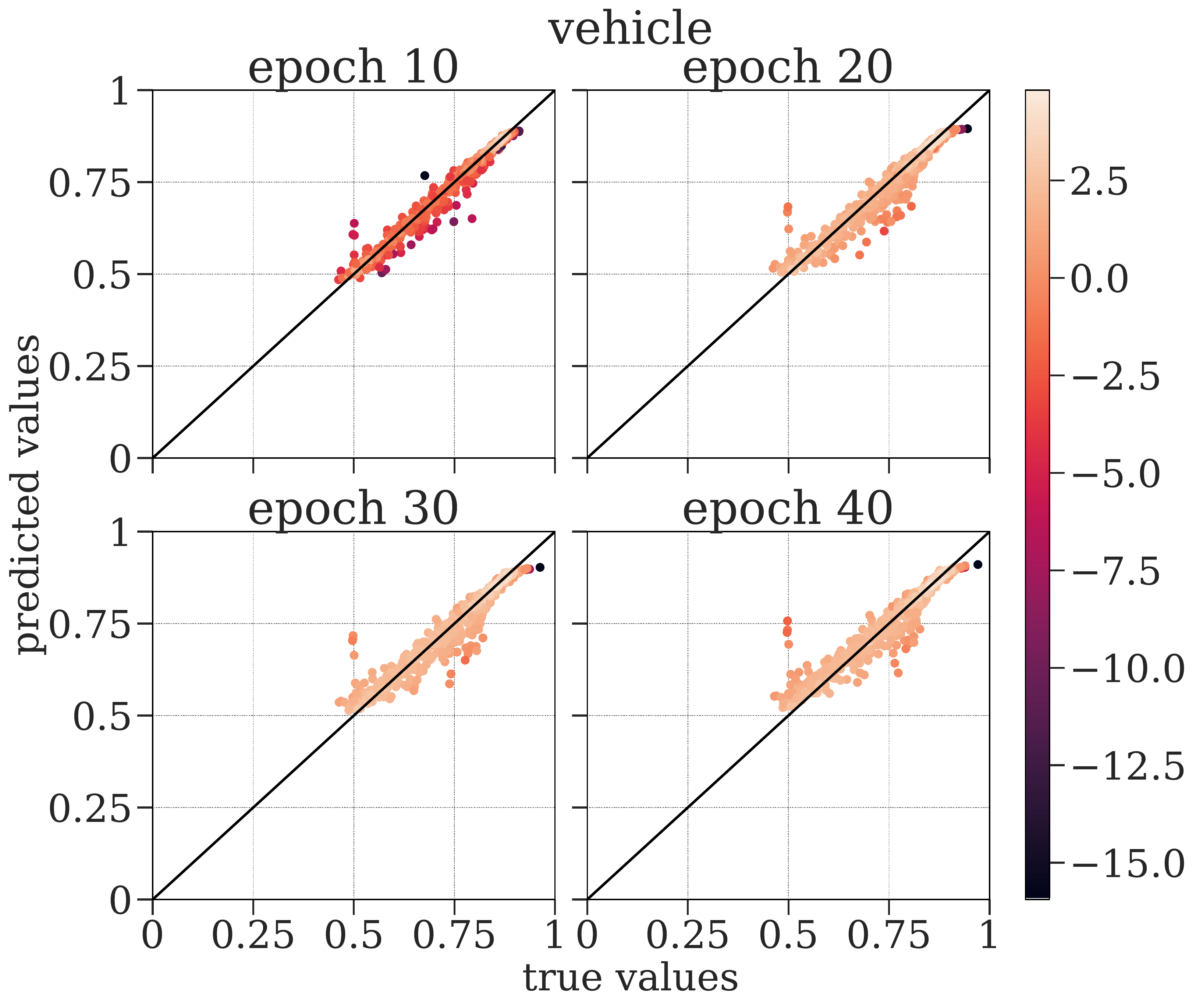}
        \includegraphics[width=.45\textwidth]{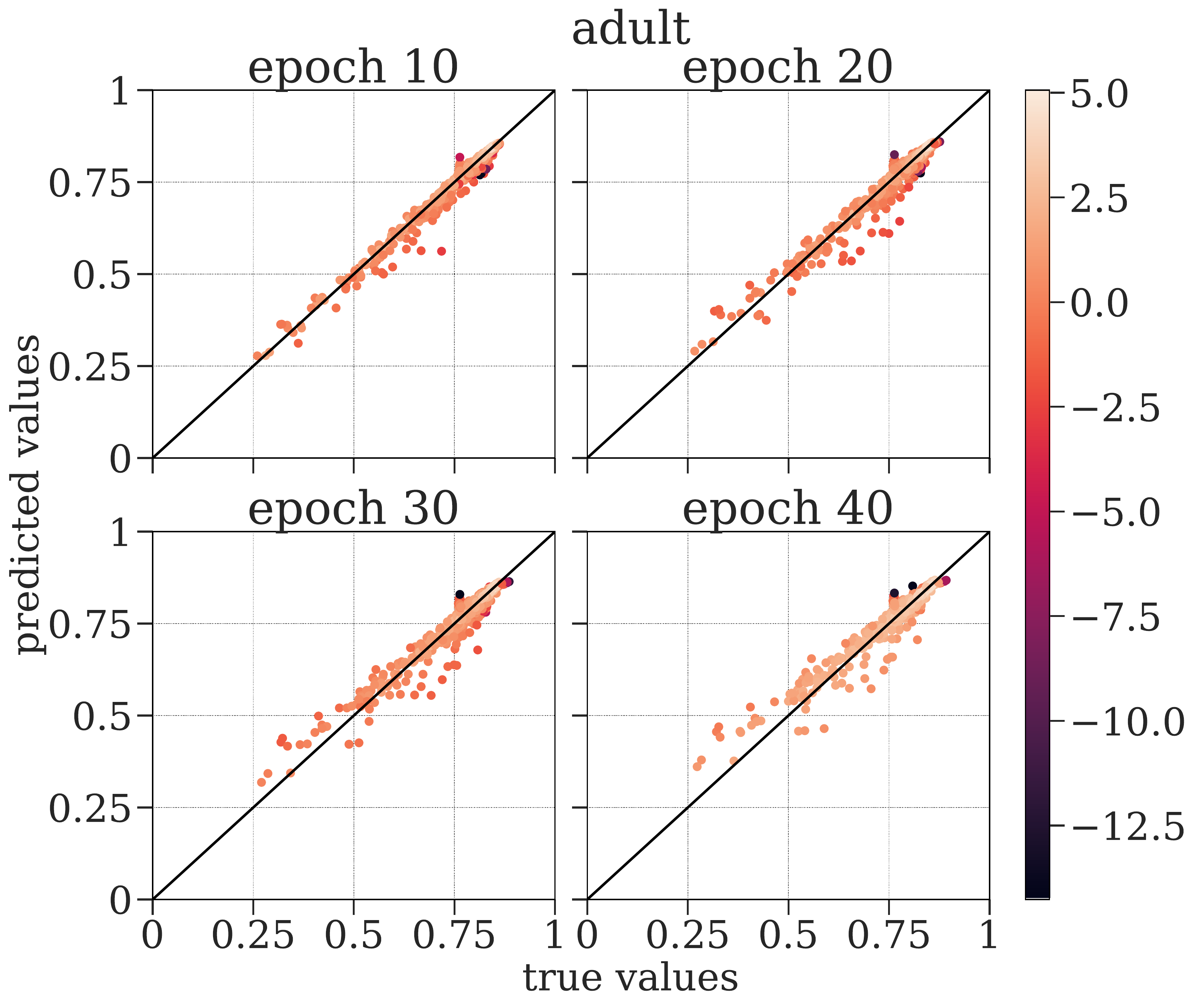}
    \end{center}
    \caption{Qualitative assessment at different target epochs of the test roll-out performances of RF 4 with 4 observed epochs on the four different datasets. Each plot shows on the horizontal axis the true values and on the vertical axis the predicted values. Each point is colored based on its log-likelihood value. }
    \label{fig:rfr_s2_predicted_true_ll}
\end{figure} 

%RFR 1 predicted true plot for 4 observed epochs 
\begin{figure}[H]
    \begin{center}
        \includegraphics[width=.45\textwidth]{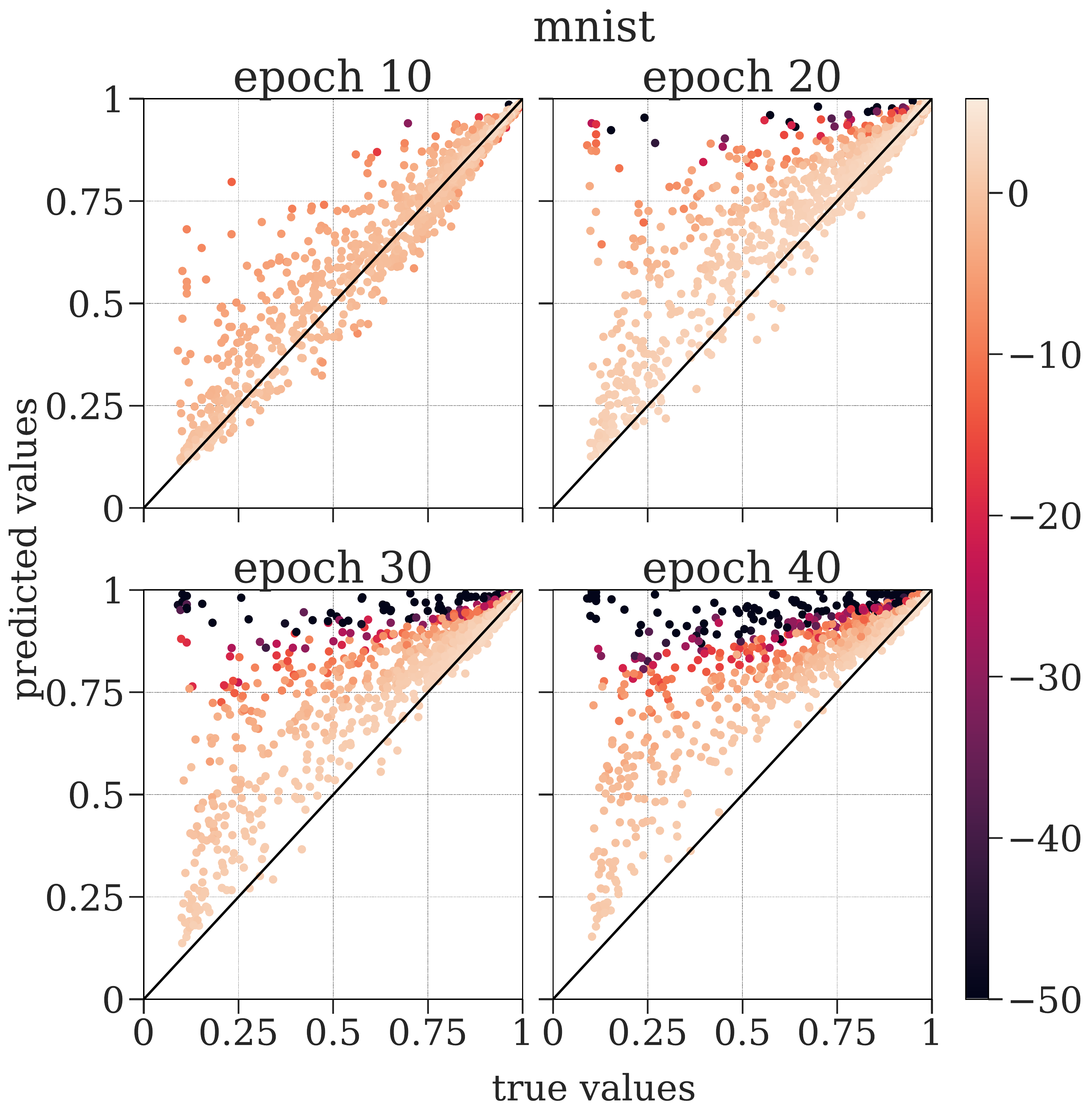}
        \includegraphics[width=.45\textwidth]{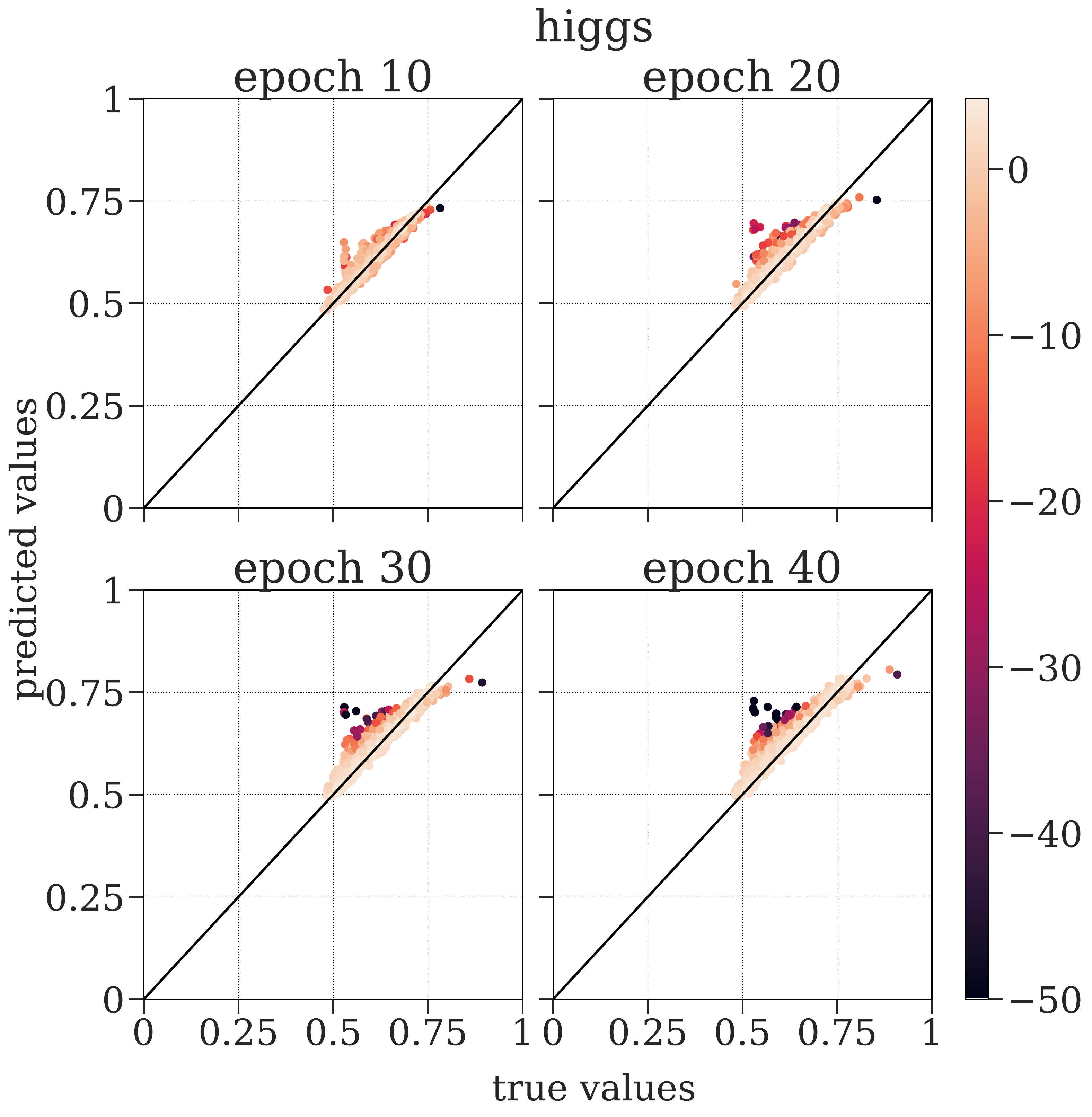}
        \includegraphics[width=.45\textwidth]{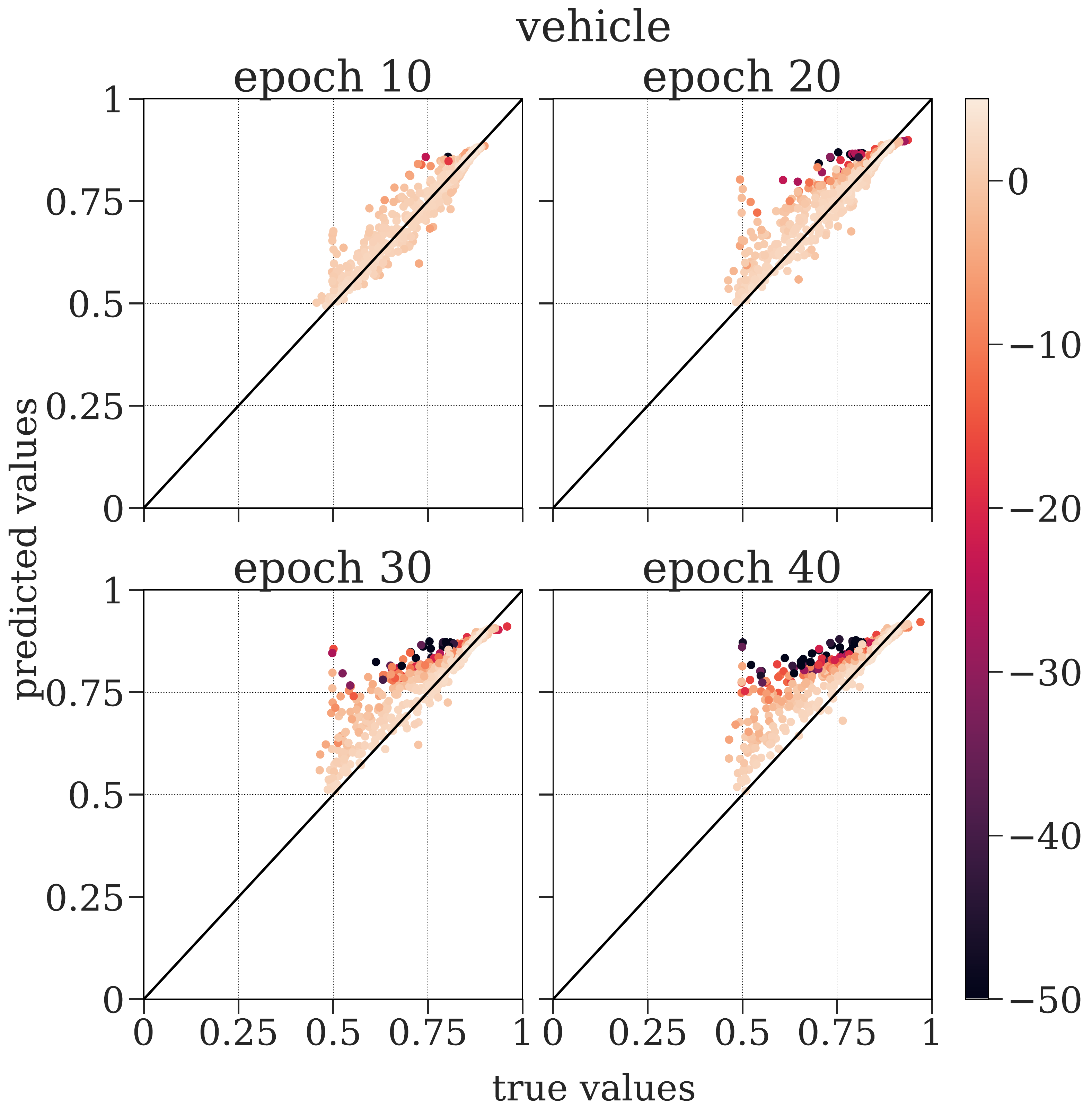}
        \includegraphics[width=.45\textwidth]{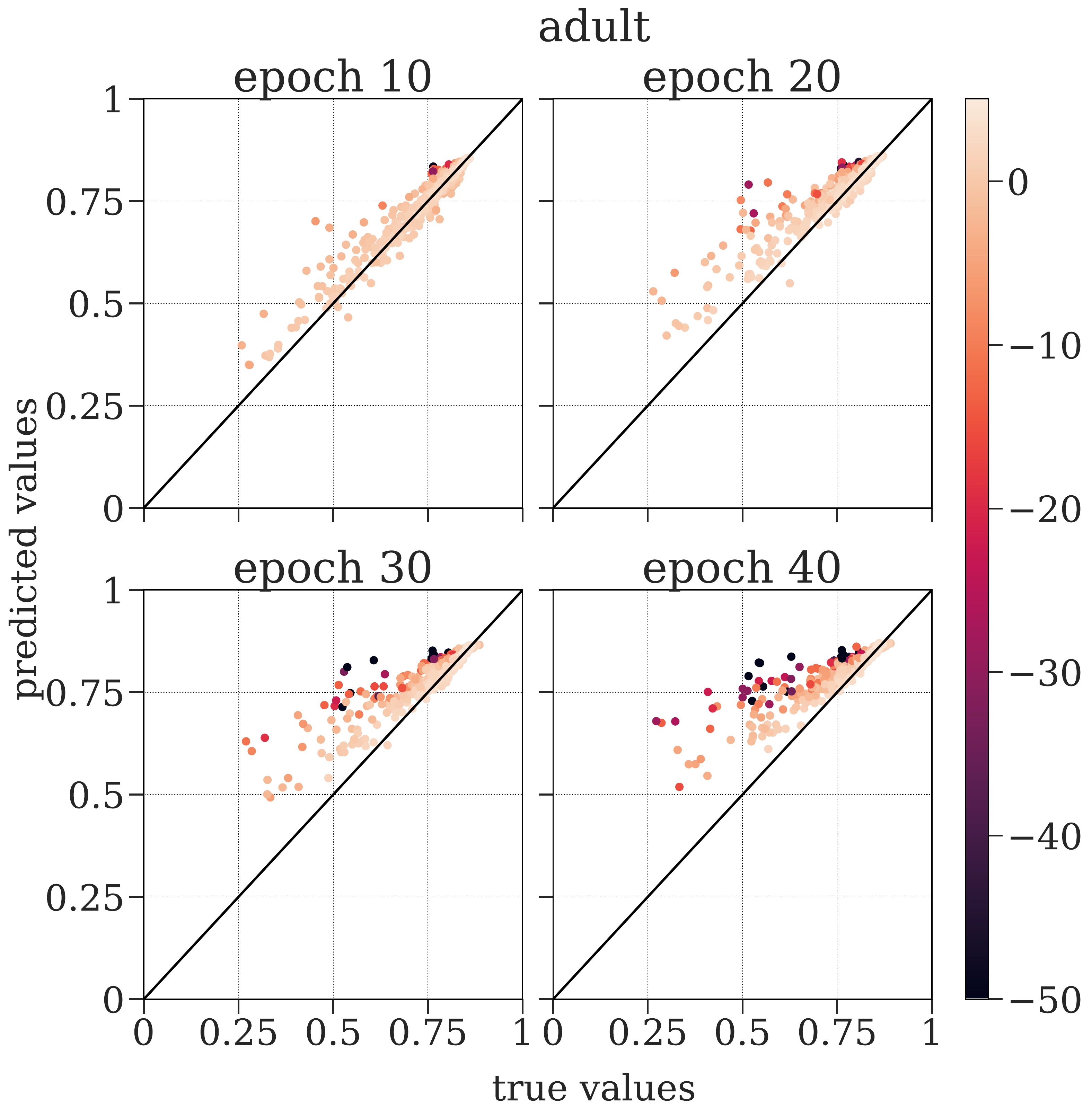}
    \end{center}
    \caption{Qualitative assessment at different target epochs of the test roll-out performances of RF 1 with 4 observed epochs on the four different datasets. Each plot shows on the horizontal axis the true values and on the vertical axis the predicted values. Each point is colored based on its log-likelihood value. }
    \label{fig:rfr_s1_predicted_true_ll}
\end{figure}

%VRNN outlier plot
\begin{figure}[H]
    \centering
    \begin{subfigure}[b]{0.73\textwidth}
    \includegraphics[width=\linewidth]{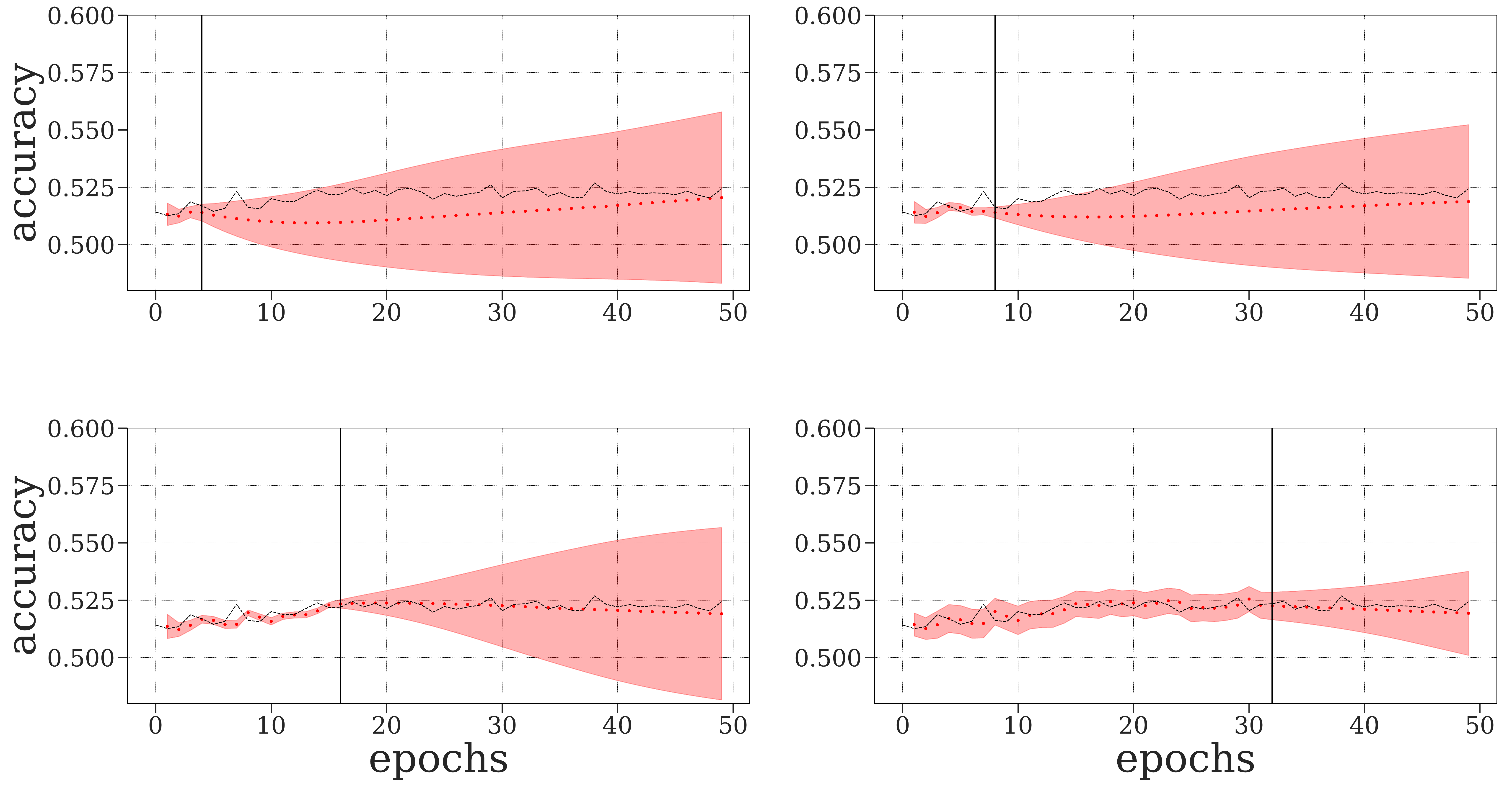}
    \caption{VRNN.}
    \label{fig:outlier_vrnn}
  \end{subfigure}
  \begin{subfigure}[b]{0.73\textwidth}
    \includegraphics[width=\linewidth]{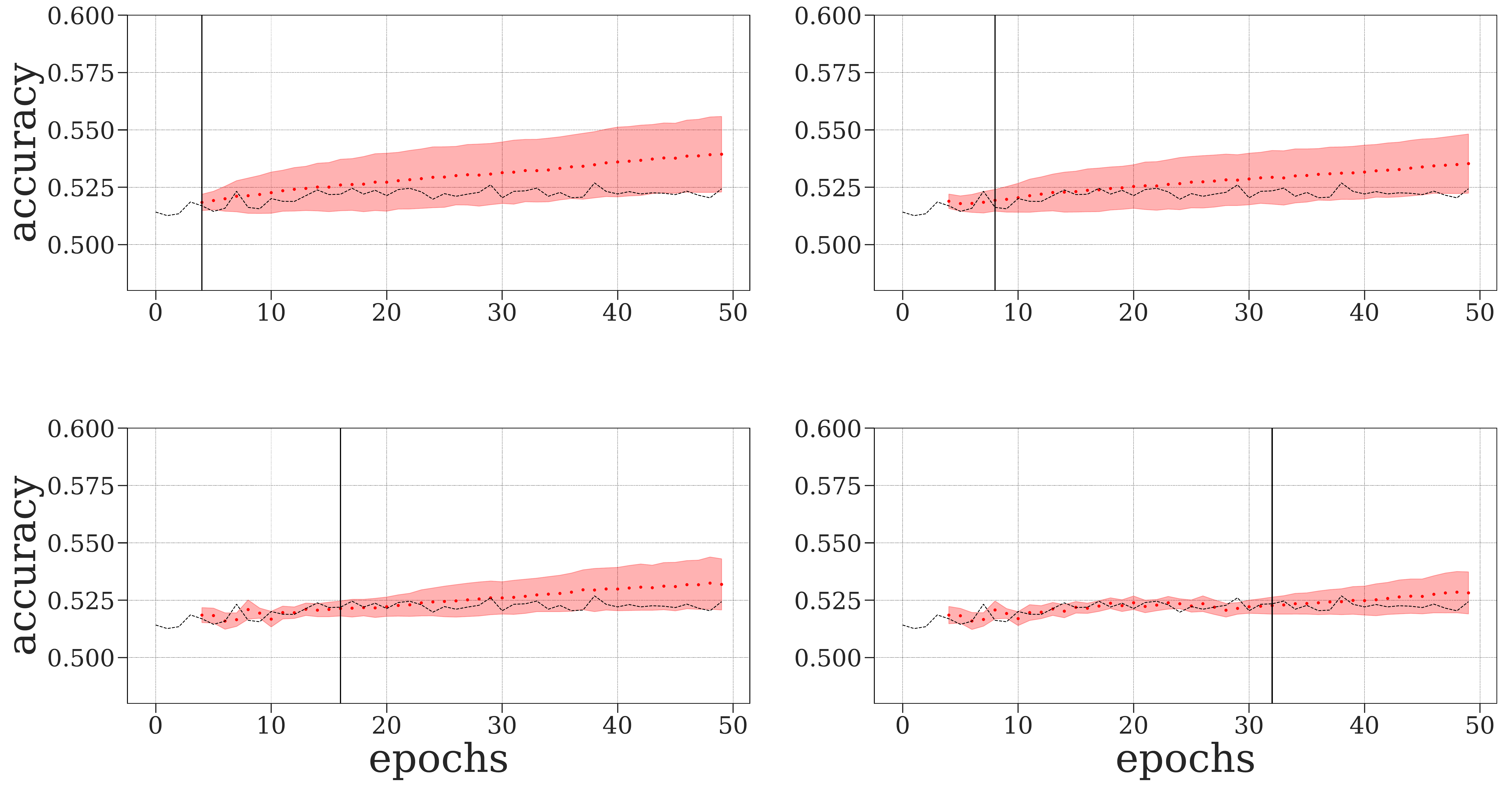}
    \caption{RF 4.}
    \label{fig:outlier_rfr_4}
  \end{subfigure}
  \begin{subfigure}[b]{0.73\textwidth}
    \includegraphics[width=\linewidth]{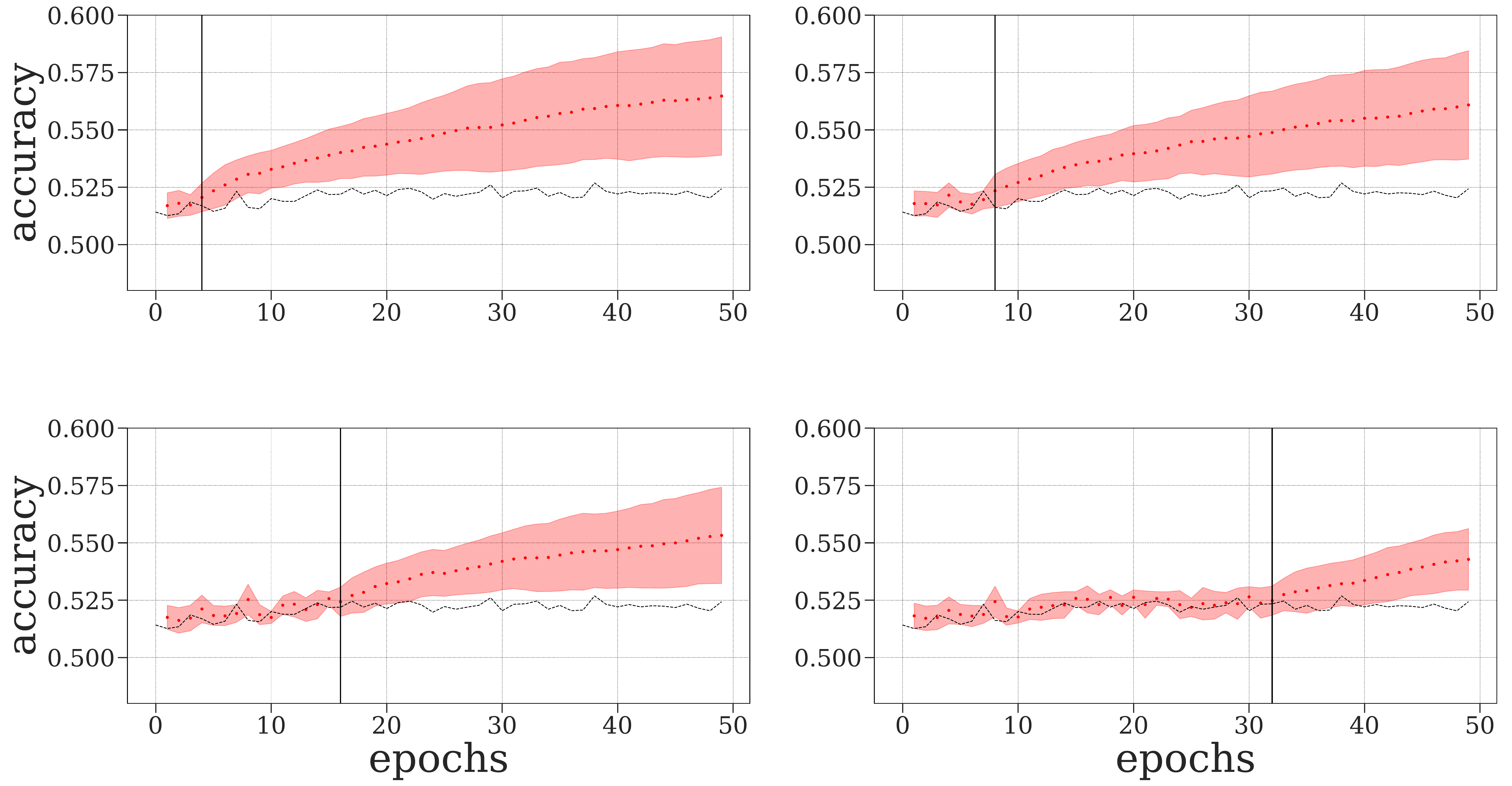}
        \caption{RF 1.}
        \label{fig:outlier_rfr_1}
  \end{subfigure}
    \caption{Predictions of roll-out models for the case of a very bumpy learning curve from the Higgs benchmark.}
    \label{fig:outlier_case}
\end{figure}

%plots with mse and average log-likelihood for different numbers of observed epochs

\begin{figure}[H]
    \begin{center}
        \includegraphics[width=.45\textwidth]{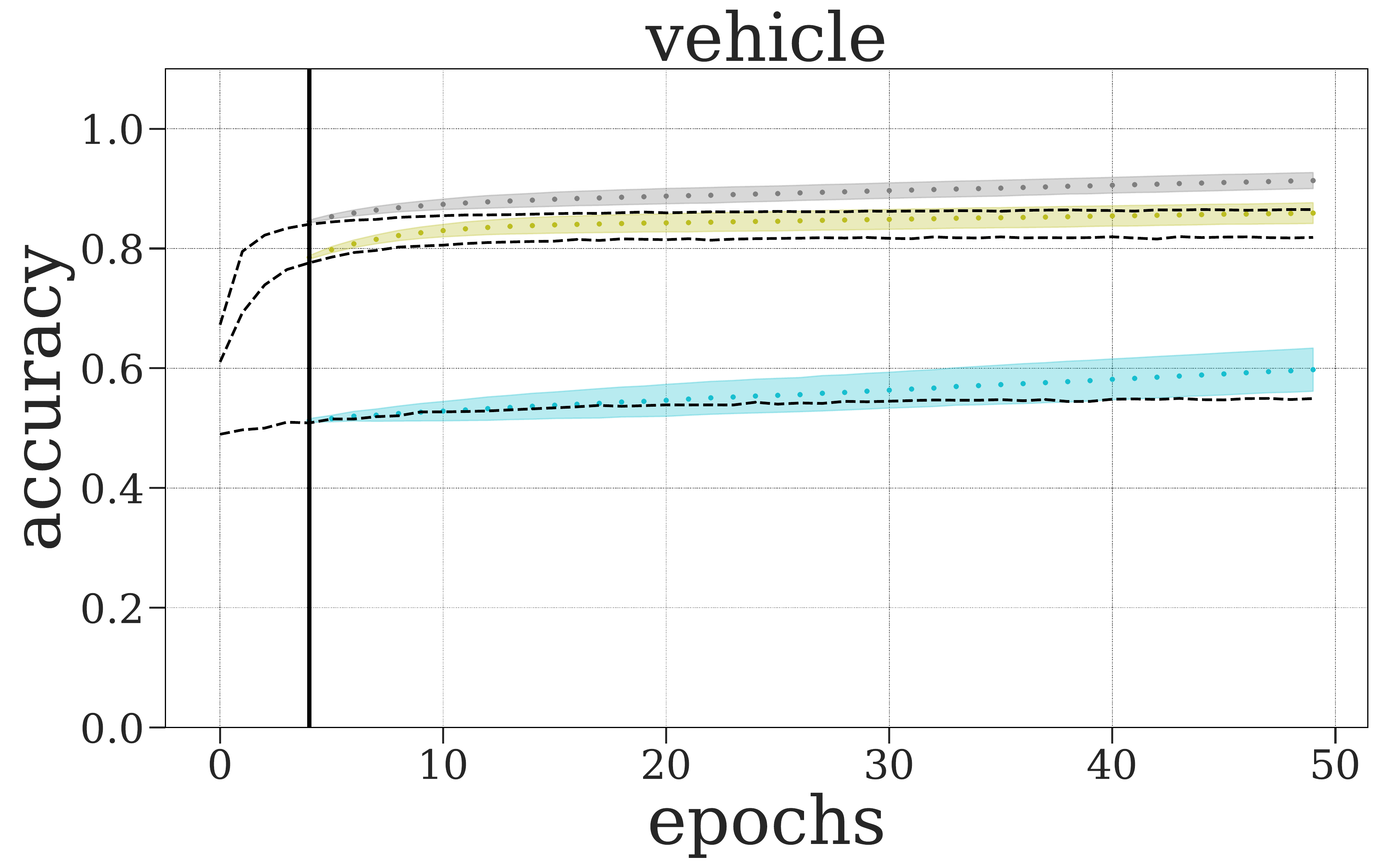}
        \includegraphics[width=.45\textwidth]{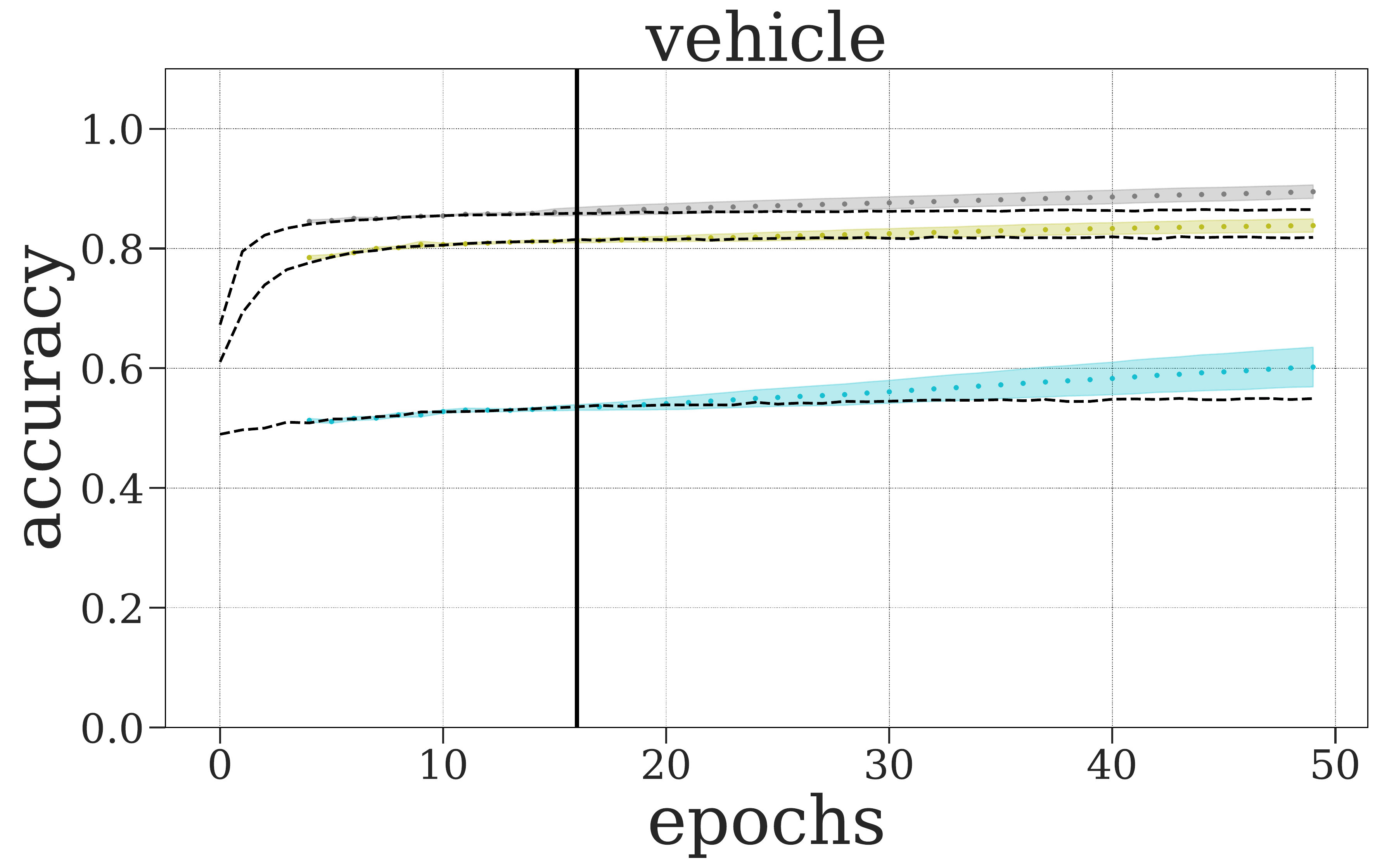}
        \includegraphics[width=.45\textwidth]{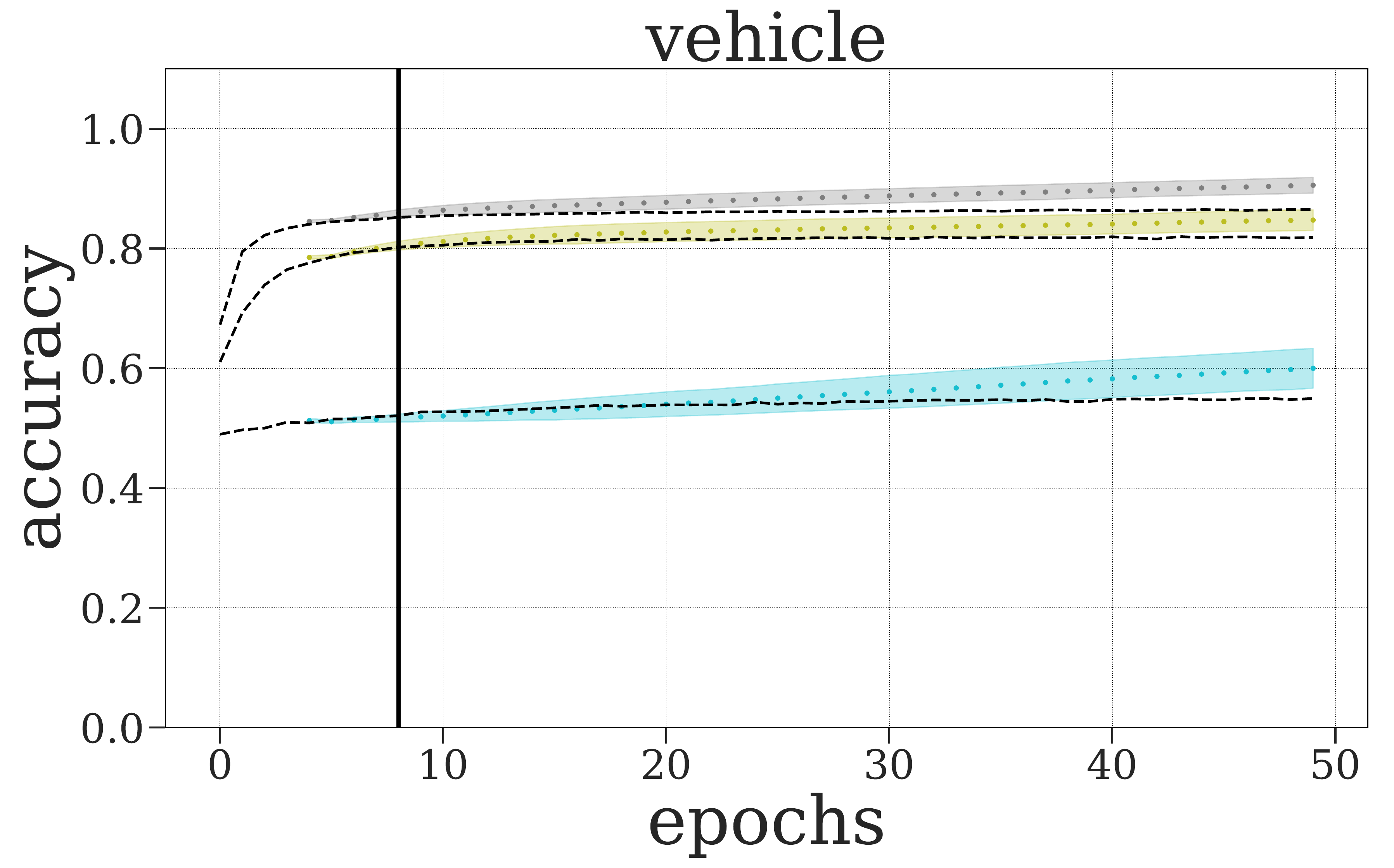}
        \includegraphics[width=.45\textwidth]{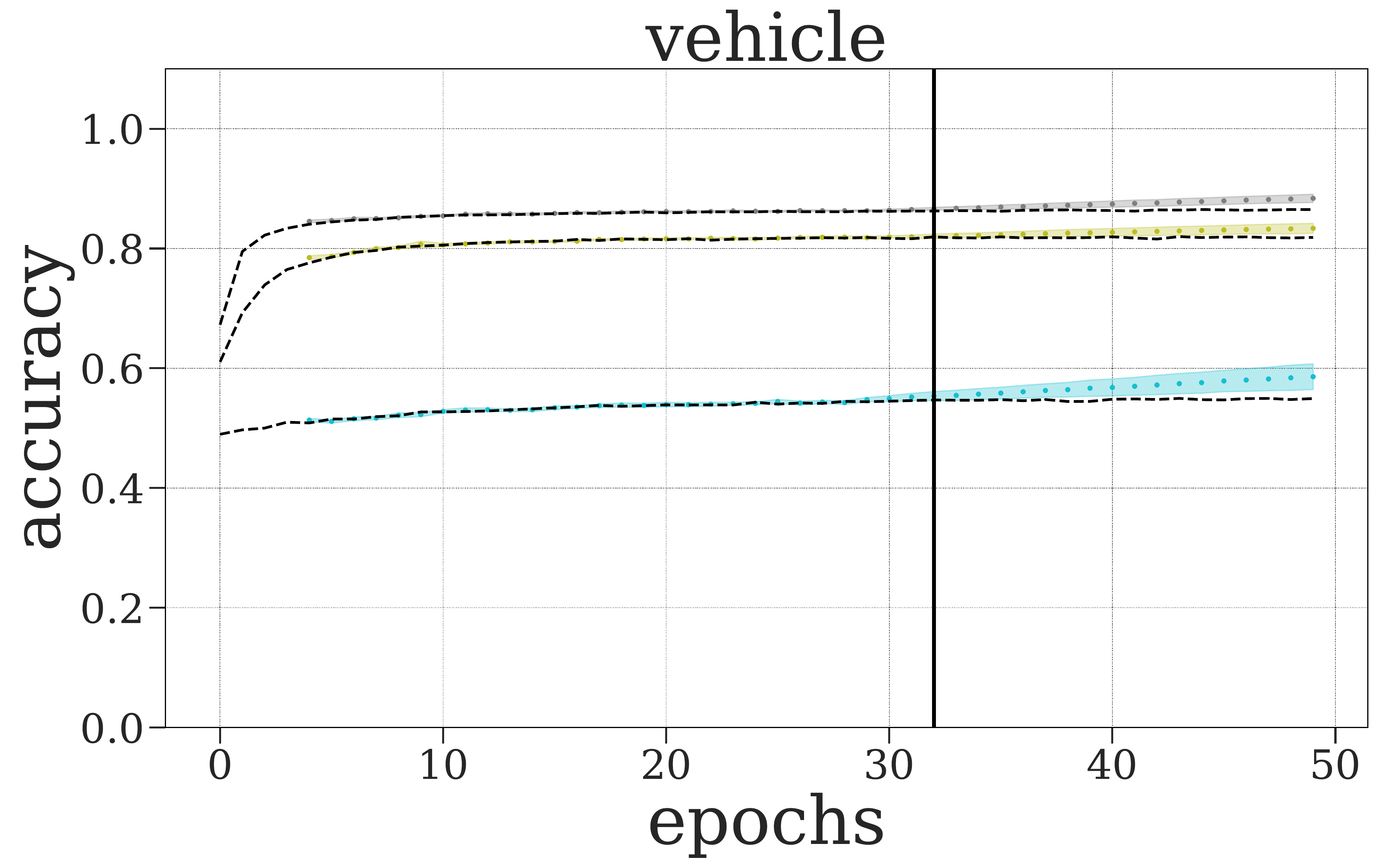}
    \end{center}
    \caption{Qualitative assessment of RF 4 predictions on the Vehicle benchmark when trained on MNIST for different numbers of observed epochs at test time (the black vertical line).}
    \label{fig:rfr_4_vehicle_unseen_dataset}
\end{figure} 

\begin{figure}[H]
    \begin{center}
        \includegraphics[width=.45\textwidth]{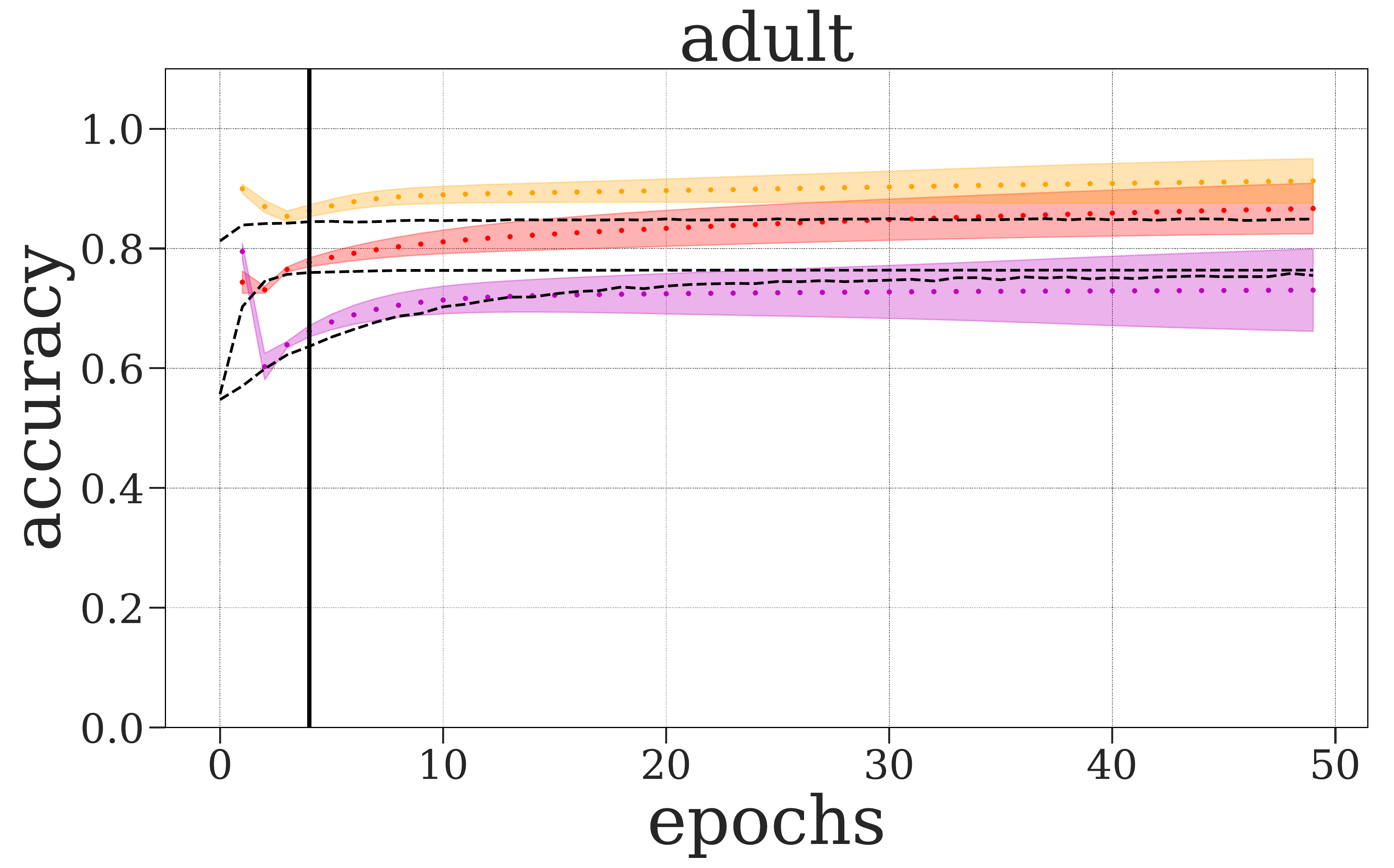}
        \includegraphics[width=.45\textwidth]{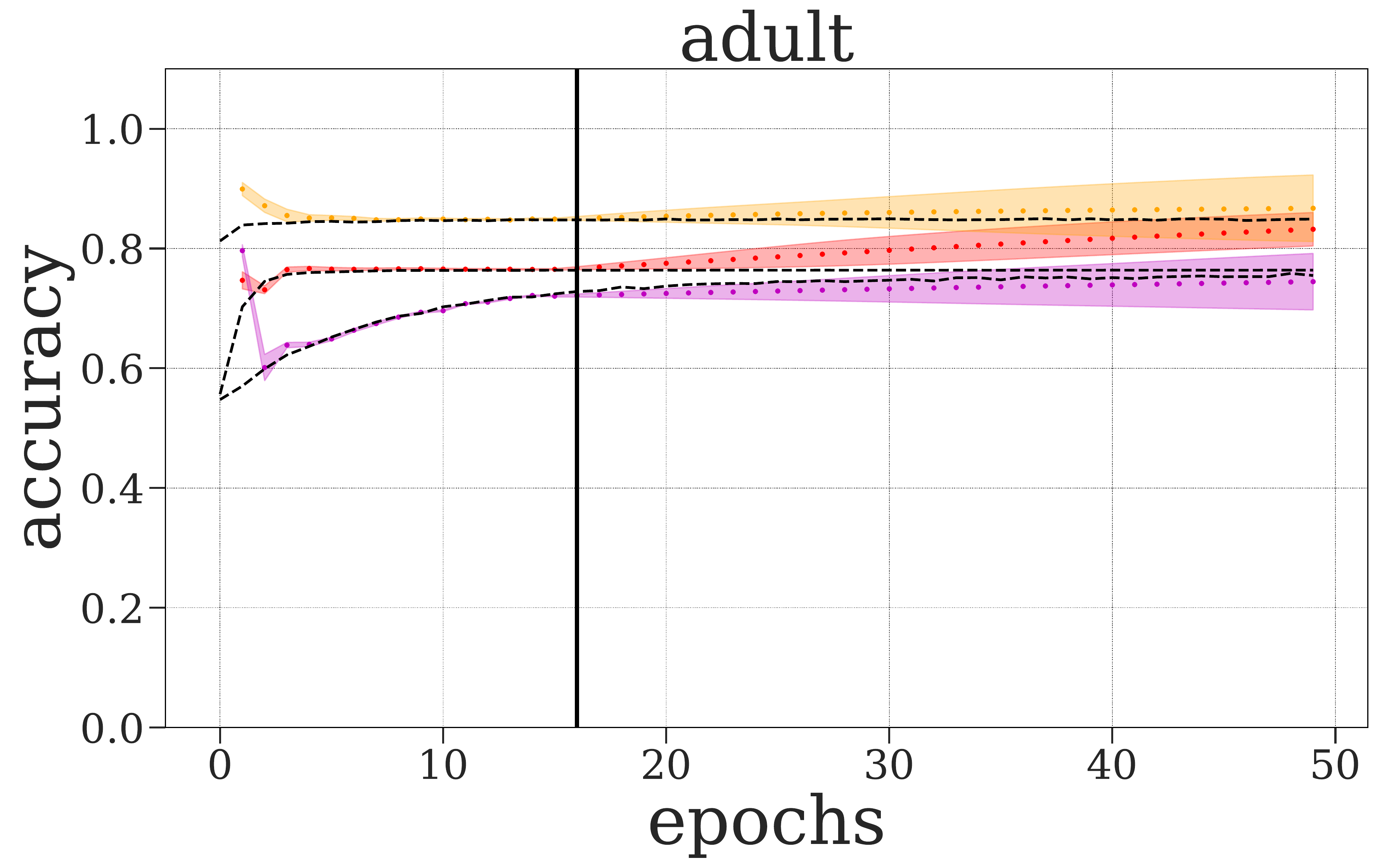}
        \includegraphics[width=.45\textwidth]{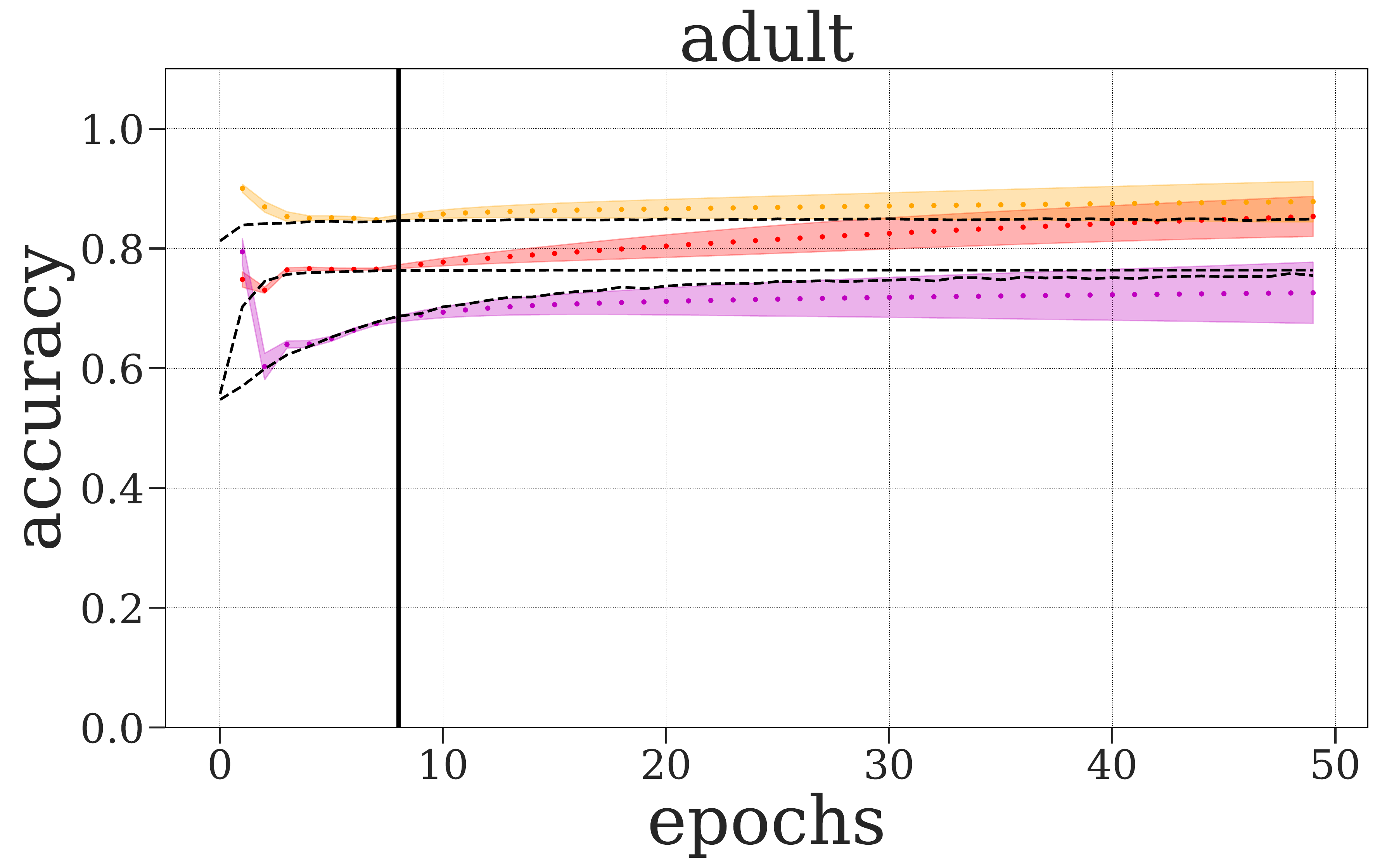}
        \includegraphics[width=.45\textwidth]{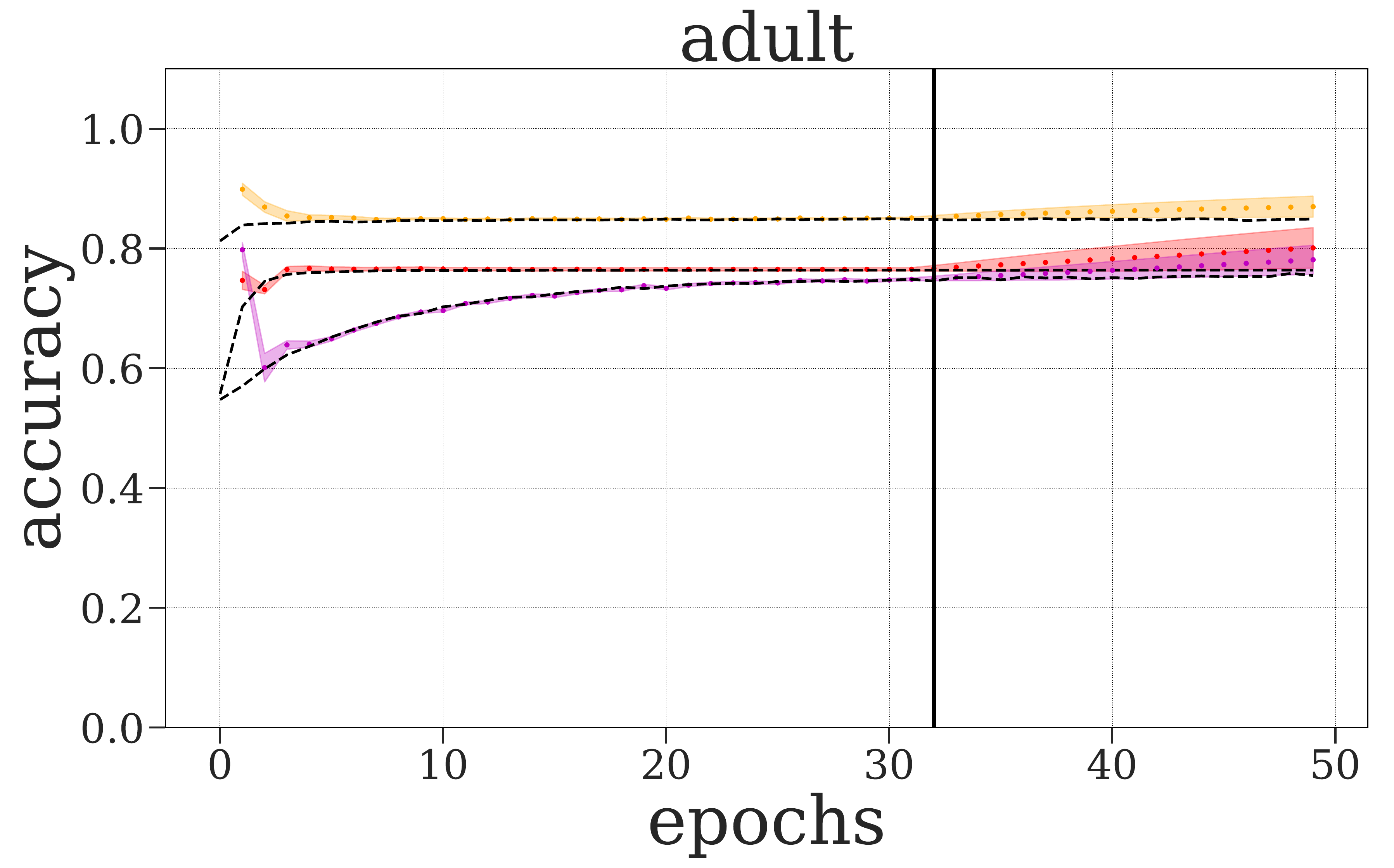}
    \end{center}
    \caption{Qualitative assessment of VRNN predictions on the Adult benchmark when trained on MNIST for different numbers of observed epochs at test time (the black vertical line).}
    \label{fig:vrnn_adult_unseen_dataset}
\end{figure} 

\begin{figure}[H]
    \begin{center}
        \includegraphics[width=.45\textwidth]{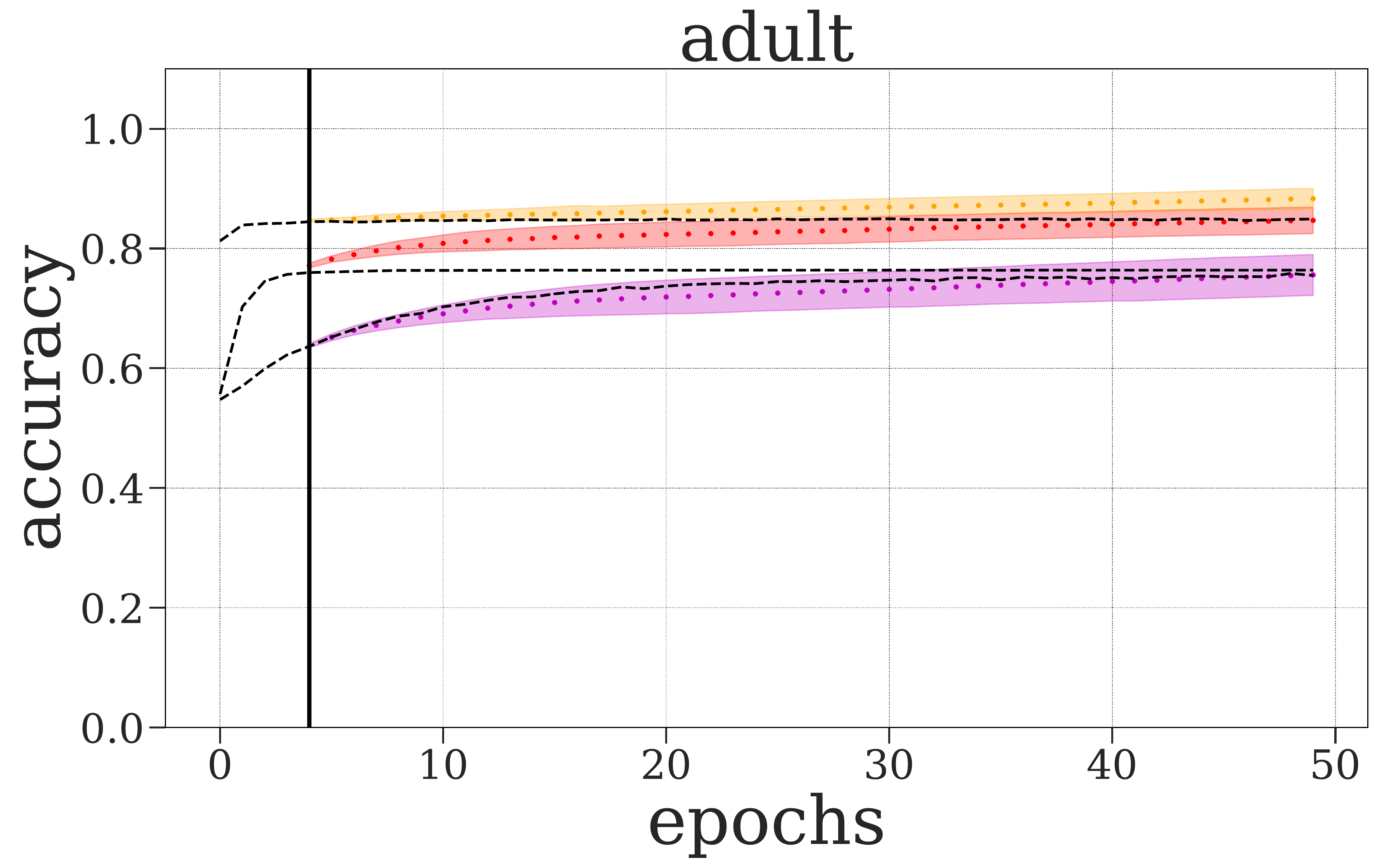}
        \includegraphics[width=.45\textwidth]{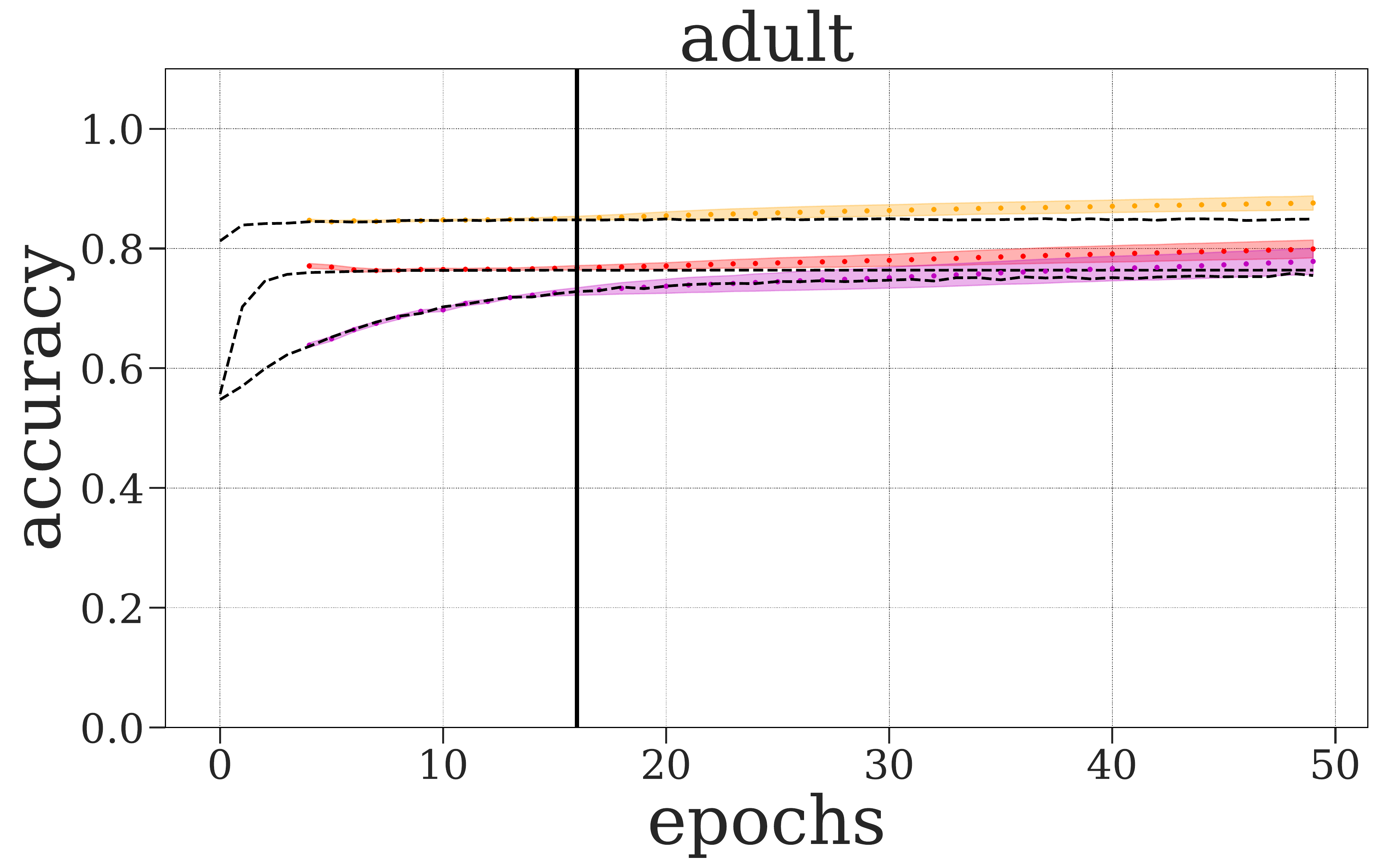}
        \includegraphics[width=.45\textwidth]{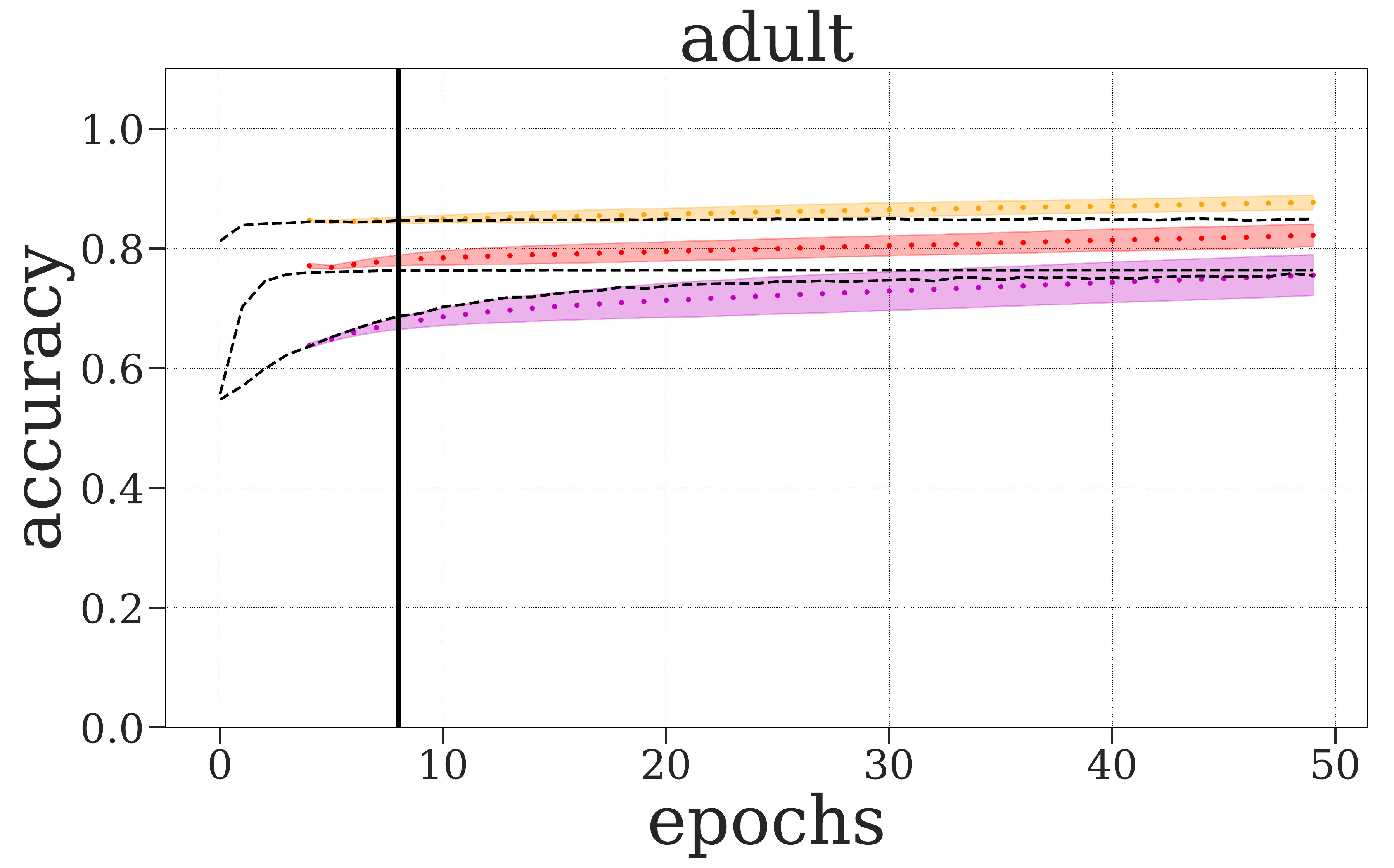}
        \includegraphics[width=.45\textwidth]{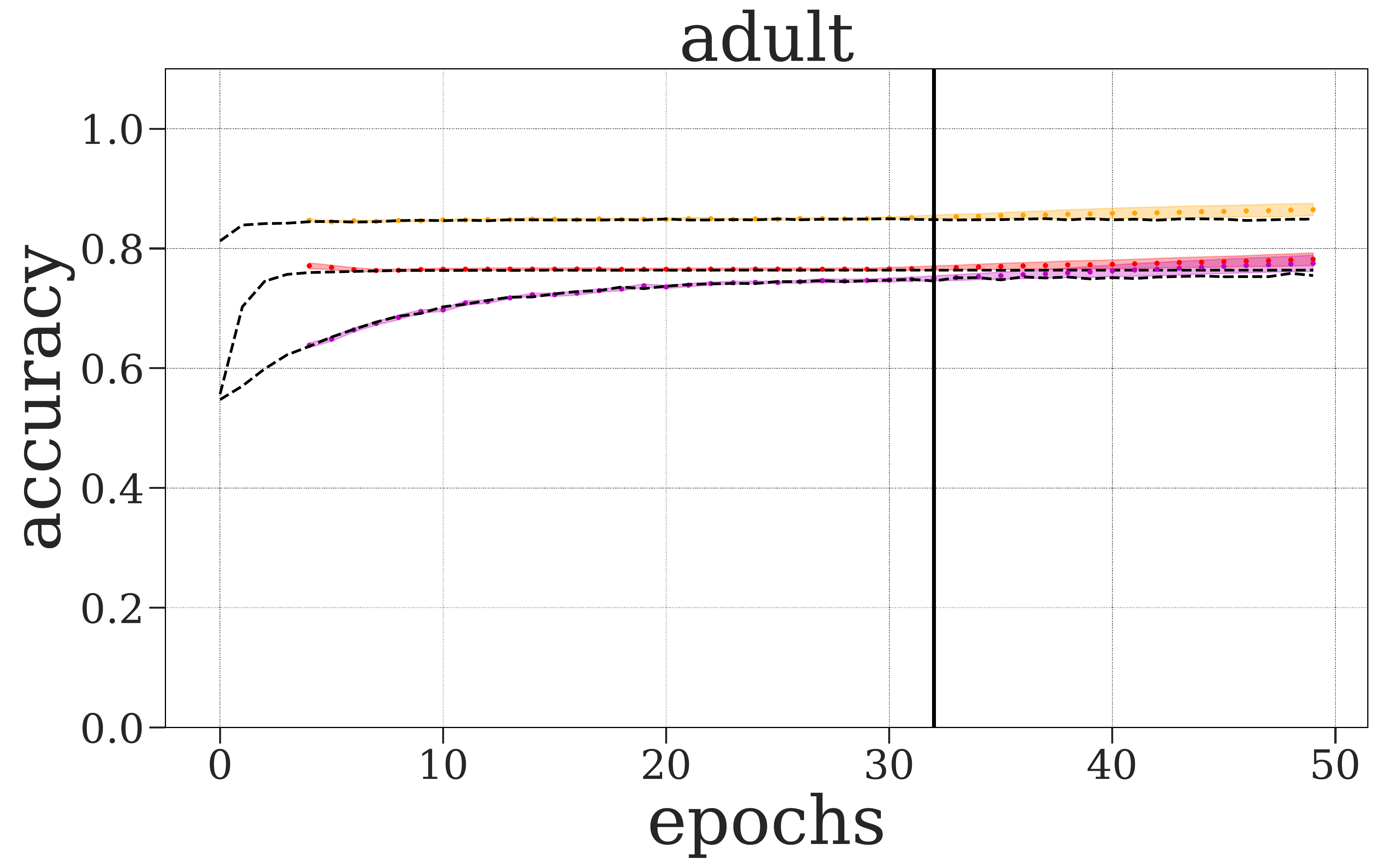}
    \end{center}
    \caption{Qualitative assessment of RF 4 predictions on the Adult benchmark when trained on MNIST for different numbers of observed epochs at test time (the black vertical line).}
    \label{fig:rfr_4_adult_unseen_dataset}
\end{figure}

\begin{figure}[H]
    \begin{center}
        \includegraphics[width=.45\textwidth]{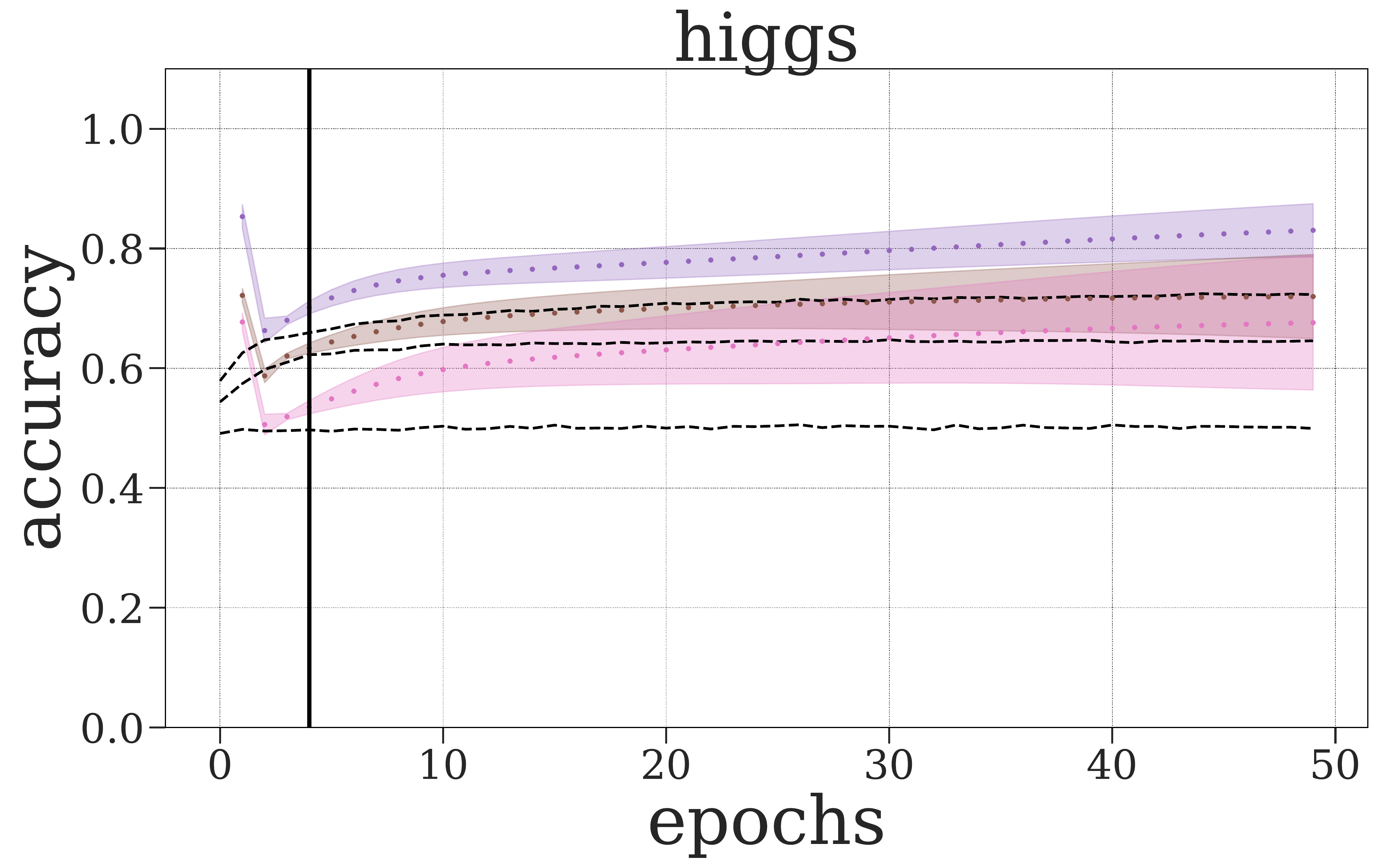}
        \includegraphics[width=.45\textwidth]{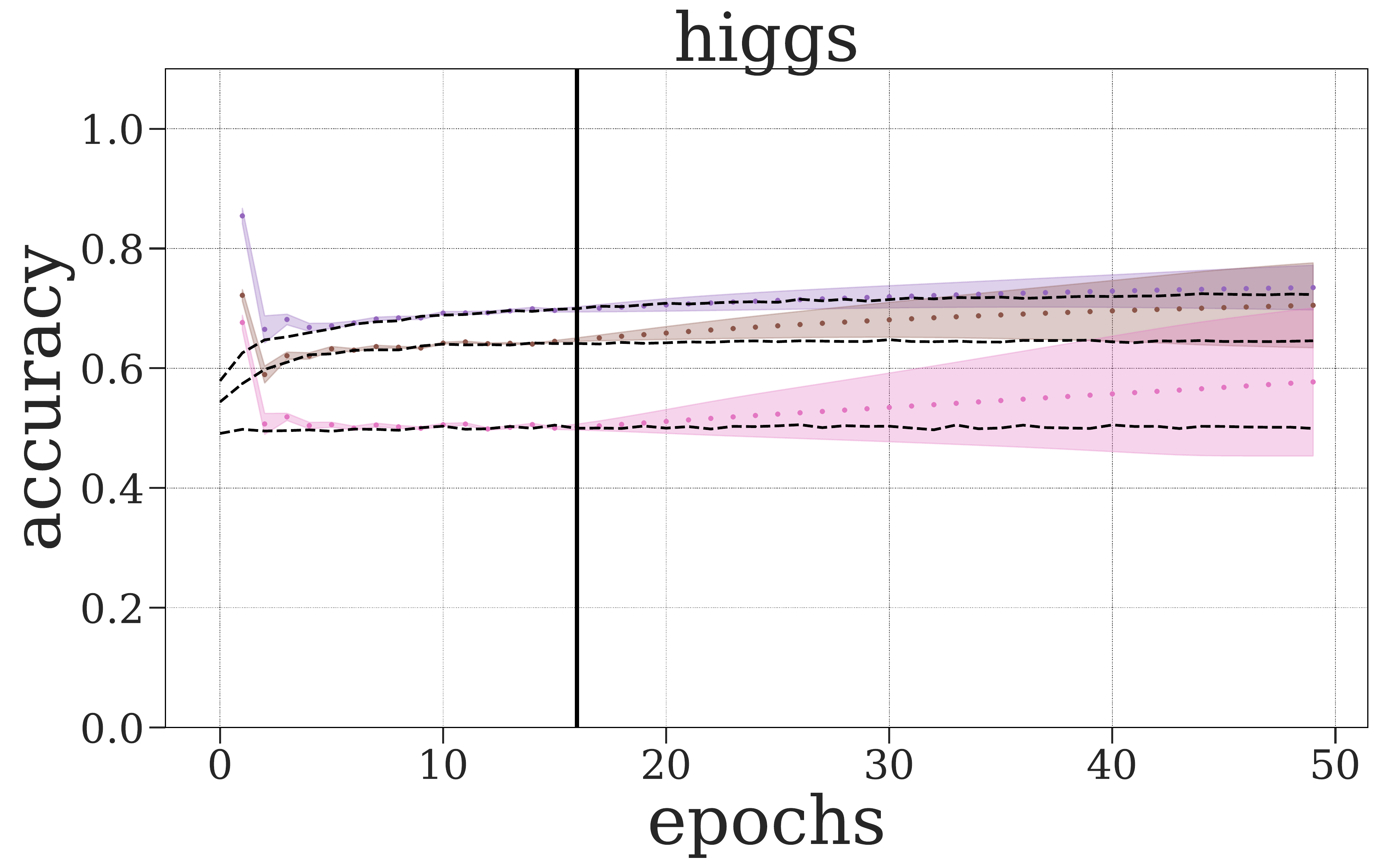}
        \includegraphics[width=.45\textwidth]{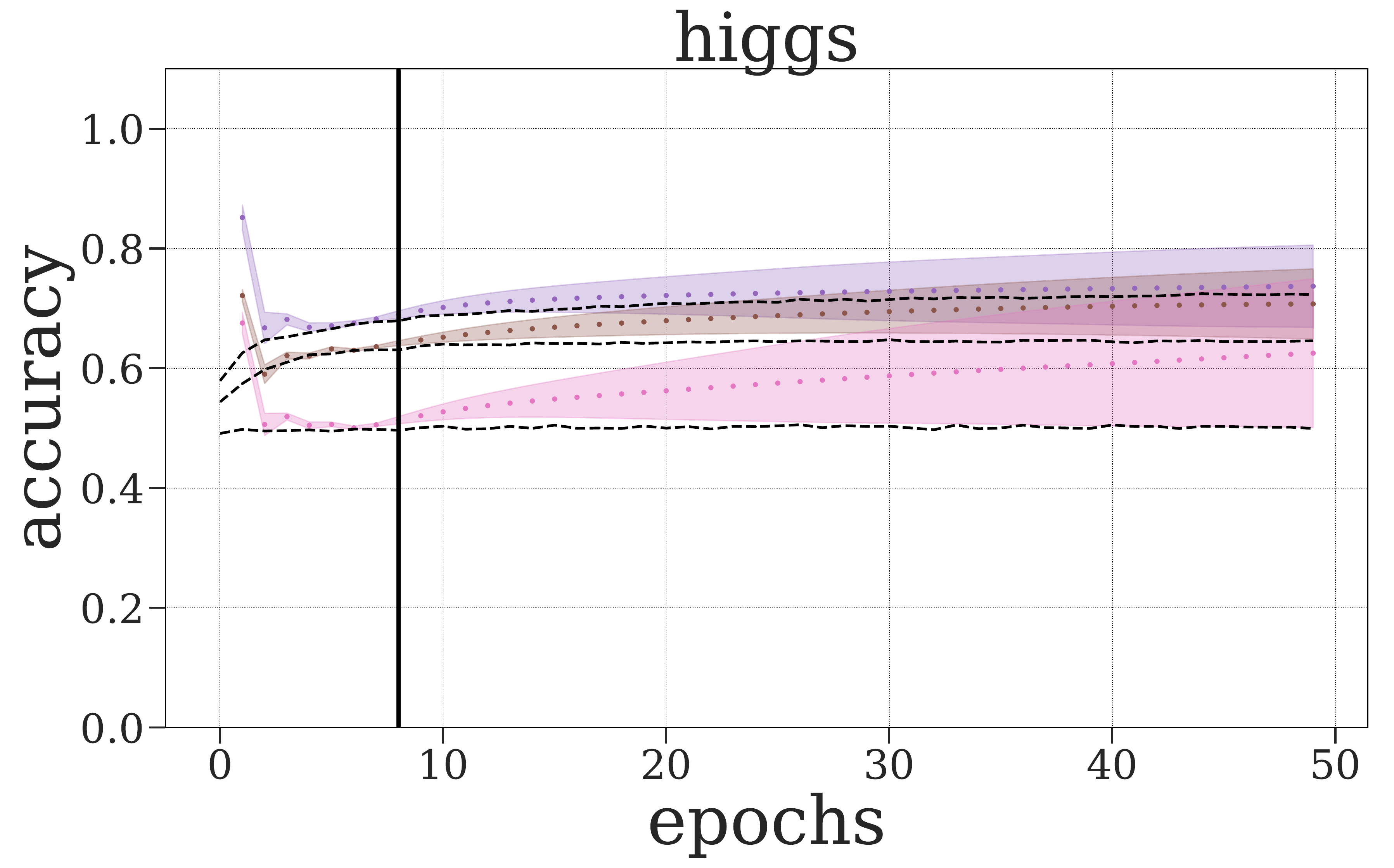}
        \includegraphics[width=.45\textwidth]{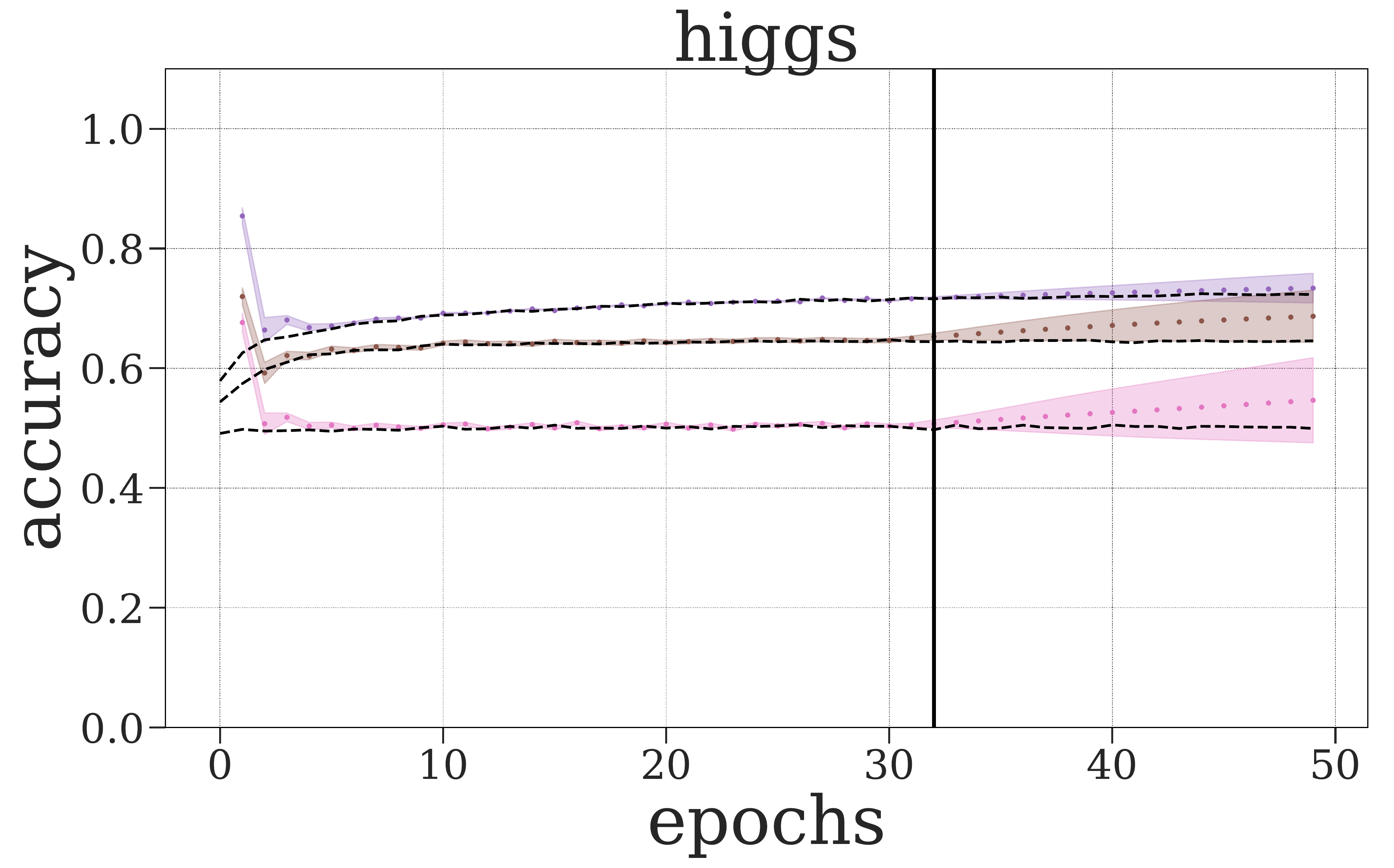}
    \end{center}
    \caption{Qualitative assessment of VRNN predictions on the Higgs benchmark when trained on MNIST for different numbers of observed epochs at test time (the black vertical line).}
    \label{fig:vrnn_higgs_unseen_dataset}
\end{figure} 

\begin{figure}[H]
    \begin{center}
        \includegraphics[width=.45\textwidth]{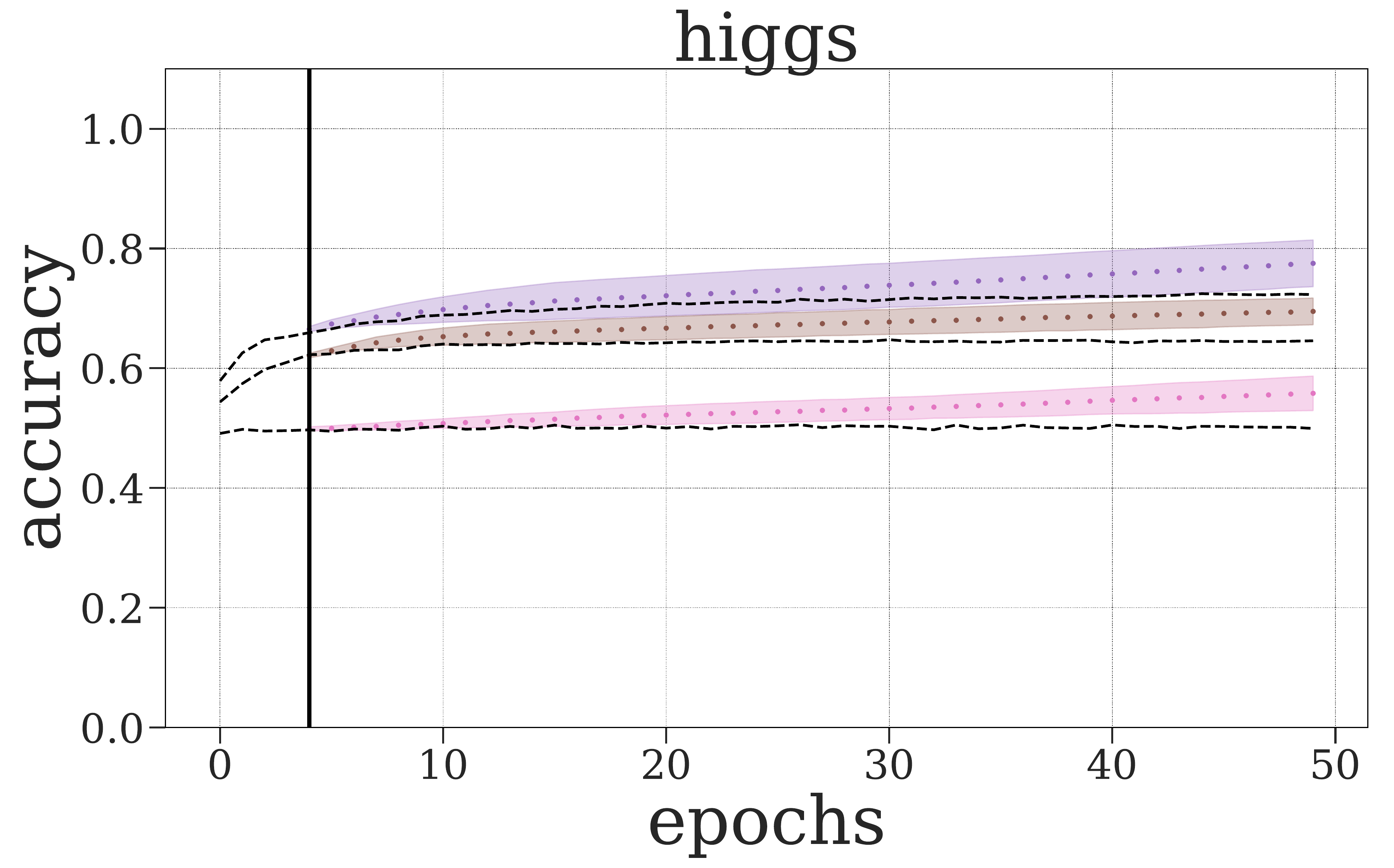}
        \includegraphics[width=.45\textwidth]{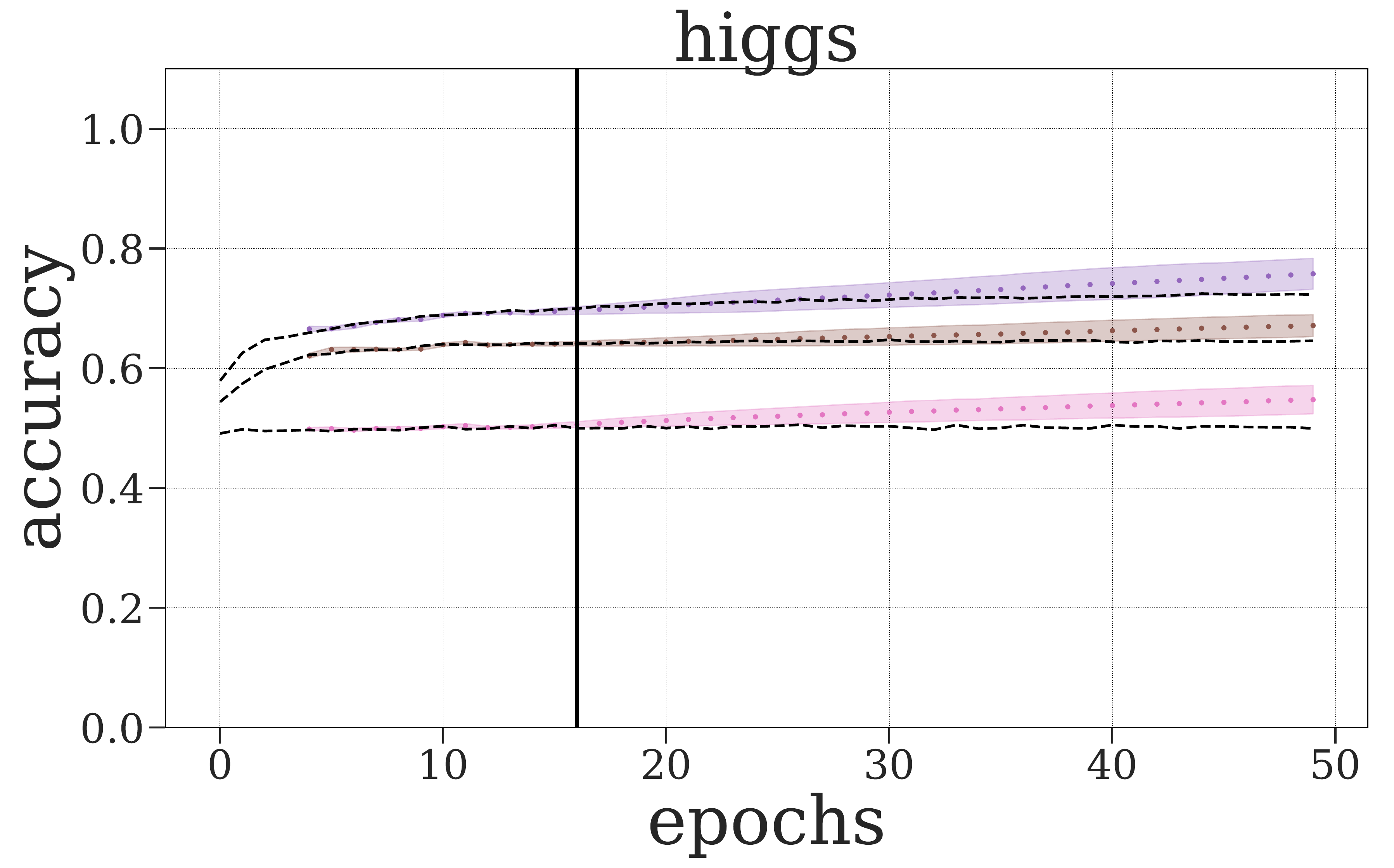}
        \includegraphics[width=.45\textwidth]{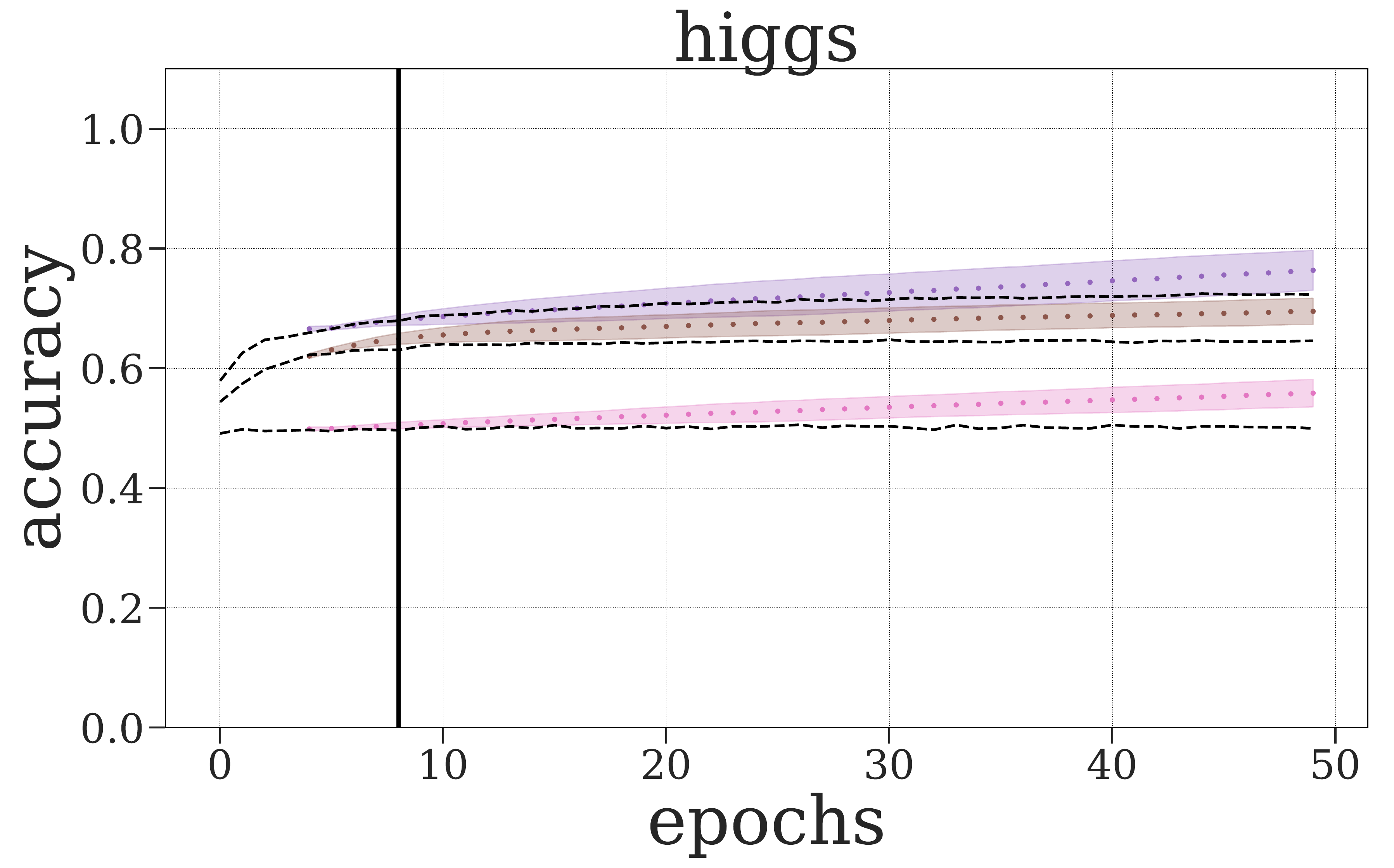}
        \includegraphics[width=.45\textwidth]{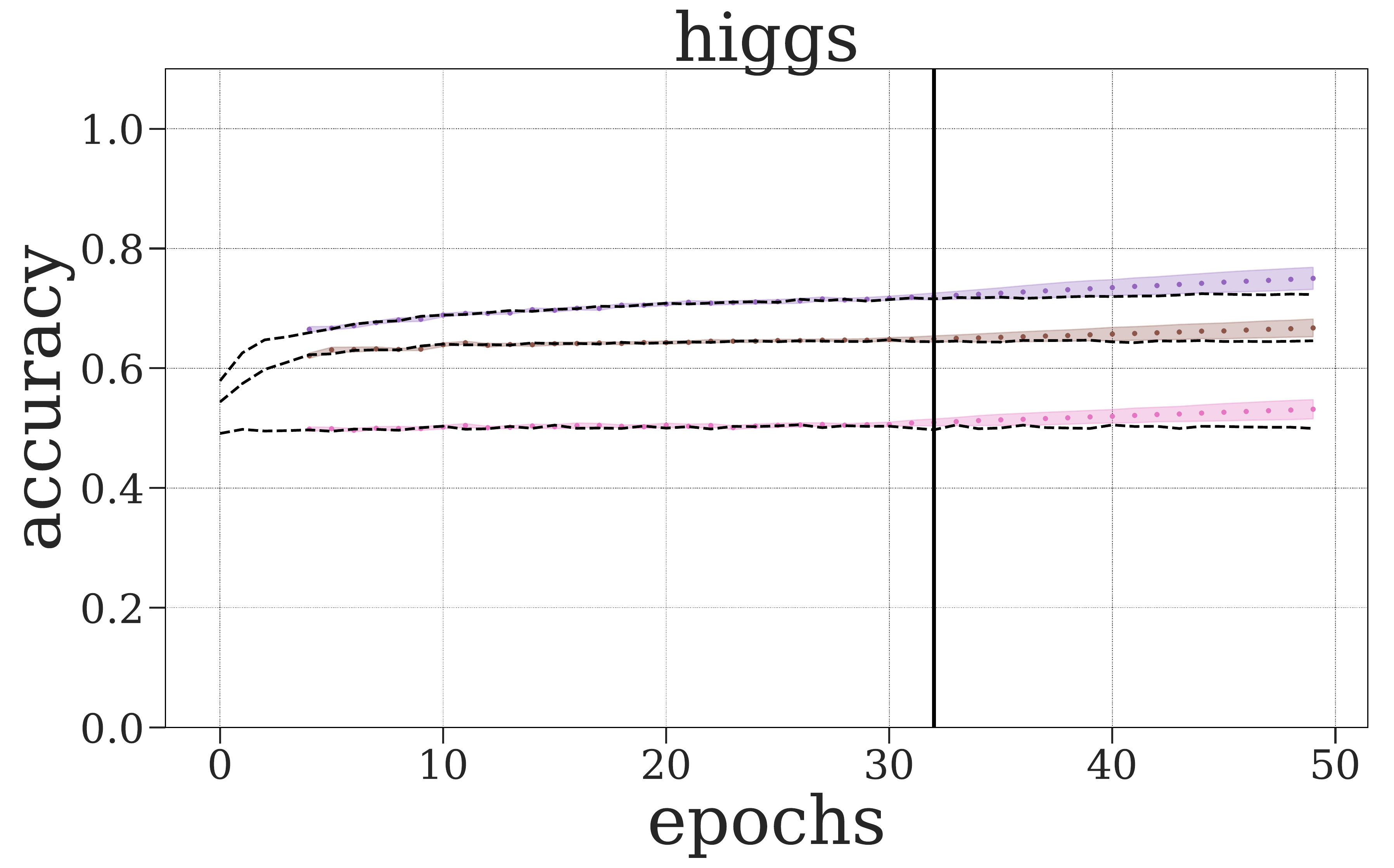}
    \end{center}
    \caption{Qualitative assessment of RF 4 predictions on the Higgs benchmark when trained on MNIST for different numbers of observed epochs at test time (the black vertical line).}
    \label{fig:rfr_4_higgs_unseen_dataset}
\end{figure} 

%%%%%%%%%%%%%%%%%%
\begin{figure}[H]
        \centering
        \begin{subfigure}[b]{0.475\textwidth}
            \centering
            \includegraphics[width=\textwidth]{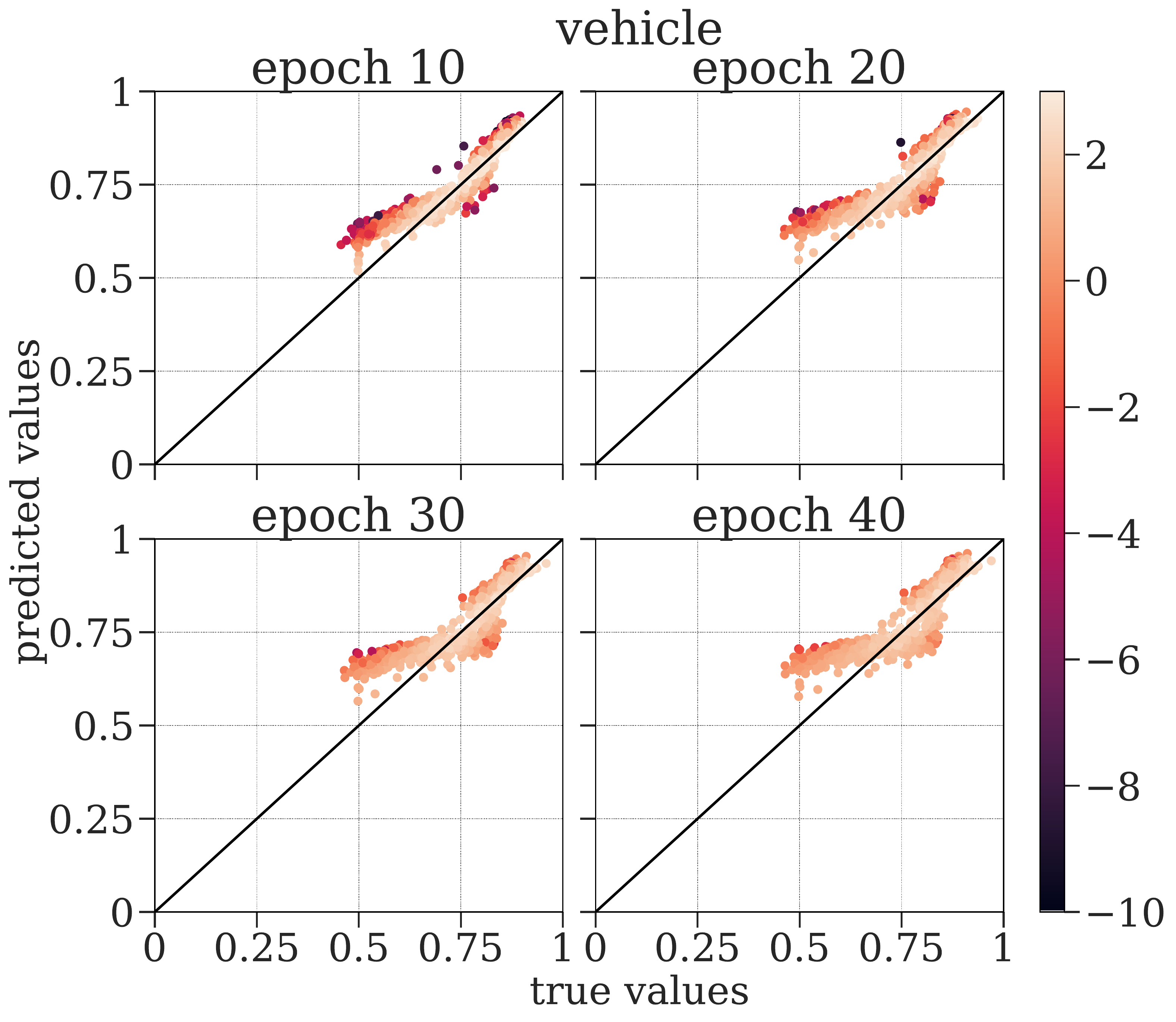}
            \caption[VRNN]%
            {{\small VRNN}}    
            \label{fig:VRNN_unseen_vehicle_predicted_true}
        \end{subfigure}
        \hfill
        \begin{subfigure}[b]{0.475\textwidth}  
            \centering 
            \includegraphics[width=\textwidth]{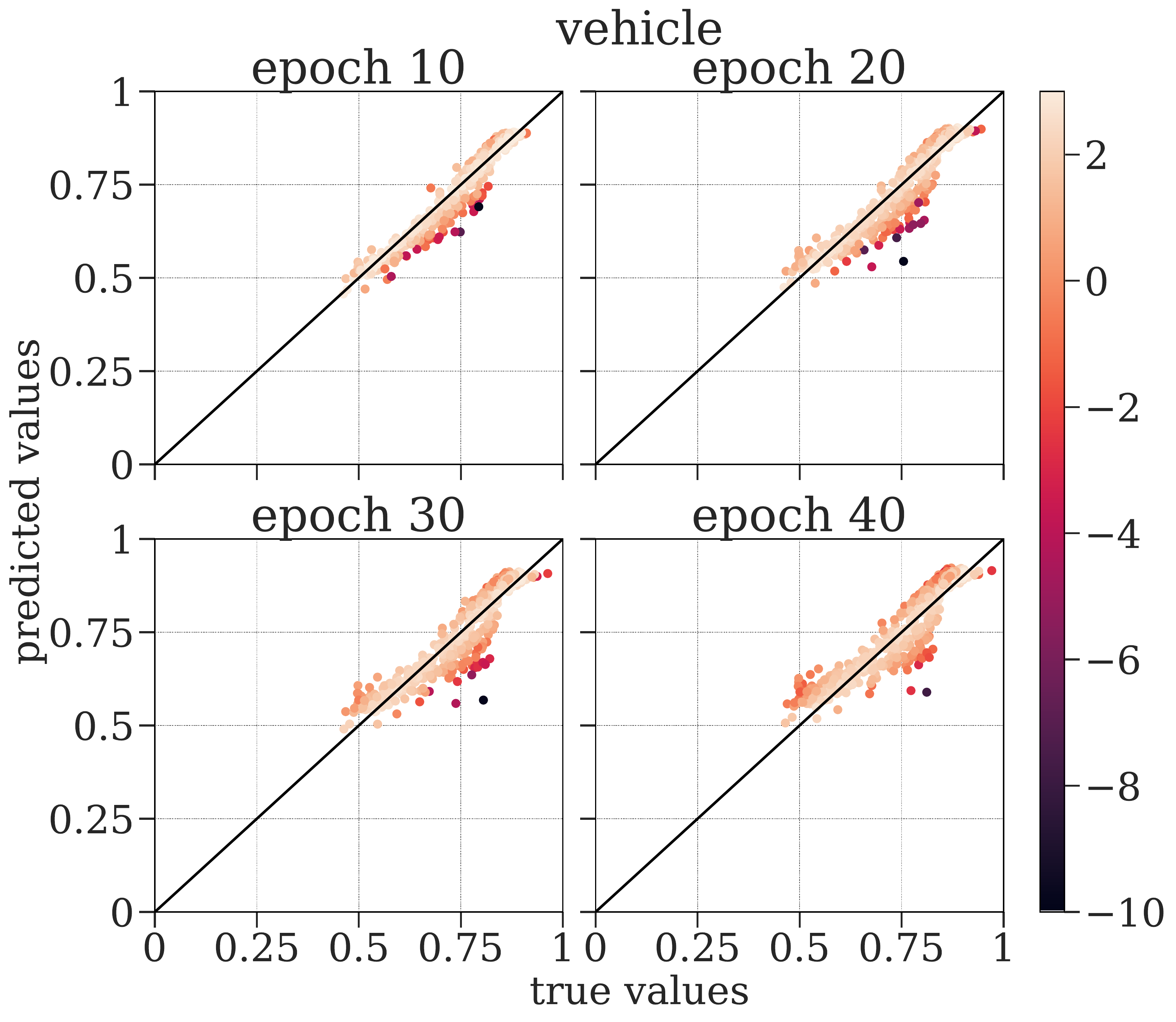}
            \caption[RF 4]%
            {{\small RF 4}}    
            \label{fig:RFR_4_unseen_vehicle_predicted_true}
        \end{subfigure}
        \vskip\baselineskip
        \caption[The panels show on the horizontal axis the true values and on the vertical axis the predicted values on the Vehicle benchmark for VRNN with 4 observed points (left) and RF 4 (right) with 4 observed epochs at test time when trained on MNIST. Each point is colored based on its log-likelihood value.]
        {\small The panels show on the horizontal axis the true values and on the vertical axis the predicted values on Vehicle benchmark for VRNN (left) and RF 4 (right) when trained on MNIST. Each point is colored based on its log-likelihood value.} 
        \label{fig:unseen_vehicle_predicted_true}
        \end{figure}
        
\begin{figure}[H]
        \centering
        \begin{subfigure}[b]{0.475\textwidth}
            \centering
            \includegraphics[width=\textwidth]{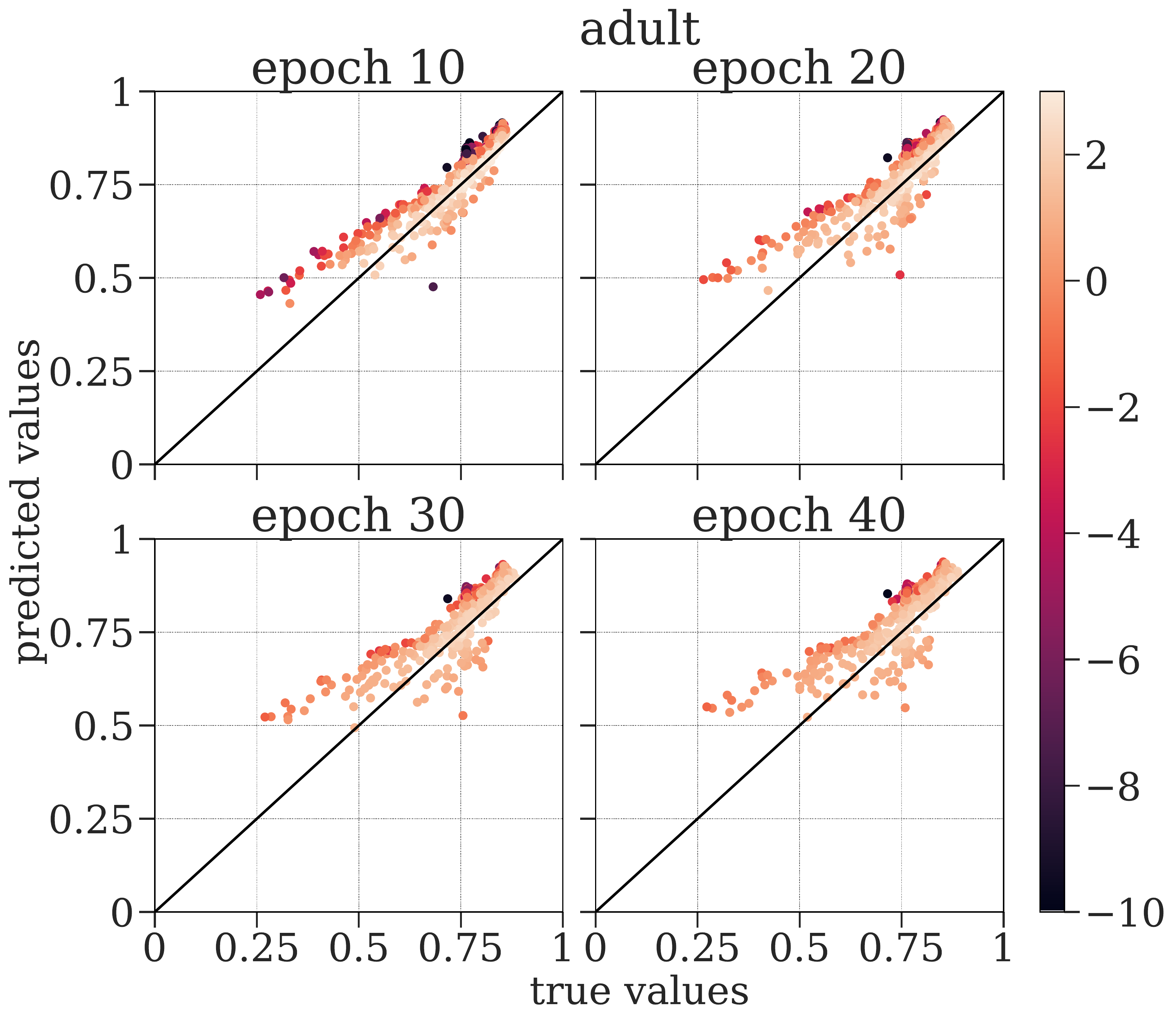}
            \caption[VRNN]%
            {{\small VRNN}}    
            \label{fig:VRNN_unseen_adult_predicted_true}
        \end{subfigure}
        \hfill
        \begin{subfigure}[b]{0.475\textwidth}  
            \centering 
            \includegraphics[width=\textwidth]{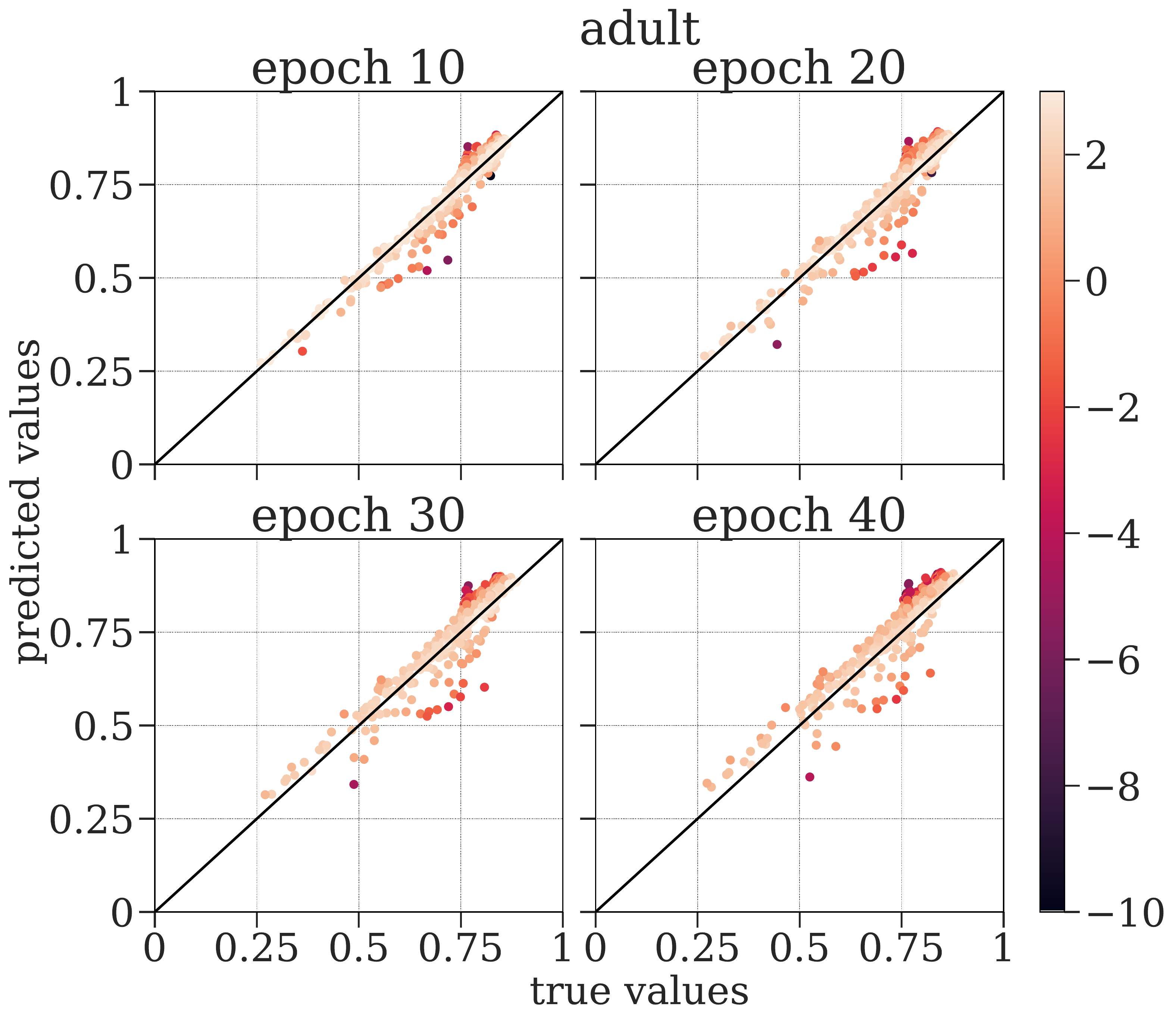}
            \caption[RF 4]%
            {{\small RF 4}}    
            \label{fig:RFR_4_unseen_adult_predicted_true}
        \end{subfigure}
        \vskip\baselineskip
        \caption[The panels show on the horizontal axis the true values and on the vertical axis the predicted values on Adult benchmark for VRNN (left) and RF 4 (right) with 4 observed epochs at test time when trained on MNIST. Each point is colored based on its log-likelihood value.]
        {\small The panels show on the horizontal axis the true values and on the vertical axis the predicted values on the Adult benchmark for VRNN with 4 observed points (left) and RF 4 (right) when trained on MNIST. Each point is colored based on its log-likelihood value.} 
        \label{fig:unseen_adult_predicted_true}
        \end{figure}
        
        \begin{figure}[H]
        \centering
        \begin{subfigure}[b]{0.475\textwidth}
            \centering
            \includegraphics[width=\textwidth]{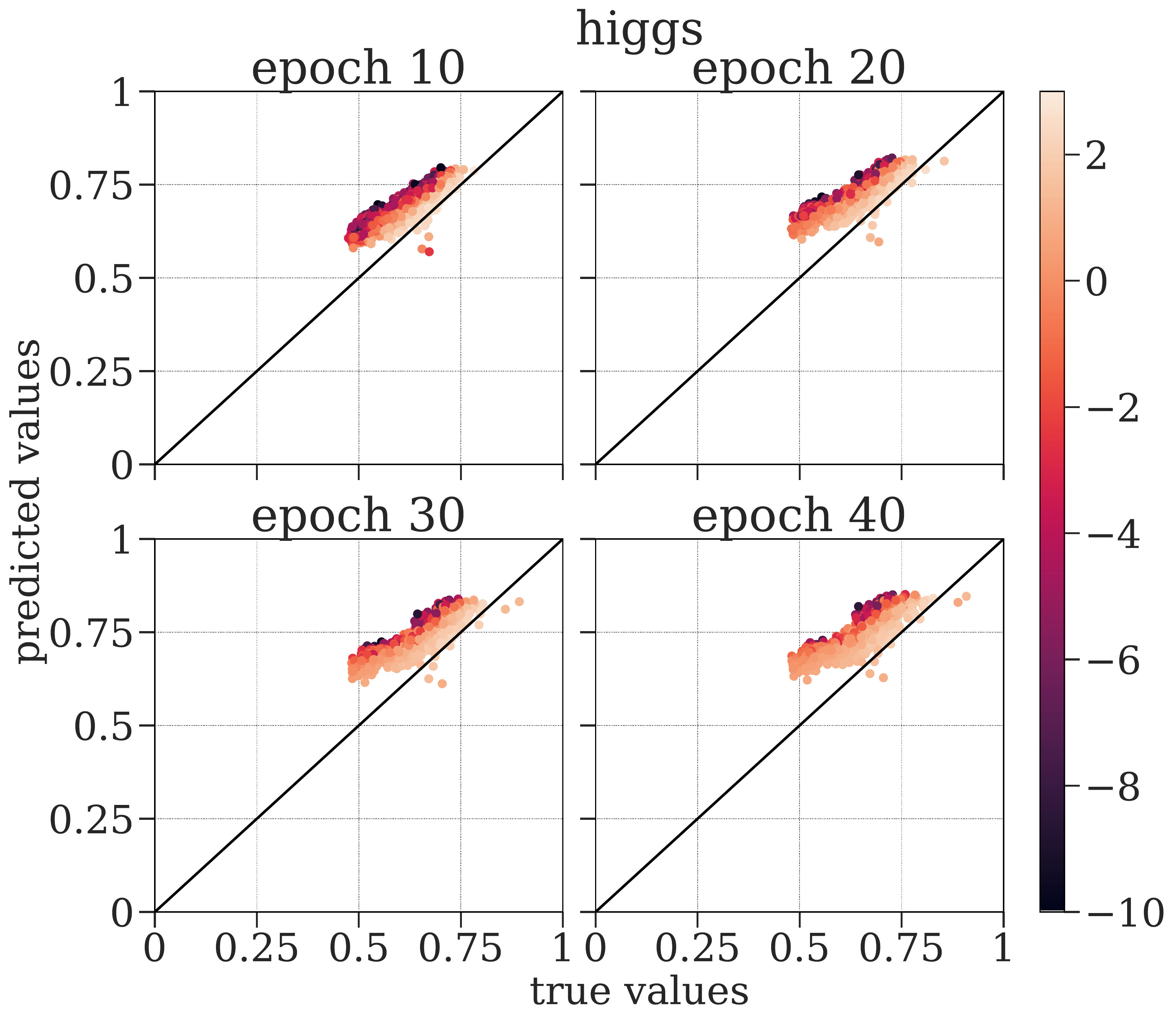}
            \caption[VRNN]%
            {{\small VRNN}}    
            \label{fig:VRNN_unseen_higgs_predicted_true}
        \end{subfigure}
        \hfill
        \begin{subfigure}[b]{0.475\textwidth}  
            \centering 
            \includegraphics[width=\textwidth]{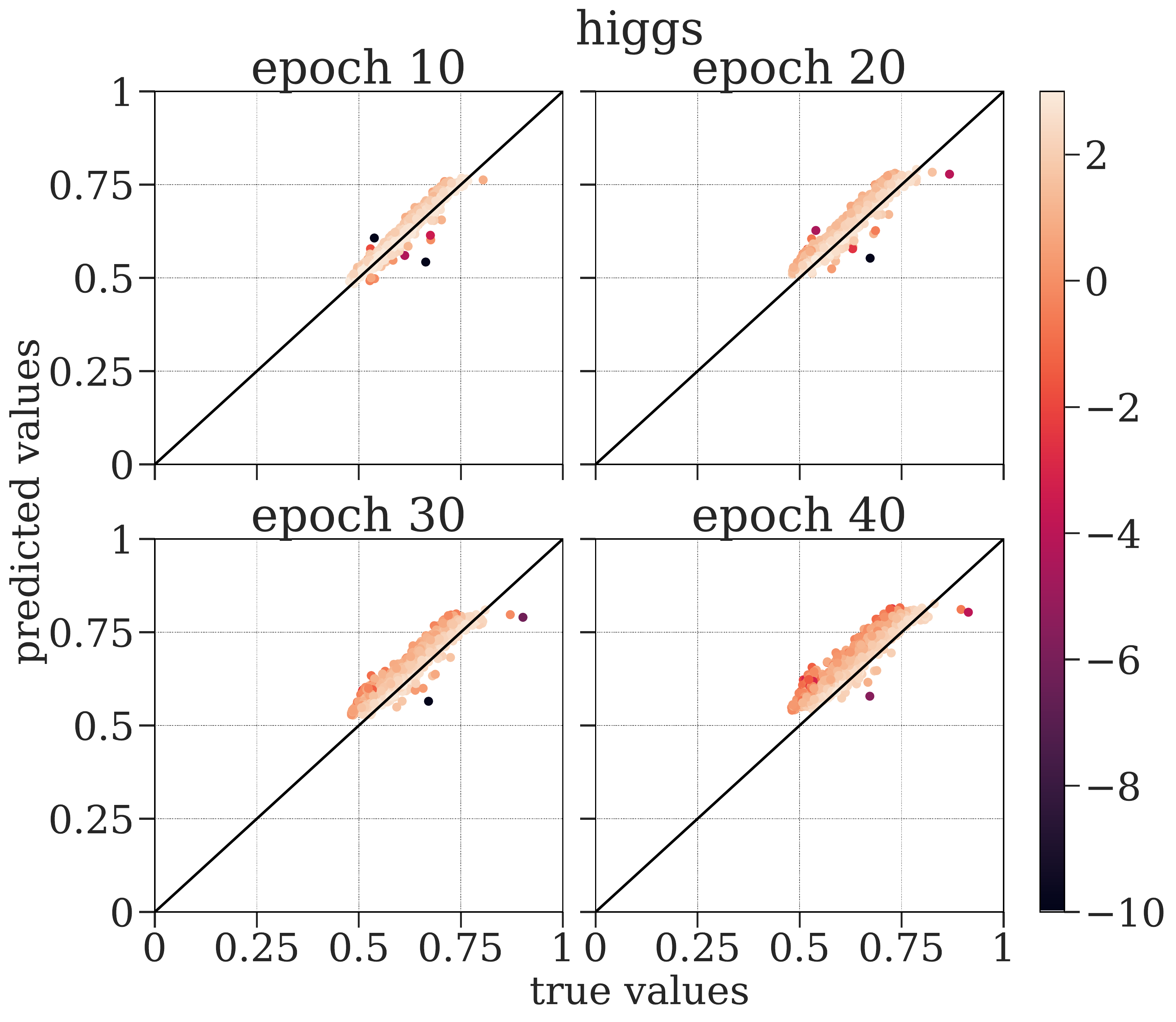}
            \caption[RF 4]%
            {{\small RF 4}}    
            \label{fig:RFR_4_unseen_higgs_predicted_true}
        \end{subfigure}
        \vskip\baselineskip
        \caption[The panels show on the horizontal axis the true values and on the vertical axis the predicted values on Higgs benchmark for VRNN (left) and RF 4 (right) with 4 observed epochs at test time when trained on MNIST. Each point is colored based on its log-likelihood value.]
        {\small The panels show on the horizontal axis the true values and on the vertical axis the predicted values on the Higgs benchmark for VRNN with 4 observed points (left) and RF 4 (right) when trained on MNIST. Each point is colored based on its log-likelihood value.} 
        \label{fig:unseen_higgs_predicted_true}
        \end{figure}
%\section{Additional Hyperband Experiments}

%\begin{figure}[ht]
%    \begin{center}
%        \includegraphics[width=.4\textwidth]{plot/comparison_0.pdf}
%        \includegraphics[width=.4\textwidth]{plot/comparison_1.pdf}
%        \includegraphics[width=.4\textwidth]{plot/comparison_2.pdf}
%        \includegraphics[width=.4\textwidth]{plot/comparison_3.pdf}
%    \end{center}
%    \caption{}
%    \label{fig:comparison_hyperband}
%\end{figure} 

\end{document}